\pdfoutput=1

\documentclass[11pt]{article}

\usepackage[final]{acl}
\usepackage{microtype}
\usepackage{graphicx}
\usepackage{subfigure}
\usepackage{booktabs} 
\usepackage{multirow}
\usepackage{makecell} 
\usepackage{array}    
\usepackage{caption}
\usepackage{amsmath}
\usepackage{adjustbox}
\usepackage{float} 
\usepackage[perpage]{footmisc} 

\usepackage{hyperref}

\usepackage{times}
\usepackage{latexsym}

\usepackage[T1]{fontenc}

\usepackage[utf8]{inputenc}

\usepackage{microtype}

\usepackage{inconsolata}

\usepackage{graphicx}

\title{Lifelong Knowledge Editing requires Better Regularization}

\author{%
  \textbf{Akshat Gupta\textsuperscript{1*}, Phudish Preteepamornkul\textsuperscript{1,2*}, Maochuan Lu\textsuperscript{1}}\\
  \textbf{Ahmed Alaa\textsuperscript{1}, Thomas Hartvigsen\textsuperscript{3}, Gopala Anumanchipalli\textsuperscript{1}}\\[1ex]
  \textsuperscript{1}UC Berkeley, \textsuperscript{2}SCB DataX, \textsuperscript{3}University of Virginia\\[1ex]
  \texttt{akshat.gupta@berkeley.edu}
}

\begin{document}

\maketitle
\begin{abstract}
Knowledge editing is a promising way to improve factuality in large language models, but recent studies have shown significant model degradation during sequential editing. In this paper, we formalize the popular locate-then-edit methods as a two-step fine-tuning process, allowing us to precisely identify the root cause of this degradation. We show that model degradation occurs due to (1) over-optimization of internal activations and (2) continuous norm-growth of edited matrices. To mitigate these issues, we introduce two regularization techniques: (1) Most-Probable Early Stopping (MPES) and (2) explicit Frobenius norm-constraint. We demonstrate that applying these simple yet effective regularization techniques at key points in the editing process can substantially mitigate model degradation. Combining these regularization methods enables scaling locate-then-edit methods to 10,000 edits while reducing editing time by 42-61\%. These results show that targeted regularization is essential for lifelong knowledge editing.

\end{abstract}

\section{Introduction}

Knowledge editing entails editing specific facts that a language model has learned during pre-training in a data and computate efficient manner \cite{metamodel, editing-survey}. The most popular editors follow a ``locate-then-edit'' approach, where the methods aim to locate and update small subsets of a model's parameters that recall target facts \cite{ROME, MEMIT, akshat-rebuilding, akshat-unified, ma2024perturbation, alphaedit}. Although these methods have shown strong results when performing singular or small-scale edits, they suffer substantial degradation when applied at scale \cite{hurt, akshat-catastrophic, thede2025understanding}, leaving the problem of lifelong knowledge editing largely unsolved. 
\let\thefootnote\relax\footnote{* Equal Contribution}


\begin{figure}[htbp]
    \centering

    \subfigure[Norm-growth as function of number of edits]{
        \includegraphics[width=0.46\linewidth]{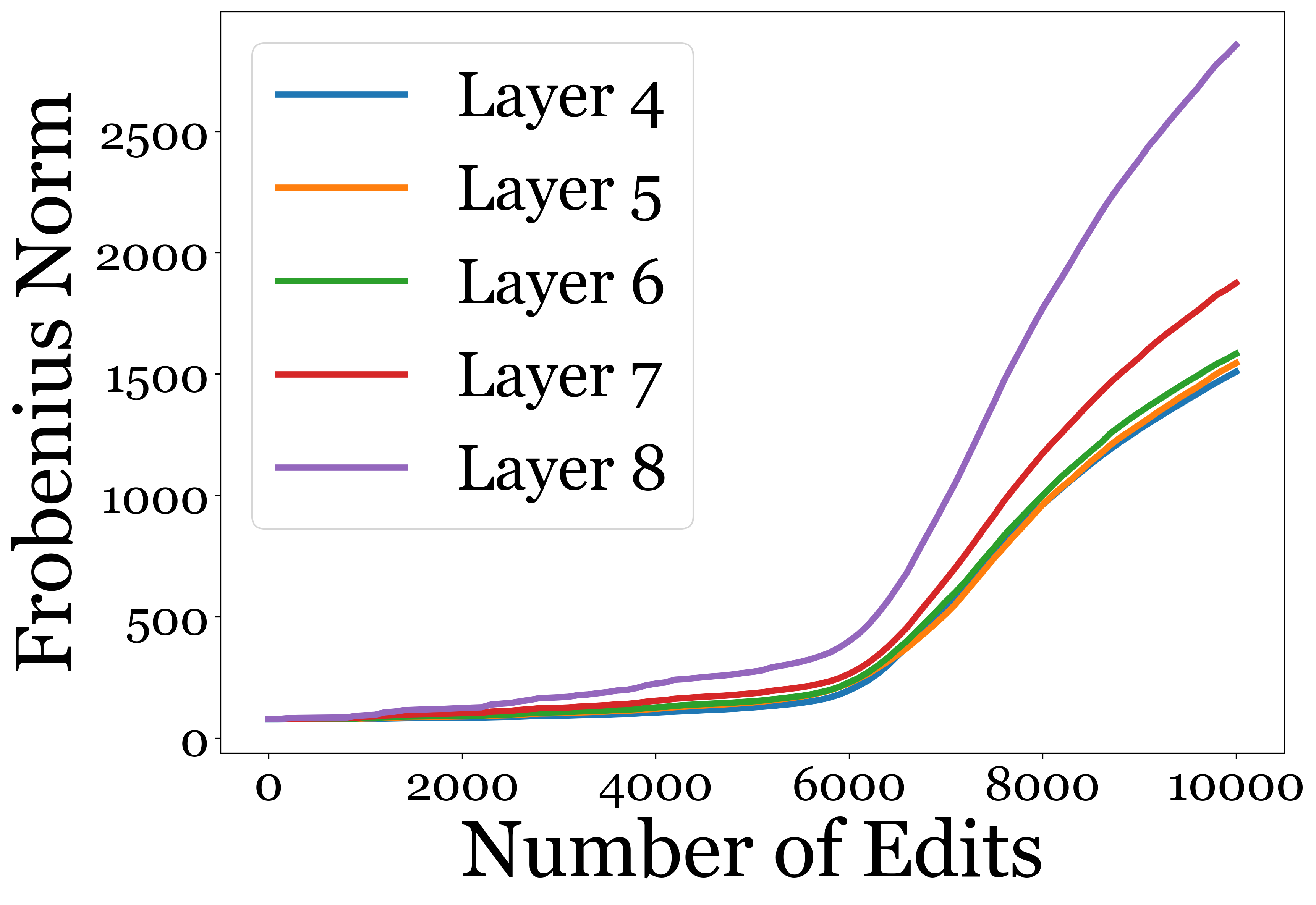} 
        \label{fig:norm-growth-MEMIT}
    }
    \subfigure[Layer wise norm post-editing]{
        \includegraphics[width=0.46\linewidth]{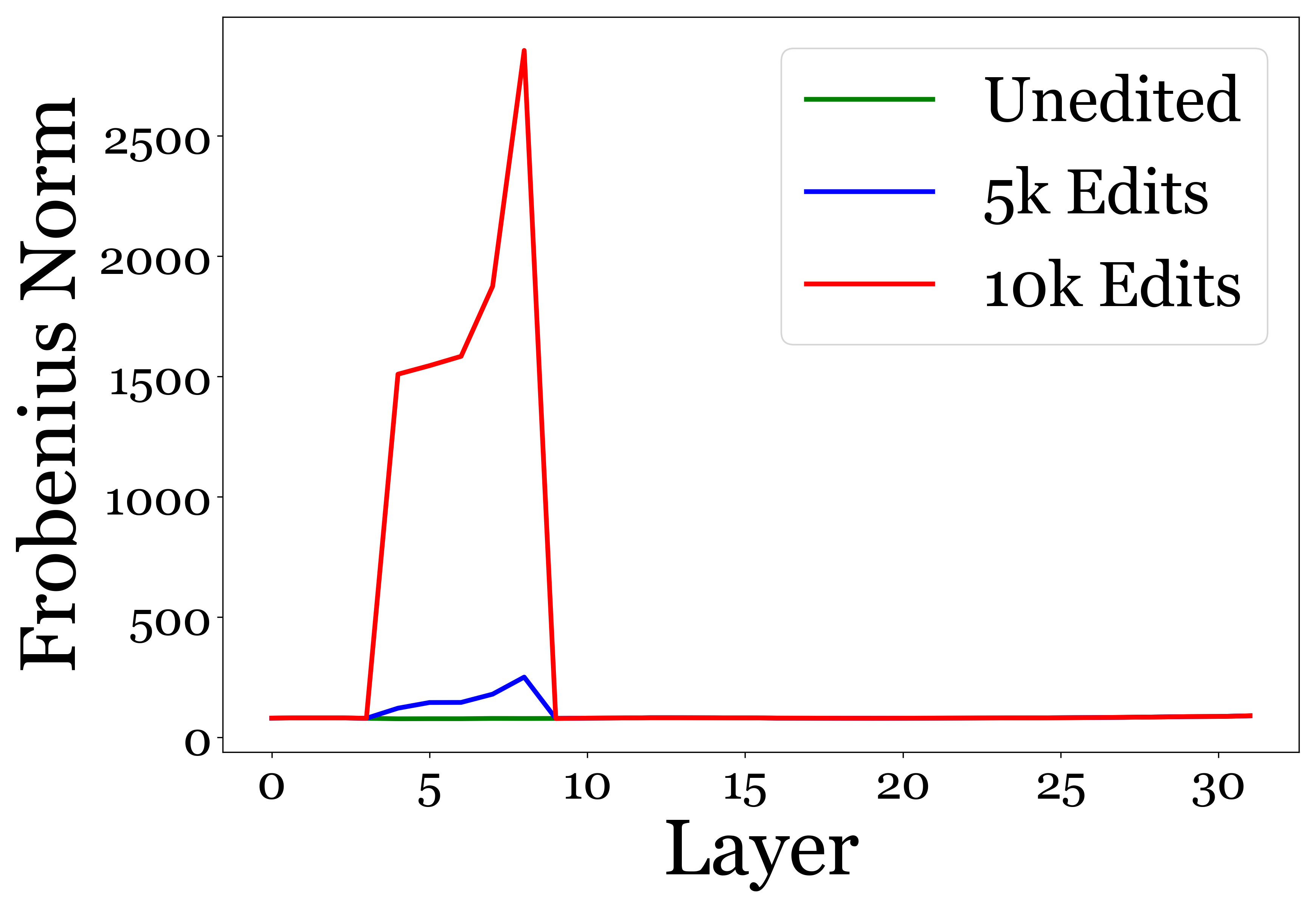} 
        \label{fig:norm-growth-MEMIT-layer-wise}
    }
    \vskip -0.1in
    \caption{The continuous growth of norm of edited MLP matrices in MEMIT Llama3-8B during sequential knowledge editing.}
    \label{fig:norm-growth-subset}
    \vskip -0.1in
\end{figure}

In this work, we systematically investigate the factors that cause model degradations and propose appropriate regularization methods to mitigate them, enabling large-scale sequential knowledge editing without sacrificing downstream performance. We begin by first formalizing locate-then-edit methods as a two-step fine-tuning process (section \ref{sec:background}), where the first step uses gradient-descent for activation optimization, and the second step performs weight-update via a least-squares update rule. This explicit structure allows us to precisely diagnose the causes of loss of downstream performance and allows us to choose appropriate interventions.


We show that model degradation occurs during continuous sequential editing because of two reasons. Firstly, we show that the gradient-descent step in locate-then-edit methods leads to over-optimization of internal activations, which makes the model predict edited facts with unnaturally high confidence (section \ref{sec:overfitting}). We mitigate this by proposing \textbf{M}ost-\textbf{P}robable \textbf{E}arly \textbf{S}topping (MPES) - a novel variant of early-stopping where we halt gradient-descent steps when edited facts become most probable across all the different contexts used to calculate the loss. Secondly, we show that sequential editing leads to disproportionate growth of the Frobenius norm of the edited matrix (Figure \ref{fig:norm-growth-subset}). This allows the outputs produced from the edited layers to have an abnormally larger influence on the final output of the model, inadvertently causing loss of general ability, which might require information coming from other parts of the model. To address this, we incorporate a Frobenius norm-constraint into the editing objective of locate-then-edit methods (section \ref{sec:norm-constraint}). While early stopping and norm-constraints are known regularization methods, their targeted stage-specific application to locate-then-edit methods is non-trivial. With our work, we demonstrate that there exists a lack of proper regularization in existing knowledge editing methods and show that explicit regularization at the appropriate stages of editing is essential and critical for scalable and stable knowledge editing. The code for our work can be found here - \url{https://github.com/scalable-model-editing/knowledge-editing-regularization}

Our proposed regularization methods mitigate the respective issues of over-optimization of intermediate activation vectors and disproportionate norm-growth, consequently preserving downstream performance for a larger number of edits. Finally, we show that combining these two regularization methods allows us to perform up to 10,000 sequential edits while maintaining original downstream performance and making editing 41-62\% faster.

\section{Related Work}
Knowledge editing methods can broadly be classified into three types \citep{editing-survey, survey-comprehensive} - \textit{hypernetwork based editing methods} \citep{metamodel, MEND}, where an additional model is trained to predict the edited weights of the model, \textit{in-context editing methods} \citep{SERAC, grace, mquake}, where edited information is added in-context, and \textit{locate-then-edit methods} \cite{ROME, MEMIT, akshat-unified, akshat-rebuilding, alphaedit}, where the knowledge source is traced to certain MLP layers \cite{localization-inform-editing, akshat-llama3} before modifying them using a two-step fine-tuning method (section \ref{sec:background}). This work focuses on performing locate-then-edit type of knowledge editing methods.

\paragraph{Overfitting During Knowledge Editing.} 
Some recent works have studied overfitting as a key challenge of knowledge editing. For example, 
\citet{lti} identifies that locate-then-edit methods overfit on edited facts. They propose to mitigate this by introducing an inference-stage regularization approach called ``learn to inference'' (LTI) that guides edited models to recall new knowledge by adding multi-stage constraints during optimization. With our proposed variant of early-stopping called "most-probable early stopping" (MPES), we present a much simpler way of reducing overfitting. We also show that LTI does not scale to multiple edits and leads to immediate model degradation when performing sequential editing. Meanwhile, MPES leads to significant improvements in downstream performance while countering overfitting (section \ref{sec:overfitting}).

\paragraph{Norm-Growth and Weight Regularization.} Prior work has shown that sequential knowledge editing leads to the norm-growth of the updated weight matrices \cite{akshat-catastrophic, disabling-butterfly}. \citet{prune} present a similar observation in the form of growth of condition number of the edited matrices as a potential cause of loss of downstream performance. However, none of these works explains why increasing norm of edited matrix is bad for the model. In our work, we analyze the effect of norm-growth on internal activations and show how it connects with model degradations (section \ref{sec:secret-mechanics}).  We verify our conclusions by adding a norm-constraint in the optimization objective of MEMIT. We show that our proposed norm-constraint combines seamlessly with MEMIT's closed-form solution, unlike prior regularization methods like PRUNE \cite{prune} and RECT \cite{rect}, which require additional regularization steps post-editing. We also show that our methods significantly outperform PRUNE and RECT in preserving downstream performance over a large number of edits.

\section{Methods, Models,    and Evaluation}\label{sec:evaluation}
We adopt the experimental setting of AlphaEdit \cite{alphaedit}, a recent locate-then-edit method that is able to perform sequential editing for up to 3,000 facts. Following their setting, we perform sequential edits in batches of 100 facts. This means that 100 facts are edited into the model with each weight update, and multiple such updates are performed sequentially. 



We evaluate all algorithms on three representative models - GPT2-XL \cite{gpt-2}, Llama2-7B \cite{llama2} and Llama3-8B \cite{akshat-llama3}. All experiments are performed on the CounterFact \cite{ROME} and zsRE \cite{MEND} datasets, which are standard knowledge editing datasets. We present the results for Llama2-7B and Llama3-8B on the CounterFact dataset in the main paper and present the remaining results in the appendix due to space constraints.


In this paper, we evaluate the editing algorithms along two dimensions - editing performance and downstream performance. The editing performance evaluates the success of the knowledge editing algorithm in making successful edits, while downstream performance evaluates the extent of model degradation following prior work \cite{alphaedit, akshat-catastrophic, hurt}.


\begin{figure}[t]
    \centering
    \subfigure[gradient-descent step which finds the target activations for the MLP matrix.]{
        \includegraphics[width=0.45\linewidth]{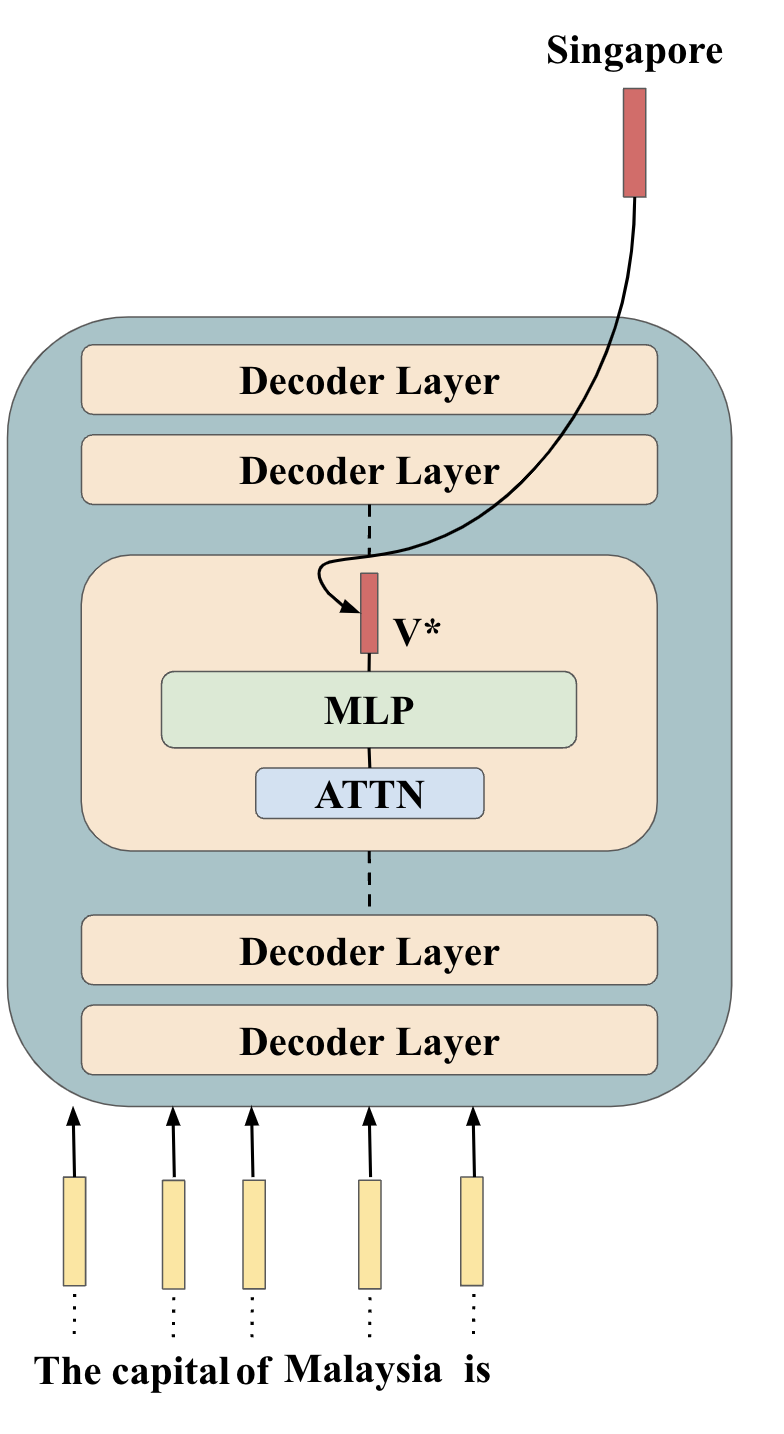} 
        \label{fig:editing-process-gd}
    }
    \hfill
    \subfigure[Target activations are used to update the second MLP matrix (in red).]{
        \includegraphics[width=0.45\linewidth]{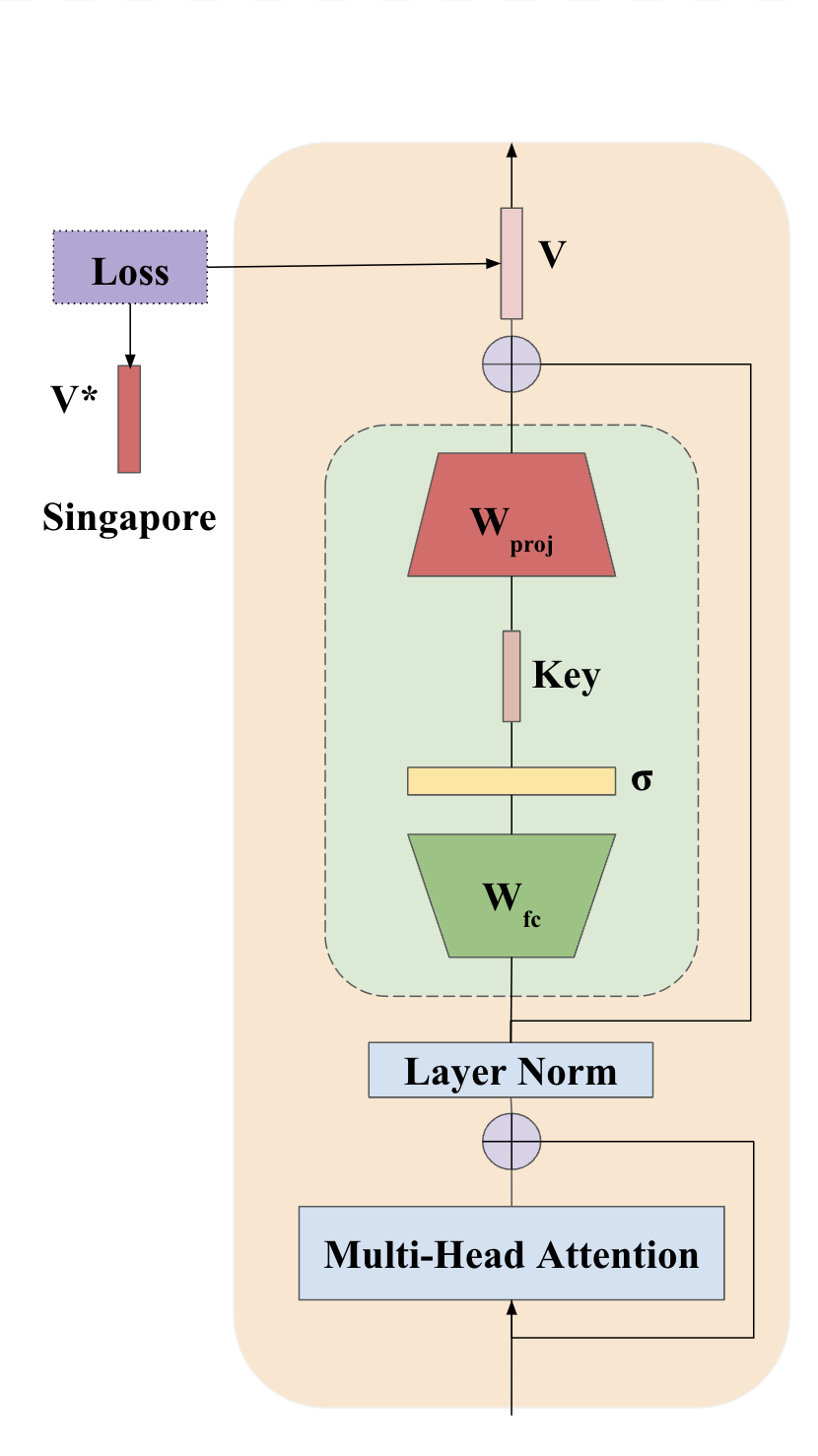} 
        \label{fig:editing-process-closed-form}
    }
    \caption{Presenting locate-then-edit knowledge editing methods as a two-step fine-tuning process.}
    \label{fig:editing-process}
    \vskip -0.1in
\end{figure}

\paragraph{Knowledge Editing Metrics:} To evaluate editing performance, we use five standard knowledge editing metrics \cite{ROME}. (i) Efficacy Score (ES), which measures if an edit was successfully made, (ii) Paraphrase Score (PS), which measures if the model is able to recall edited facts in different scenarios, (iii) Neighborhood Score (NS), which measures if edited facts disturb unrelated knowledge, (iv) Overall Score (S), which is the harmonic mean of ES, PS, and NS, and (v) Generation Entropy (GE), which measures the fluency of a model. A detailed explanation of these metrics is given in Appendix \ref{appendix:knowledge-editing-metrics}. The editing metrics for each model are calculated after making all 10,000 sequential edits \cite{alphaedit}. This approach ensures that the metrics capture both the success of the edits of the latest batch of facts as well as facts that were edited in the past.

\paragraph{Downstream Performance Metrics:} Following prior work \cite{alphaedit, akshat-catastrophic}, we measure downstream performance during knowledge editing using 6 tasks - massive multitask language understanding (MMLU) \cite{MMLU}, natural language inference (NLI, RTE) \cite{nli1, nli2, nli3}, sentiment analysis (SST2) \cite{sst2}, paraphrase detection (MRPC) \cite{mrpc}, and linguistic acceptability classification (CoLA) \cite{cola}. Performance is assessed every 1000 edits, following \citet{alphaedit}. Additional details are provided in Appendix \ref{appendix:downstream}.


\section{Knowledge Editing as a Two-Step Fine-tuning Process}\label{sec:background}

``Locate-then-edit" family of methods like ROME \cite{ROME} , MEMIT \cite{MEMIT} and AlphaEdit \cite{alphaedit} update factual knowledge expressed as (subject, relation, object) or (s,r,o). Instead of updating all parameters of a model to incorporate new knowledge, these methods only update certain matrices that are most responsible for fact recall \cite{ROME}. The location of an edit within a model is described by a two-dimensional address - (i) an intermediate layer to be edited and (ii) a token from the list of input tokens used to create the target representation.

The exact layer to be edited is found via causal tracing \cite{ROME} or an empirical sweep across the model layers \cite{localization-inform-editing, akshat-llama3}. Only the second MLP matrix in the FFN module of the editing layer is updated \citep{key-value-memories, ROME}, forming the first part of the editing address. \citet{ROME} also showed that using the output representation of the last subject token produces the best editing results, providing the second part of the address. The entire editing process is presented in Figure \ref{fig:editing-process}.

We explain this with an example. Given a fact to be edited, for example - \textit{``The capital of Malaysia is Singapore"}, the query phrase for the editing process is \textit{``The capital of Malaysia is"} and the target phrase is \textit{``Singapore"}. The first part of the editing address, the exact layer whose second MLP matrix gets edited, is decided before the editing begins. The second part of the editing address is the token index of the last subject token, which in this case would be the last subword-token in \textit{``Malasiya"}. The intermediate hidden representation of this last subject token is used to make the edit. Once the editing address has been decided, instead of updating the chosen MLP weight matrix directly using gradient-descent, locate-then-edit knowledge editing proceeds in two steps.

\begin{enumerate}
    \item In the first step (Figure \ref{fig:editing-process-gd}), gradient-descent is used to find the appropriate activation vector that acts as a target for the weight matrix to be edited. In the example, the found activation will cause the model to generate ``Singapore" in response to the question. The loss function for the gradient-descent step is shown in equation \ref{eq:mpes-loss}. Note that in this step, no weights are updated and just an intermediate activation vector is found.

    \item  
    The weight update occurs in the second editing step (Figure \ref{fig:editing-process-closed-form}), where the MLP matrix is updated using the target activation vector found previously. This update uses a least square loss function, which preserves the MLP outputs for unrelated contexts while generating the target activation when the input corresponds to the query phrase. 
    
\end{enumerate}

The loss function used to update the MLP weight matrix is formulated using least-squares in the form of a preservation-memorization objective \cite{akshat-unified}:

\vskip -0.5cm
\begingroup
\small
\begin{equation}\label{eq:memit_objective}
\begin{aligned}
     \underset{\hat{W}}{\operatorname{argmin}} \hspace{5pt} L(\hat{W}) \hspace{10pt} \text{where}& \hspace{50pt}\\ 
     L(\hat{W}) = \hspace{4pt} \underbrace{\lambda \sum^{P}_{i=1} \left\| \hat{W} k^i_0 - W_0 k^i_0 \right\|^2_2}_{\text{preservation}}  +&
     \underbrace{\sum^{B}_{j=1} \left\|\hat{W} k^j_e - v^j_e\right\|^2_2}_{\text{memorization}}
\end{aligned}
\end{equation}
\endgroup

Specifically, $W_0$ represents the initial weights of the second MLP matrix, which is being edited to $\hat{W}$. $k_0$ represent input to the MLP matrix for information we want to preserve from the original model, and $k_e$ is input activation vectors representing facts we want to insert into the model. $v_e$ is the target activation vector found in step 1 of editing using gradient-descent. Since the above objective is linear in the argument, we do not need to use gradient-descent for optimization.



Thus, locate-then-edit methods can be seen as a unique type of \textit{2-step fine-tuning method}. Instead of updating the MLP matrix directly using gradient-descent on the desired data, the weight update happens in two steps using two different types of objective functions for each step. The first step uses gradient-descent whereas the second step uses a closed-form solution. 




\begin{table*}[t]
\vspace{0.1em}
\centering
\setlength{\tabcolsep}{4pt}      
{\small                
\begin{adjustbox}{max width=\textwidth} 
\begin{tabular}{ccccc|ccc}
\toprule
\textbf{Model} &
\multicolumn{4}{c|}{\textbf{Prediction Probability}} &
\multicolumn{3}{c}{\textbf{Time / edit (s)}} \\ \cmidrule(lr){2-8}
 & \makecell{Original\\fact} &
   \makecell{MEMIT} &
   \makecell{MEMIT\\w/ LTI} &
   \makecell{MEMIT\\w/ MPES} &
   \makecell{MEMIT} &
   \makecell{MEMIT\\w/ LTI} &
   \makecell{MEMIT \\w/ MPES} \\
\midrule
Llama-2 7B & 0.52 & 0.78 & 0.30 & 0.45 & 4.84 & 8.67 & 2.79 ($\downarrow$ 42 \%) \\
Llama-3 8B & 0.49 & 0.79 & 0.29 & 0.41 & 8.71 & 9.21 & 3.31 ($\downarrow$ 61 \%) \\
\bottomrule
\end{tabular}
\end{adjustbox}
}
\caption{Comparison between prediction probabilities of edited facts along with editing time when using MPES and LTI \cite{overfitting-modelediting} with MEMIT.}\label{tab:overfitting}
\vspace{-0.1in}
\end{table*}

\section{Over-Optimization of Target Activations}\label{sec:overfitting}


The gradient-descent step for finding target activations as described in section \ref{sec:background} minimizes average cross-entropy loss for predicting the target fact for the query phrase augmented by `$N$' random prefixes. The random prefixes are supposed to represent different contexts in which the edited fact can be recalled, thus aiming to create a more general query representation \cite{ROME}. Let $p$ represent the query-phrase input to the model, $o^*$ represent the target fact to be edited into the model and $x_j$ represent random prefixes added to the query phrase. Then, the gradient-descent step minimizes the following loss function:

\vskip -0.2in 
\begin{equation}\label{eq:mpes-loss}
   L (\theta) = \frac{1}{N} \sum^{j = N}_{j = 1} -log \mathcal{P}_\theta [ o^*| x_j + p ],
\end{equation}

\begin{table*}[t]
\vskip 0.1in
\begin{center}
{\Huge
\begin{adjustbox}{max width=\textwidth}
\begin{tabular}{l cc cc cc cc cc}
\toprule
\textbf{Method} & 
\multicolumn{2}{c}{\textbf{Edit Score}} & 
\multicolumn{2}{c}{\textbf{Paraphrase Score}} & 
\multicolumn{2}{c}{\textbf{Neighborhood Score}} & 
\multicolumn{2}{c}{\textbf{Overall Score}} & 
\multicolumn{2}{c}{\textbf{Generation Entropy}} \\
\cmidrule(lr){2-3} \cmidrule(lr){4-5} \cmidrule(lr){6-7} \cmidrule(lr){8-9} \cmidrule(lr){10-11}
& Llama2-7B & Llama3-8B & Llama2-7B & Llama3-8B & Llama2-7B & Llama3-8B & Llama2-7B & Llama3-8B & Llama2-7B & Llama3-8B \\
\midrule
MEMIT                         & 81.04 & 49.68 & 64.67 & 49.29 & 60.95 & \textbf{51.31} & 67.86 & 50.08 & 442.59 & 373.48 \\
MEMIT + LTI                   & 55.34 & 51.85 & 53.57 & 51.23 & 49.21 & 49.21 & 52.58 & 50.70 & 562.07 & 372.08 \\
MEMIT + MPES                  & \textbf{93.21} & \textbf{74.40} & \textbf{83.43} & \textbf{66.32} & \textbf{62.25} & 50.78 & \textbf{77.36} & \textbf{62.23} & \textbf{569.84} & \textbf{466.97} \\
\midrule
RECT                          & 82.42 & 63.17 & 66.84 & 56.92 & \textbf{67.39} & 52.89 & 71.54 & 57.36 & 549.35 & \textbf{588.39} \\
RECT + LTI                    & 53.87 & 51.55 & 52.17 & 51.08 & 50.35 & 50.43 & 52.09 & 51.02 & 498.97 & 544.41 \\
RECT + MPES                   & \textbf{92.12} & \textbf{66.52} & \textbf{78.06} & \textbf{59.11} & 67.16 & \textbf{59.44} & \textbf{77.81} & \textbf{61.51} & \textbf{596.62} & 410.82 \\
\midrule
PRUNE                         & 70.80 & 49.38 & 62.11 & 49.63 & 51.86 & 51.09 & 60.60 & 50.02 & 280.83 & 340.22 \\
PRUNE + LTI                   & 55.19 & 49.88 & 52.98 & 49.63 & 50.33 & 50.19 & 52.76 & 49.90 & 548.23 & 239.17 \\
PRUNE + MPES                  & \textbf{91.48} & \textbf{92.66} & \textbf{83.65} & \textbf{82.20} & \textbf{61.91} & \textbf{64.82} & \textbf{76.85} & \textbf{78.16} & \textbf{568.91} & \textbf{512.83} \\
\midrule
AlphaEdit                     & 61.10 & 72.67 & 55.86 & 63.44 & 53.75 & 52.90 & 56.74 & 61.95 & 540.92 & 465.81 \\
AlphaEdit + LTI               & 53.82 & 53.27 & 53.03 & 51.70 & 49.68 & 49.72 & 52.11 & 51.52 & 524.11 & 240.07 \\
AlphaEdit + MPES              & \textbf{84.15} & \textbf{88.43} & \textbf{74.94} & \textbf{82.08} & \textbf{62.87} & \textbf{56.60} & \textbf{72.93} & \textbf{72.83} & \textbf{583.40} & \textbf{565.32} \\
\bottomrule
\end{tabular}
\end{adjustbox}
}
\end{center}
\caption{Sequential knowledge-editing performance after 10,000 edits.  
Scores are shown for nine MEMIT-based and AlphaEdit algorithms under two models.  
MPES consistently improves editing metrics across all settings.}\label{tab:editing-performance-overfitting}
\vskip -0.1in
\end{table*}

We observe that \textbf{when the gradient-descent step stops in these algorithms, the intermediate target representations become over-optimized, assigning unusually high probabilities to edited facts}. More specifically, the average target probability at which the gradient-descent step stops for all locate-then-end algorithms lies between 95-100\%. When this target representation is used to update the the edited matrix using equation \ref{eq:memit_objective}, the resultant edits are predicted with abnormally high probabilities. This can be seen in Table \ref{tab:overfitting}, where the edited facts are predicted with a much larger probability compared to the original ``unedited'' models. More details on the experimental settings can be found in Appendix \ref{appendix:overfitting}. This phenomenon was also observed by \citet{overfitting-modelediting}.



\subsection{MPES and Knowledge Editing}

As seen in Table \ref{tab:overfitting}, the probability with which the unedited Llama2-7B, and Llama3-8B naturally recall a fact are 52\% and 49\% respectively. However, facts that get edited into the model with algorithms like MEMIT are predicted with an average probability of 78-79\%. To overcome this hyper-optimization over a small subset of edited facts, we propose a variant of early stopping called ``most-probable early stopping" (MPES). 

Conventionally, early stopping is used during training while monitoring validation loss, where training is halted when the validation loss stops improving. In MPES, we stop the gradient-descent step in knowledge editing when the target fact becomes the most probable token for all `$N$' query phrases used for optimization (equation \ref{eq:mpes-loss}).  This contrasts with the current stopping criteria in these methods, where gradient-descent is halted after either 20 iterations or upon reaching a loss threshold. These criteria leave open the possibility of over-optimization (and hence overfitting, as observed empirically in Table \ref{tab:overfitting}), or underfitting. Apart from preventing the model from becoming overly optimized towards a specific target fact, using MPES for halting gradient-descent also has two other advantages. \textit{Firstly, it simplifies monitoring of the gradient-descent process and provides a principled approach to stopping gradient-descent which is directly tied to the knowledge editing objective, that is, accurately recalling edited facts in a variety of scenarios without overfitting. Secondly, by optimally stopping the gradient-descent process, we also improve the efficiency of locate-then-edit algorithms. MPES reduces the average edit time by 42-61\% compared to the standard MEMIT.}



\begin{figure*}[ht]
    \subfigure[Llama2-7B]{
        \includegraphics[width=0.49\linewidth]{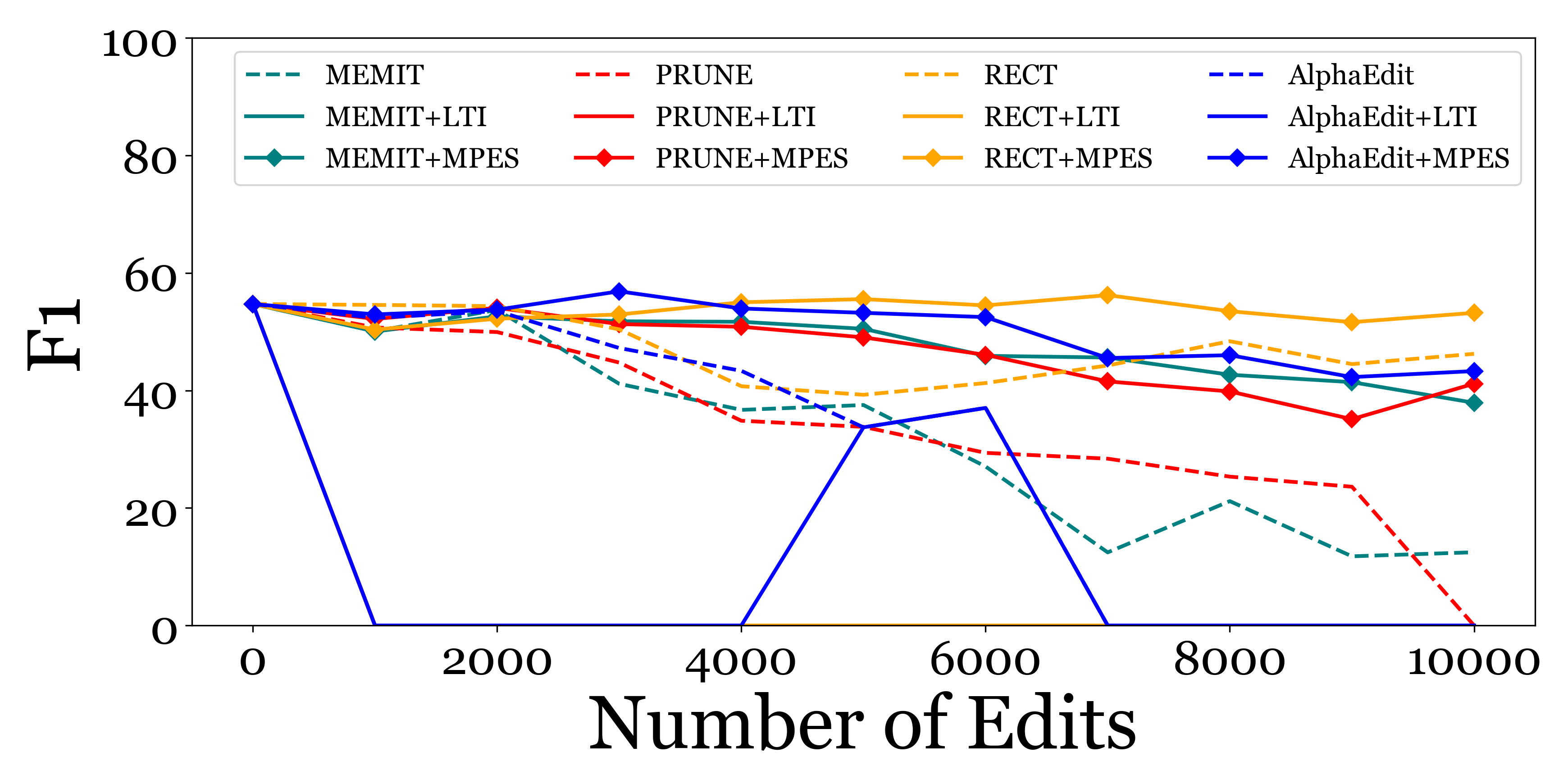} 
        \label{fig:downstream-mpes-memit}
    }
    \subfigure[Llama3-8B]{
        \includegraphics[width=0.49\linewidth]{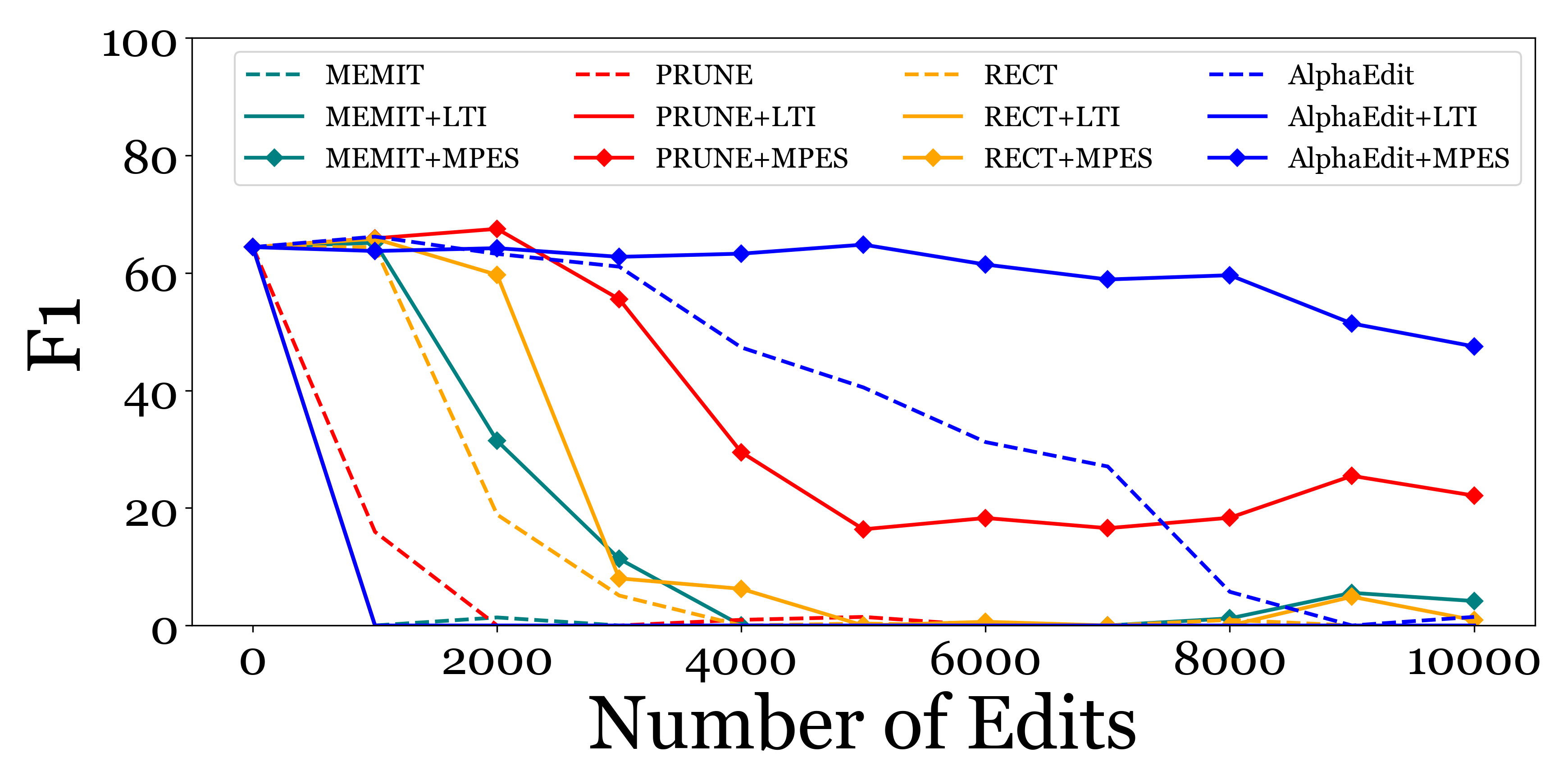} 
        \label{fig:downstream-mpes-lti-memit}
    }

    \caption{Downstream evaluation when comparing MPES (our method) with LTI for regularizing over-optimization of target activations during knowledge editing.  }\label{fig:downstream-baseline-main}
\end{figure*}

We compare MPES with LTI \cite{overfitting-modelediting}, which prevents over-optimization during the gradient-descent step using multiple additional loss functions that also increase the average time per edit (LTI takes approximately three times as long as MPES as shown in Table \ref{tab:overfitting}). We combine MPES with multiple knowledge editing methods including MEMIT \cite{MEMIT}, RECT \cite{rect}, PRUNE \cite{prune} and AlphaEdit \cite{alphaedit}. We perform 10,000 sequential edits in batches of 100 as described in section \ref{sec:evaluation}. The editing and downstream evaluation metrics for sequential editing for both regularization methods can be seen in Table \ref{tab:editing-performance-overfitting} and Figure \ref{fig:downstream-baseline-main} respectively. We see that MPES provides substantial gain across all key metrics. 

Table \ref{tab:editing-performance-overfitting} shows that all editing metrics are enhance significantly for all knowledge editing methods when combined with MPES. Infact,  Table \ref{tab:editing-performance-overfitting} shows that LTI harms the editing metrics for sequential editing when compared to even the respective baselines. The downstream performance metrics can be seen in Figure \ref{fig:downstream-baseline-main}. We see that knowledge editing methods augmented with MPES (represented by solid lines with diamond dots) are able to maintain their downstream performance for much longer when compared to the corresponding baselines and LTI. However, despite these improvements, downstream degradation still occurs after a few thousand edits especially in larger models like Llama3-8B. This suggests that MPES alone may not fully address all failure modes, which we explore further in the next section. To summarize, \textbf{with MPES we present a principled way of stopping the gradient-descent step during knowledge editing} which results in improved editing performance, delays loss of downstream performance, and makes current knowledge editing methods much faster.

\vskip -0.2in 
\section{Norm Growth during Sequential Knowledge Editing}
Previous studies show that sequential knowledge editing increases the norm of the edited matrix \cite{akshat-catastrophic, disabling-butterfly, prune}. Figure \ref{fig:norm-growth-subset} illustrates this issue, highlighting the extreme norm growth in edited layers of Llama3-8B using MEMIT. Here, the norms of the edited layers increase by more than 10 times, while the unedited layer remains unchanged (Figure \ref{fig:norm-growth-MEMIT-layer-wise}). Furthermore, norm growth persists continuously during editing, as shown in Figure \ref{fig:norm-growth-subset}, for not even one edit does the norm remain constant or decrease.

While the anomalous norm-growth was observed in prior work, they do not explain how it affects the general ability of the model. We answer this question by analyzing the residual stream of the model during large-scale knowledge editing.




\vskip -0.1in 
\subsection{Explaining Loss of Downstream Performance due to Norm-Growth}\label{sec:secret-mechanics}

To understand the impact of this norm growth, we analyze how residual connections work in decoder-only LLMs. The intermediate hidden vector at layer $l$, represented by $h^l$, is also sometimes referred to as the \textit{residual stream}. Each decoder-only layer contains one attention and FFN module that feeds directly into the residual stream through residual connections. The exact computations happening within transformer-based decoder-only LLMs can be found in Appendix \ref{appendix:internal-computations}. Let the output of the attention module at layer $l$ be represented by $a^l$, and the output of the FFN module be represented by $m^l$. As the vectors computed in the attention and MLP modules get added back to the residual stream at each layer, the residual stream represents a summation of an increasing number of vectors as we go deeper into the model. A non-recursive formula for the output of the transformer just before unembedding and final layernorm is shown below:

\begin{equation}\label{eq:resiudal-summation}
    h^L = h^{0} + \sum^{i = L}_{i = 1} a^i + \sum^{i = L}_{i = 1} m^i
\end{equation}

Here, $L$ represents the total number of layers in a model and $h^L$ represents the residual stream after the final layer. Thus, the output vector at the final layer is a summation of the outputs of individual self-attention and MLP sub-modules. 

Now, if the norm of the edited MLP matrix grows as disproportionately as shown in Figure \ref{fig:norm-growth-subset}, the norm of the vectors that are produced from those edited MLP sub-modules will also grow. This means that the norms of the vectors $m^l$ in the summation corresponding to the edited layers will grow substantially. As the norm of a few vectors in the summation grows, these vectors will begin to dominate the sum. Proof for this is shown in Appendix \ref{appendix:activation-norm-growth-proof}, where we show that if the norm of a vector in a summation grows, the overall sum effectively tends towards that vector.

\begin{figure}[ht]
    \centering

    \includegraphics[width=\linewidth]{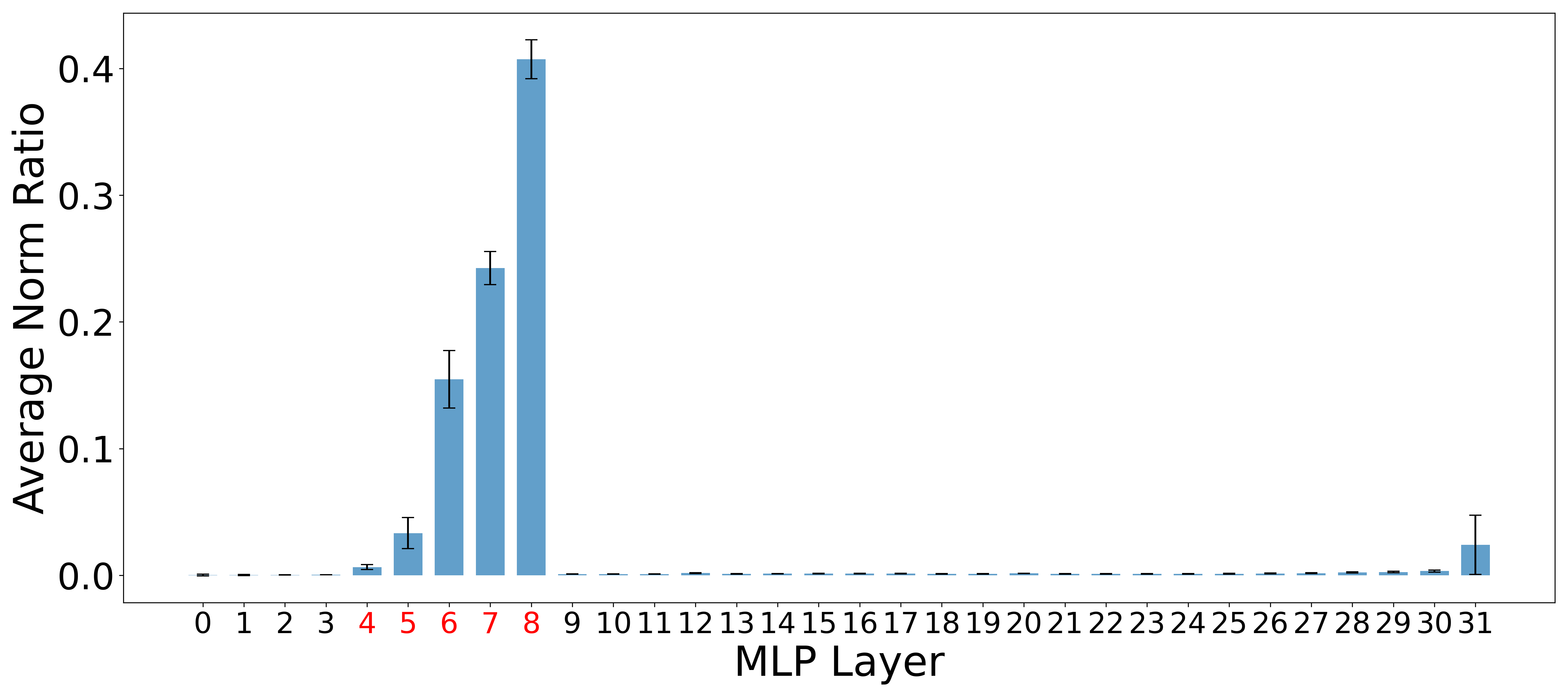}
    \label{fig:activation-norm-growth-llama3-memit}
    \vskip -0.2 in
    \caption{The figure shows the proportion of contribution of activation vectors from each sub-module to the residual stream. Edited layers are shown in red.}
    \label{fig:activation-norm-growth}
\end{figure}

We also show this effect empirically. The growing norm of activation vectors produced by edited layers after editing Llama3-8B can be seen in Figure \ref{fig:activation-norm-growth}. After editing using MEMIT, which edits layers 4-8 for 10,000 sequential edits (edited layers are shown in red color on the x-axis), we see a drastic increase in the norm of activation vectors produced by edited layers. For example, the activation vectors produced by layer 8 account for almost 40\% of the total norm, and vectors produced by all edited layers account for 85\% of the total. \textit{To emphasize how extreme this is, the residual stream for Llama-3-8B is made up of a summation of 65 vectors\footnote{Llama-3-8B has 32 layers, each contributing two vectros $a^l$ and $m^l$ to the residual stream along with the input vector.}, and 4 out of the 65 vectors coming from the edited layers account for 85\% of the total norm.} The norm-contributions for an unedited model can be seen in Figure \ref{fig:activation-norm-growth-llama3-baseline-appendix} (appendix), where the four edited layers account for less than 4\% of the overall norm. Note that this effect cannot be mitigated by LayerNorm or RMSNorm, which just normalize the incoming vectors \cite{brody2023expressivity}, whereas the norm growth of edited layers changes the content of the final hidden representation by making it consist of mostly the output of the edited layers.



This gives us a crucial insight into why the growing norm of edited matrices leads to a loss of downstream performance. \textbf{As the norms of the edited matrices increase, and as a consequence, the norm of the activation vectors produced from those matrices also increases, the activation vector at the final layer ($h^L$) is dominated by the output of edited layers, thus giving the edited layers a larger authority over the final representations. This allows edited layers to override the information produced from other parts of the model}, possibly also helping make successful edits. However, as the norm of the edited matrix continues to grow when we edit it sequentially, suddenly the final representations begin to largely be composed of the outputs of only the edited layers. This makes it impossible for the model to account for the information processed from other sub-modules, which might be important to perform other downstream or unrelated tasks.

\vskip -0.3in 
\subsection{Introducing Norm Constraint}\label{sec:norm-constraint}

In the above discussion, we show how growing norm of edited matrices is detrimental to the functioning of edited models and hypothesize how it causes loss of downstream performance. To test these conclusions, we propose to add an additional term to the editing objective to control this norm growth. Thus, we augment the MEMIT objective with a Frobenius norm-constraint:

\begingroup
\small
\begin{equation}\label{eq:encore_objective}
\begin{aligned}
     L(\hat{W}) = \hspace{4pt} &\underbrace{\lambda_p \sum^{P}_{i=1} \left\| \hat{W} k^i_0 - W_0 k^i_0 \right\|^2_2}_{\text{preservation}}  +
     \underbrace{\sum^{B}_{j=1} \left\|\hat{W} k^j_e - v^j_e\right\|^2_2}_{\text{memorization}} \\
     & \hspace{50pt} + \underbrace{\lambda_n \left\|\hat{W}- W_0\right\|^2_F}_{\text{norm-constraint}}
\end{aligned}
\end{equation}
\endgroup

The original weight matrix is represented by $W_0$ and $\hat{W}$ represents the edited matrix. The above objective has a closed-form solution as shown below (proof, Appendix \ref{appendix:encore-proof}):
\begingroup
\small
\begin{equation}
\begin{aligned}
     \hat{W} &= W_0 + \Delta \hspace{10pt}\text{where} \\
     \Delta = (V_{1} - W_{0}K_{1})&K_{1}^{T} (\lambda_p K_{0}K_{0}^{T} + K_{1}K_{1}^{T} + \lambda_n I)^{-1}
\end{aligned}
\end{equation}
\endgroup

\begin{figure*}[ht]
    \centering
    \subfigure[Llama2-7B]{
        \includegraphics[width=0.48\linewidth]{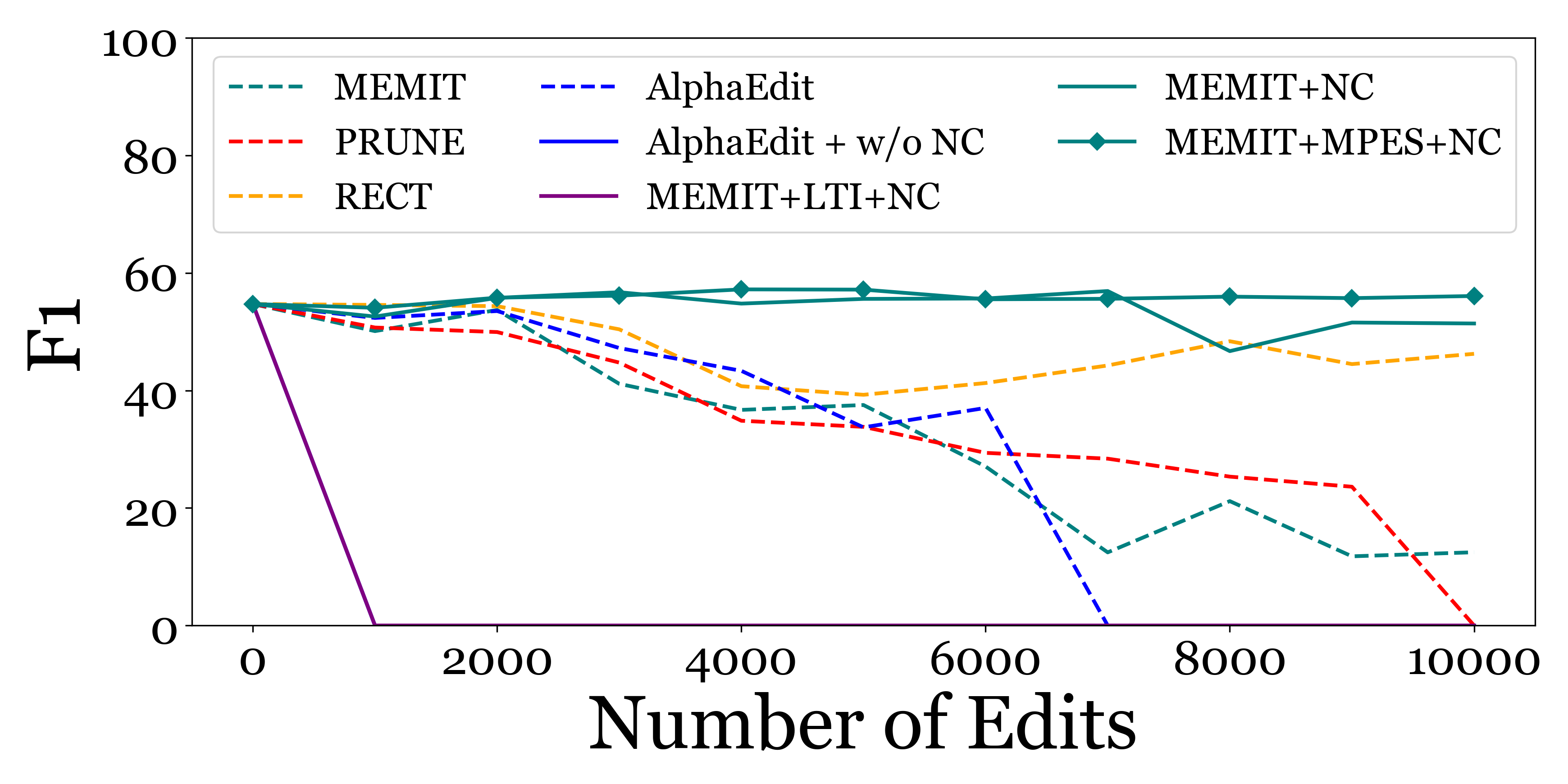} 
        \label{fig:downstream-encore-main-llama2}
    }
    \subfigure[Llama3-8B]{
        \includegraphics[width=0.48\linewidth]{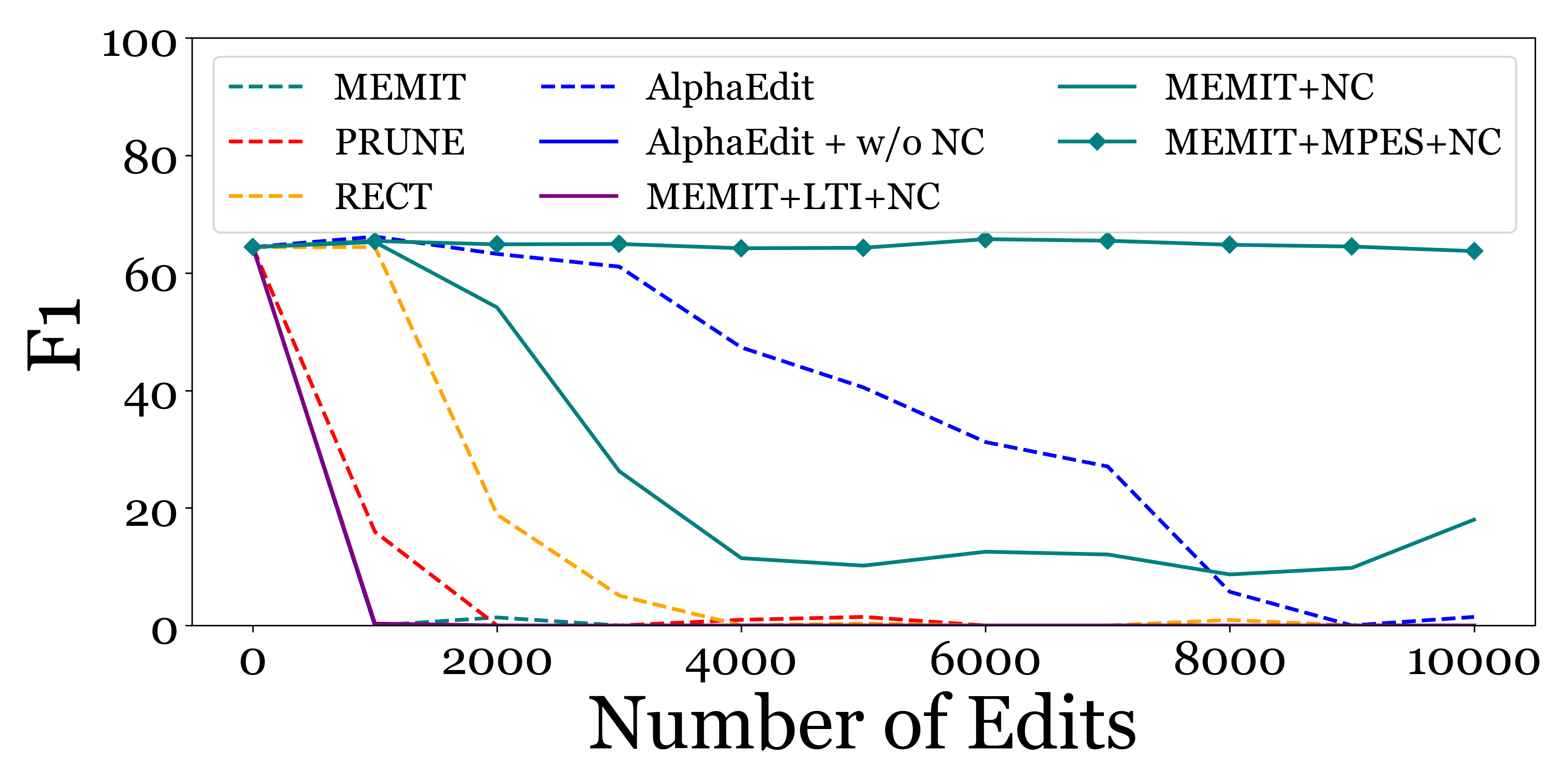} 
        \label{fig:downstream-encore-main-llama3}
    }
    
    \caption{Average downstream performance for during sequential editing with compared to baseline of MEMIT and addition of MPES and Norm-Constraint (NC).}\label{fig:downstream-norm-constraint}
\end{figure*}

Table \ref{tab:editing-performance-encore-main} and Figure \ref{fig:downstream-norm-constraint} show the editing and downstream performance after 10,000 sequential edits using the norm-constraint. Compared to baseline methods such as MEMIT, AlphaEdit, as well as regularization approaches such as PRUNE and RECT, Frobenius norm-constraint version of MEMIT (represented by MEMIT + NC) produces consistent and substantial improvements in editing and downstream performance metrics. Specifically, we see in Table \ref{tab:editing-performance-encore-main} that MEMIT + NC outperforms all previous models on all editing metrics except for the generation entropy metric. This shows that explicitly controlling norm growth is more effective than prior regularization methods like PRUNE and RECT, particularly when editing at scale. When looking at downstream performance in Figure \ref{fig:downstream-norm-constraint}, MEMIT + NC also outperforms all prior editing methods when looking for Llama-2-7B, and all methods except AlphaEdit for Llama3-8B. Moreover, the observed gains in downstream performance support our hypothesis that uncontrolled norm growth in edited layers leads to downstream degradation. 

\begin{table*}[t]
\vskip 0.1in
\begin{center}
{\Huge
\setlength{\tabcolsep}{4pt}
\begin{adjustbox}{max width=\textwidth}
\begin{tabular}{l cc cc cc cc cc}
\toprule
\textbf{Method} & 
\multicolumn{2}{c}{\textbf{Edit Score}} & 
\multicolumn{2}{c}{\textbf{Paraphrase Score}} & 
\multicolumn{2}{c}{\textbf{Neighborhood Score}} & 
\multicolumn{2}{c}{\textbf{Overall Score}} & 
\multicolumn{2}{c}{\textbf{Generation Entropy}} \\
\cmidrule(lr){2-3} \cmidrule(lr){4-5} \cmidrule(lr){6-7} \cmidrule(lr){8-9} \cmidrule(lr){10-11}
& Llama2-7B & Llama3-8B & Llama2-7B & Llama3-8B & Llama2-7B & Llama3-8B & Llama2-7B & Llama3-8B & Llama2-7B & Llama3-8B \\
\midrule
MEMIT                         & 81.04 & 49.68 & 64.67 & 49.29 & 60.95 & 51.31 & 67.86 & 50.08 & 442.59 & 373.48 \\
PRUNE                         & 70.80 & 49.38 & 62.11 & 49.63 & 51.86 & 51.09 & 60.60 & 50.02 & 280.83 & 340.22 \\
RECT                          & 82.42 & 63.17 & 66.84 & 56.92 & \textbf{67.39} & 52.89 & 71.54 & 57.36 & \textbf{549.35} & \textbf{588.39} \\
AlphaEdit                     & 61.10 & 72.67 & 55.86 & 63.44 & 53.75 & 52.90 & 56.74 & 61.95 & 540.92 & 465.81 \\
AlphaEdit w/o NC & 0.00 & 0.00 & 0.00 & 0.00 & 0.00 & 0.00 & 0.00 & 0.00 & 434.17 & 560.92 \\
MEMIT + LTI + NC      & 54.41 & 50.76 & 51.87 & 48.89 & 52.05 & 51.17 & 52.75 & 50.25 & 516.24 & 575.59 \\
MEMIT + NC            & 87.89 & 85.72 & \textbf{79.10} & 77.08 & 59.72 & 58.45 & 73.59 & 71.86 & 517.86 & 367.46 \\
MEMIT + MPES + NC     & \textbf{91.09} & \textbf{87.54} & 78.02 & \textbf{77.97} & 59.78 & \textbf{59.30} & \textbf{74.03} & \textbf{72.97} & 545.72 & 536.43 \\
\bottomrule
\end{tabular}
\end{adjustbox}
}
\end{center}
\caption{Editing performance of our approach when compared to baseline MEMIT, AlphaEdit and MEMIT regularization method such as PRUNE and RECT. }\label{tab:editing-performance-encore-main}
\vskip -0.1in
\end{table*}

Additionally, we want to note that AlphaEdit inherently has an explicit norm-constraint in its optimization objective. To present the importance of norm-constraint during large scale editing, we remove this term from the objective and evaluate the method, as shown in Table \ref{tab:editing-performance-encore-main}. To our surprise, we find that AlphaEdit completely fails without the norm constraint, as shown in the row \textit{`AlphaEdit w/o NC'} in Table \ref{tab:editing-performance-encore-main} . Specifically, after 10,000 edits, the model effectively collapes, where Llama2-7B outputs \texttt{<unk>} and Llama3-8B loops on "!" and all editing scores drop to \underline{zero}. We also illustrate the unbound growth of norm in Figures \ref{fig:alphaedit-llama2-without-norm} and \ref{fig:alphaedit-llama3-without-norm} in Appendix \ref{appendix:unbound-growth}.  These results support our analysis in section \ref{sec:secret-mechanics} and re-emphasize that managing norm-growth is essential for stable and successful large-scale editing.





\subsection{Combining MPES and Norm Constraint}


In this section, we combine MPES, which intervenes in the gradient-descent step, with explicit Frobenius norm constraint, which regularizes the weight-update step in MEMIT. This combined approach, MEMIT + MPES + NC, achieves the strongest results across both editing score and downstream metrics. As shown in Table \ref{tab:editing-performance-encore-main}, it consistently improves the consistency of editing scores between models, with the largest gains observed on Llama3-8B. More importantly, Figures \ref{fig:downstream-encore-main-llama2} and \ref{fig:downstream-encore-main-llama3} show that the combination of the two regularization methods beautifully preserves downstream task performance with great effectiveness when compared to other algorithms, even after 10,000 sequential edits.

These findings suggest that MPES and norm-constraint address complementary failure modes during large-scale sequential editing. Their combination provides regularization methods that enable scalable knowledge editing while having minimal loss of downstream task performance. The generalizability of these findings can be seen by the compatibility of MPES with other knowledge editing methods, and the neccesity and importance of managing norm-growth in both MEMIT and AlphaEdit.

\section{Conclusion}
In this paper, we show that existing knowledge editing methods require appropriate regularization when scaled sequentially to a large number of edits.
Specifically, we diagnose and address two core failure modes in large-scale knowledge editing, (1) over-optimization of intermediate activations and (2) uncontrolled norm growth of edited matrices. To mitigate these, we introduce stage-specific regularization techniques in the form of ``most-probable earlys stopping'' (MPES) and Frobenius Norm-Constraint, which generalize to multiple knowledge editing methods. With the combination of both these regularization methods, we are able to make 10,000 sequential edits while consistently maintaining the downstream performance levels of the unedited model.


\section{Limitation}

Although our method significantly improves editing speed, scaling beyond 10,000 edits still remains challenging due to compute constraints. Performing 10,000 sequential edits on Llama3-8B already required a full day of GPU time on our hardware. This is because running these edits is combined with a lot of analysis, including downstream performance evaluation and measuring editing accuracy metrics on past edited facts. In contrast, baseline methods such as MEMIT combined with LTI were substantially slower, making larger-scale comparisons infeasible. Future work could explore more efficient optimization techniques or hardware acceleration to push beyond this limit.



\section*{Acknowledgements}
This work was supported in part by the NVIDIA Academic Grant Program award.

\bibliography{custom}

\appendix

\section{Knowledge Editing Metrics}\label{appendix:knowledge-editing-metrics}
A more detailed explanation of the knowledge editing metrics used in this paper is below:

\begin{enumerate}
    \item \textbf{Efficacy Score (ES)}: assesses whether an edit has been successful. It is calculated as the percentage of edits where $P(\text{new fact}) > P(\text{old fact})$ when evaluated on paraphrases of the query prompt.   
    \item \textbf{Paraphrase Score (PS)}: measures the model's ability to generalize after an edit. Specifically, it is the percentage of edits where $P(\text{new fact}) > P(\text{old fact})$ for paraphrased versions of the query prompt.
    \item \textbf{Neighborhood Score (NS)}: evaluates the locality of a model edit by determining whether editing one fact affects other facts stored in the model. It is the percentage of unaffected facts in the neighborhood of the edited fact.
    \item \textbf{Generation Entropy (GE)}: measures the fluency of the model's text generation post-edit. GE is computed as the weighted average of bi-gram and tri-gram entropies in the text generated by the edited model. A lower GE indicates repetitive text generation, a common failure mode \cite{ROME, akshat-catastrophic}. 
    \item \textbf{Score (S)}: introduced by \cite{ROME}, this composite metric combines edit success, generalization, and locality into a single score. It is calculated as the harmonic mean of the Efficacy Score (ES), Paraphrase Score (PS), and Neighborhood Score (NS).
\end{enumerate}

\section{Experimental Detail on Overfitting During Knowledge Editing}\label{appendix:overfitting}

For each method and model, we conducted three different experiments: 

\begin{enumerate}

    \item Unedited fact recall probability - In this case, we calculate the average probability with which a fact is recalled by the unedited/original model. These are the facts that the model learnt through its pre-training. The model is asked questions from the CounterFact dataset, and we average the probability with which the model predicts the fact correctly. 
    
    \item Edited fact probability without MPES - In this case, we evaluate the probability with which a model recalls a fact that is edited into the model. This is the standard baseline case without MPES.

    \item Edited fact probability with LTI - Here we also evaluate the average probability with which a model recalls an edited fact. In this case we used LTI during editing the fact.
    
    \item Edited fact probability WITH MPES - Here we also evaluate the average probability by which a model recalls an edited fact. In this case, MPES is used during editing the fact.
\end{enumerate}

In each of the experiments we performed, we passed average numbers over 1000 edited facts with a batch size of 1. We use the CounterFact dataset \cite{ROME} for all of these experiments. We show the result in table \ref{tab:overfitting}. As we can see from the table, the unedited fact token probability is pretty low but once we run the edited fact the probability increases to almost 1 for some cases. MPES brings the probability of fact recall for edited facts down to a more natural value, which prevents the overfitting problem that we present in this paper.

\section{GPT2-XL Result}
\begin{figure}[ht]
    \centering
    \subfigure[GPT2-XL downstream performance comparison with MPES]{
        \includegraphics[width=\linewidth]{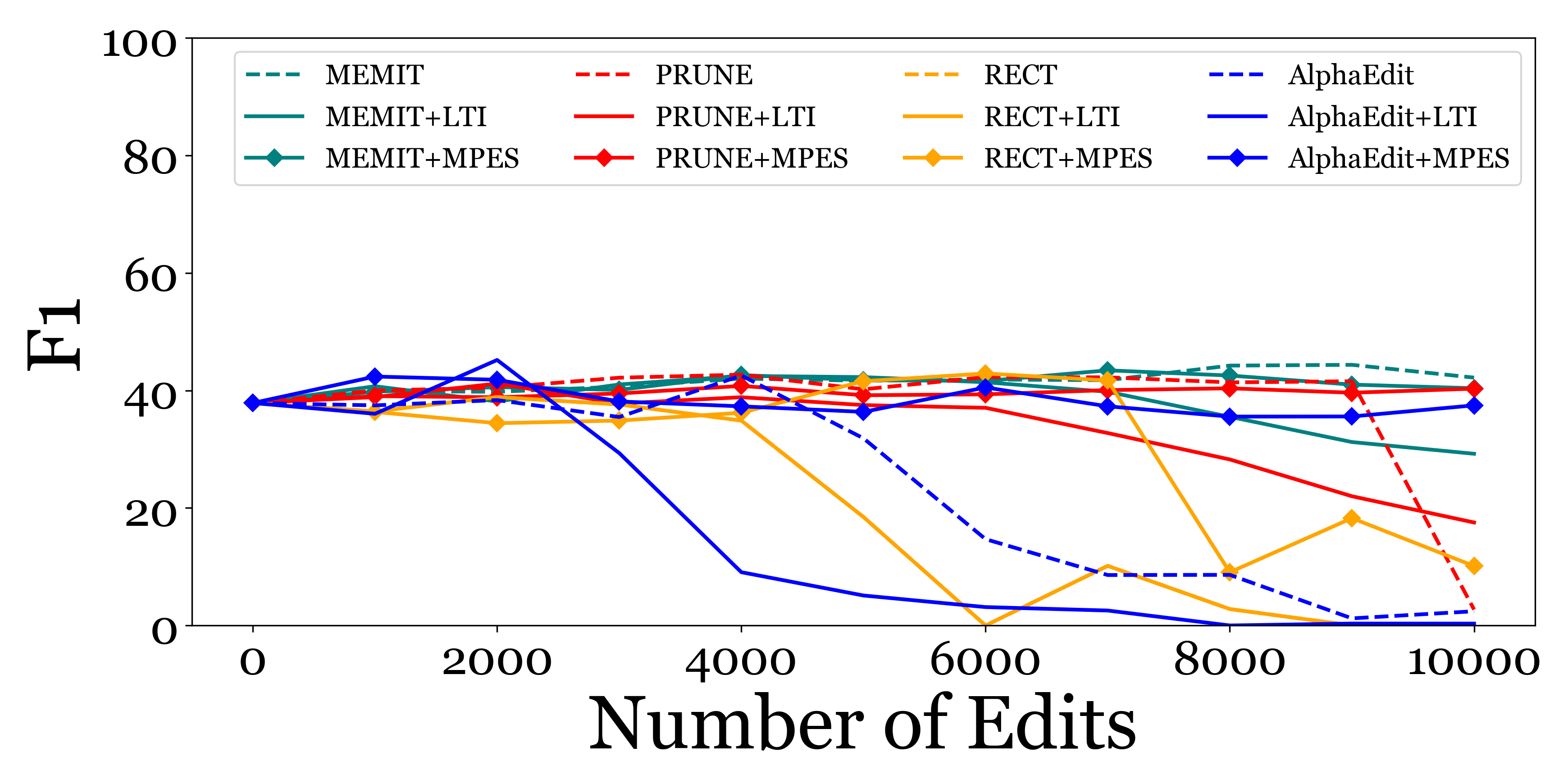} 
        \label{fig:downstream-overfit-gpt2xl}
    }
    \subfigure[GPT2-XL downstream performance comparison with MPES + NC]{
        \includegraphics[width=\linewidth]{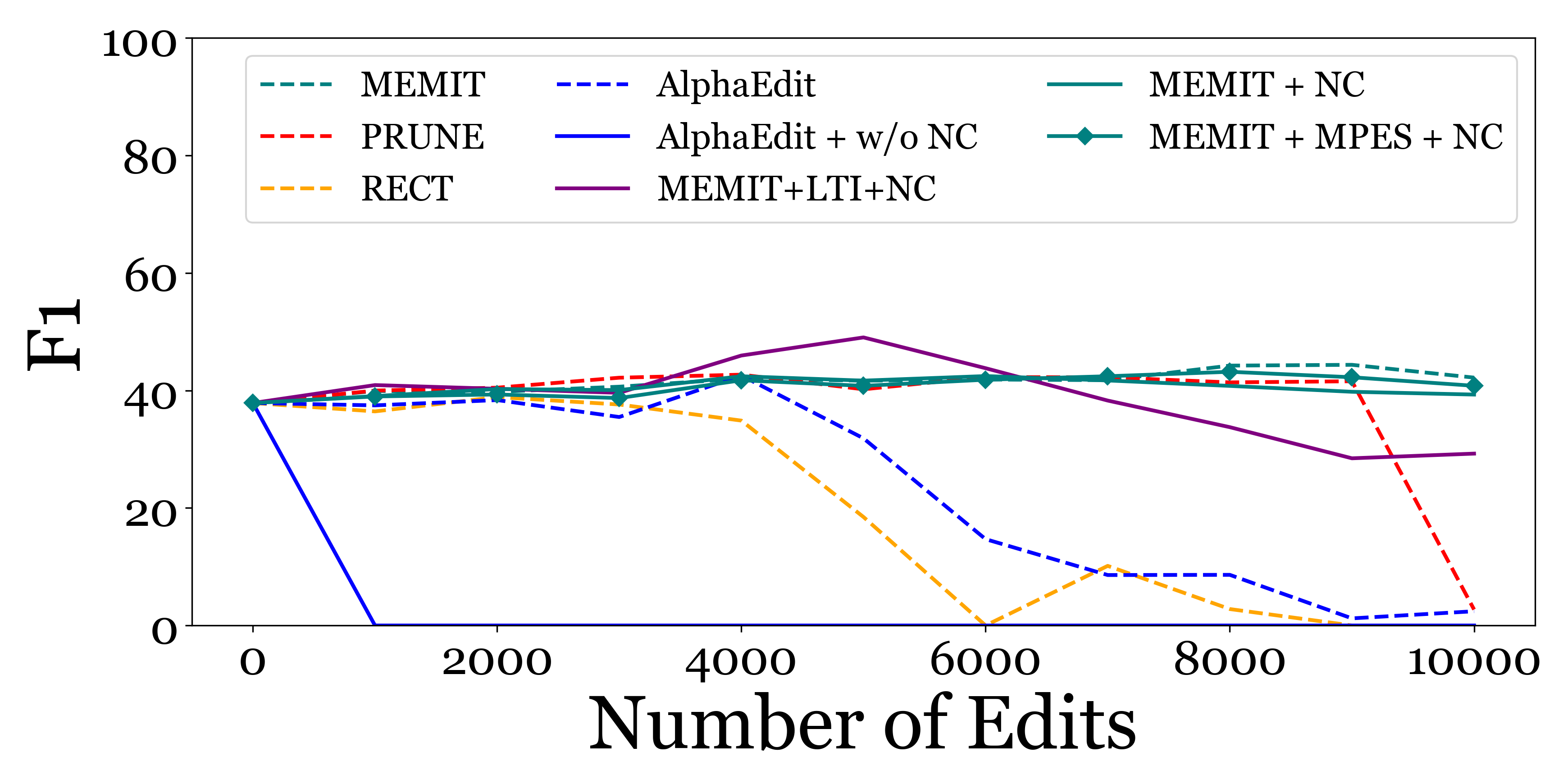} 
        \label{fig:downstream-encore-gpt2xl}
    }
    
    \caption{Average downstream performance for during sequential editing with compared to baseline of MEMIT and addition of MPES and Norm-Constraint (NC).}\label{fig:downstream-norm-constraint-gpt2xl}
\end{figure}

In this section we show the result for GPT2-XL on CounterFact dataset in Tables \ref{tab:editing-performance-overfitting-gpt2xl}
and \ref{tab:editing-performance-gpt2xl} and also the downstream average downstream performance in Figure \ref{fig:downstream-norm-constraint-gpt2xl}. As we can see that the result for GPT2-XL also improves the results as well.

\begin{table}[H] 
\vskip 0.1in
\begin{center}
\begin{adjustbox}{max width=0.5\textwidth}
\begin{sc}
\begin{tabular}{lccccccr}
\toprule
Method  & \multirow{2}{*}{\makecell{Edit \\ Score}} & \multirow{2}{*}{\makecell{Paraphrase \\ Score}} & \multirow{2}{*}{\makecell{Neighborhood \\ Score}} & \multirow{2}{*}{\makecell{Overall \\ Score}} & \multirow{2}{*}{\makecell{Generation \\ Entropy}} \\
& & & & & & \\
\midrule
MEMIT    & \textbf{94.04} & \textbf{79.91} & 57.90 & \textbf{74.22} & 517.37  \\
MEMIT + LTI & 82.03 & 70.51 & 56.29 & 67.97 & 508.35 \\
MEMIT + MPES   & 92.17 & 77.01 & \textbf{60.38} & 72.26 & \textbf{523.57}  \\
\midrule
PRUNE & 61.05 & 58.05 & 50.00 & 55.96 & \textbf{579.69} \\
PRUNE + LTI & \textbf{80.06} & \textbf{69.77} & 56.07 & \textbf{67.18} & 542.41\\
PRUNE + MPES & 76.91 & 59.11 & \textbf{65.96} & 66.54 & 563.57\\
\midrule
RECT & 51.40 & 49.83 & 52.17 & 51.12 & 409.42\\
RECT + LTI & 47.51 & 47.37 & 52.74 & 49.08 & 185.96 \\
RECT + MPES & \textbf{51.99} & \textbf{49.93} & \textbf{54.29} & \textbf{52.01} & \textbf{523.63}  \\
\midrule
AlphaEdit    & 88.58 & 70.33 & 56.04 & 69.20 & 580.27  \\
AlphaEdit + LTI & 74.20 & 59.90 & 54.86 & 61.98 & \textbf{596.47} \\
AlphaEdit + MPES    & \textbf{95.52} & \textbf{82.08} & \textbf{60.03} & \textbf{76.32} & 565.44  \\
\bottomrule
\end{tabular}
\end{sc}
\end{adjustbox}
\end{center}
\caption{Knowledge editing performance for GPT2-XL on the CounterFact dataset for different algorithms in combination with MPES.}
\label{tab:editing-performance-overfitting-gpt2xl}
\vskip -0.1in
\end{table}

\begin{table}[H]
\vskip 0.15in
\begin{center}
\begin{adjustbox}{max width=0.5\textwidth}
\begin{sc}
\begin{tabular}{lccccccr}
\toprule
Method  & \multirow{2}{*}{\makecell{Edit \\ Score}} & \multirow{2}{*}{\makecell{Paraphrase \\ Score}} & \multirow{2}{*}{\makecell{Neighborhood \\ Score}} & \multirow{2}{*}{\makecell{Overall \\ Score}} & \multirow{2}{*}{\makecell{Generation \\ Entropy}} \\
& & & & & & \\
\midrule
MEMIT    & \textbf{94.04} & \textbf{79.91} & 57.90 & 74.22 & 517.37  \\
PRUNE & 61.05 & 58.05 & 50.00 & 55.96 & 579.69 \\
RECT & 51.40 & 49.83 & 52.17 & 51.12 & 409.42\\
AlphaEdit    & 88.58 & 70.33 & 56.04 & 69.20 & \textbf{580.27}  \\
AlphaEdit + w/o NC & 47.37 & 47.37 & 52.63 & 49.00 & 573.40 \\
MEMIT + LTI + NC  & 84.15 & 73.12 & 54.91 & 68.54 & 554.47 \\
MEMIT + NC  & 93.89 & 80.9 & 58.00 & 74.53 & 504.68 \\ 
MEMIT + MPES + NC & 93.99 & 79.79 & \textbf{59.98} & \textbf{75.29}& 517.44 \\ 
\bottomrule
\end{tabular}
\end{sc}
\end{adjustbox}
\end{center}
\caption{Knowledge editing performance for GPT2-XL on the CounterFact dataset for different algorithms in combination with MPES and NC.}
\label{tab:editing-performance-gpt2xl}
\vskip -0.1in
\end{table}

\section{Experimental Details on Activation Norm Growth}\label{appendix:activation-norm-growth}
To assess the impact of the activation vectors generated by the edited layers before and after editing, we conducted an experiment where we edited our model on a total of 10,000 facts using a batch size of 100. Once the model was edited, we evaluated the norm of the activations produced by each layer by passing it through a Wikipedia dataset containing 30,000 examples. For each example, the model performed a one-word prediction task given an input context, and we measured the norm of the activation vectors produced from each layer in the model. We repeated the same process for the unedited model to compare the differences. 

\begin{figure}[htbp]
    \centering
    \subfigure[Average Norm Proportion For Unedited GPT2-XL]{
        \includegraphics[width=\linewidth]{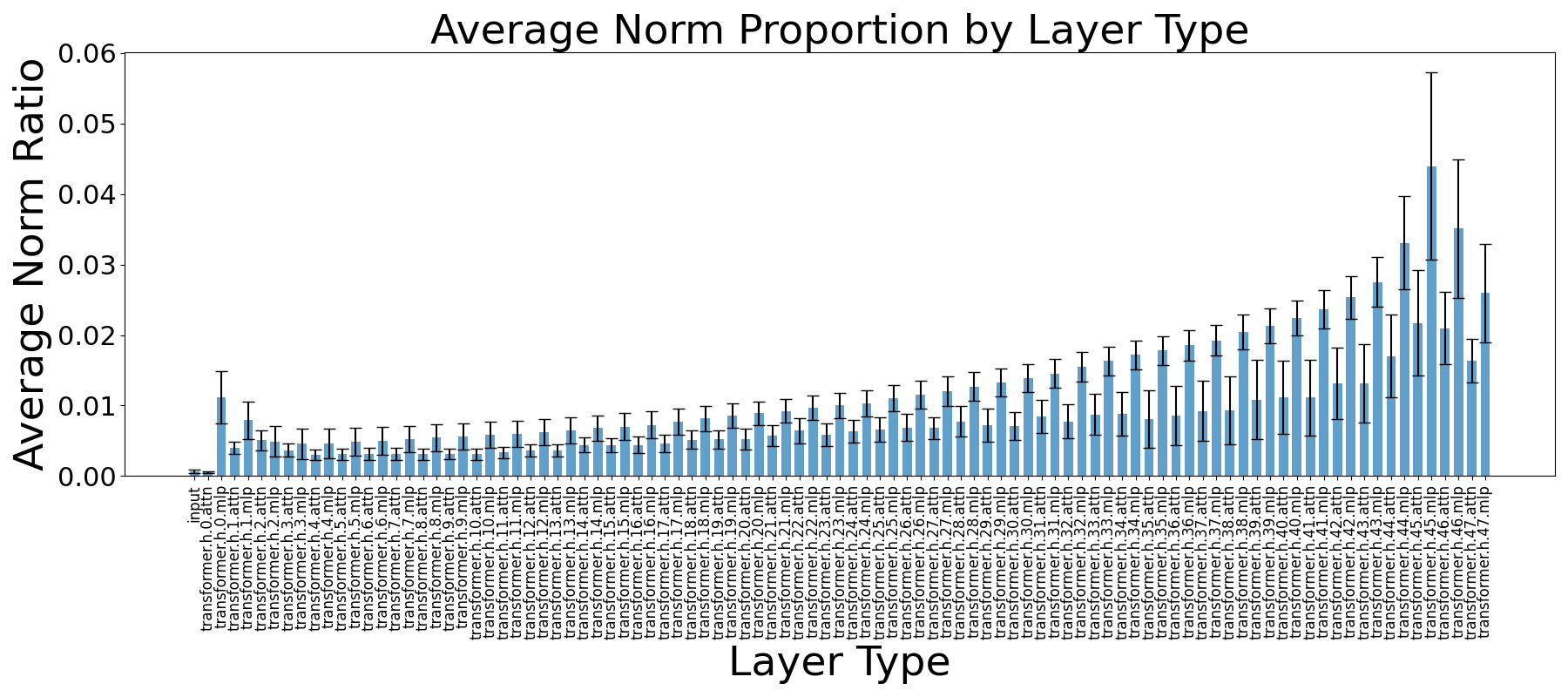} 
        \label{fig:activation-norm-growth-gpt2-xl-baseline}
    }
    \subfigure[Average Norm Proportion for GPT2-XL using Alphaedit]{
        \includegraphics[width=\linewidth]{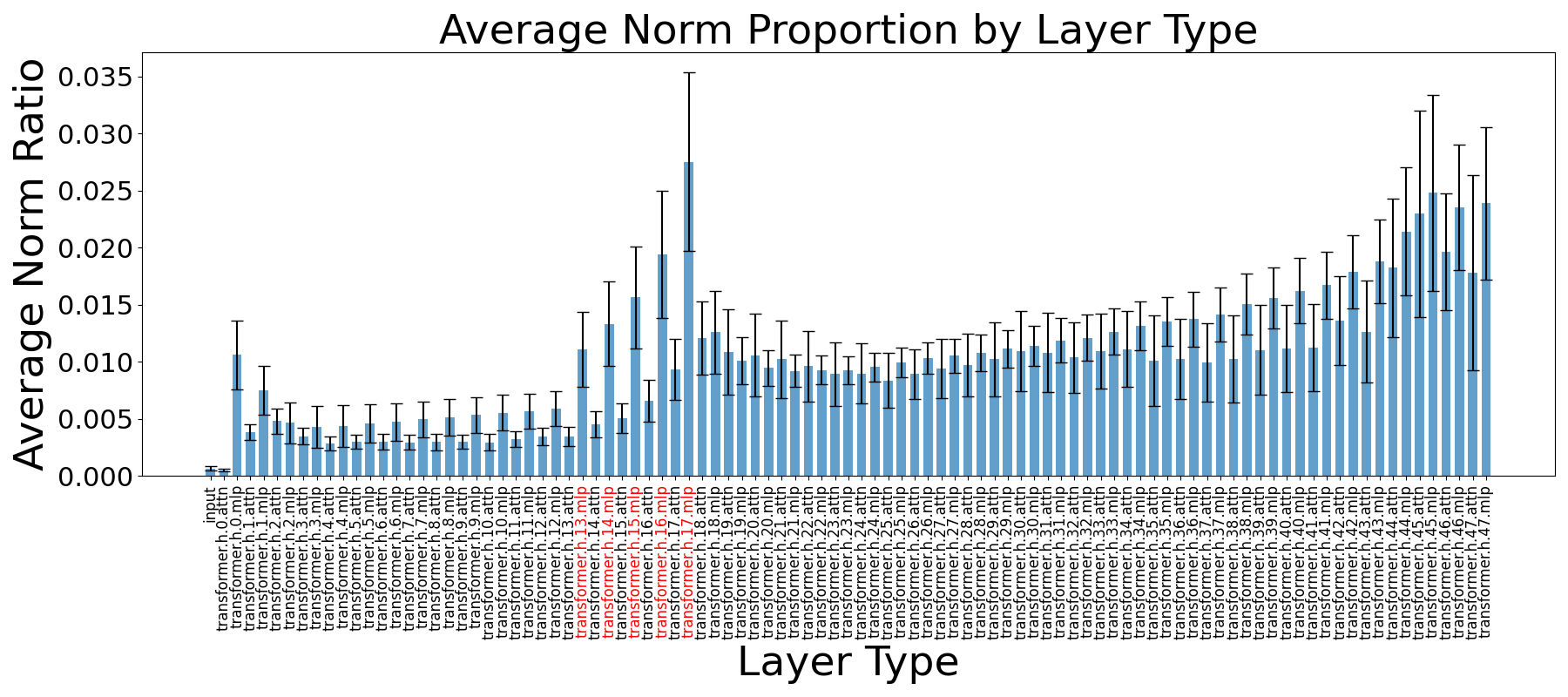} 
        \label{fig:activation-norm-growth-gpt2=xl-alphaedit}
    }
    \subfigure[Average Norm Proportion for GPT2-XL using MEMIT]{
        \includegraphics[width=\linewidth]{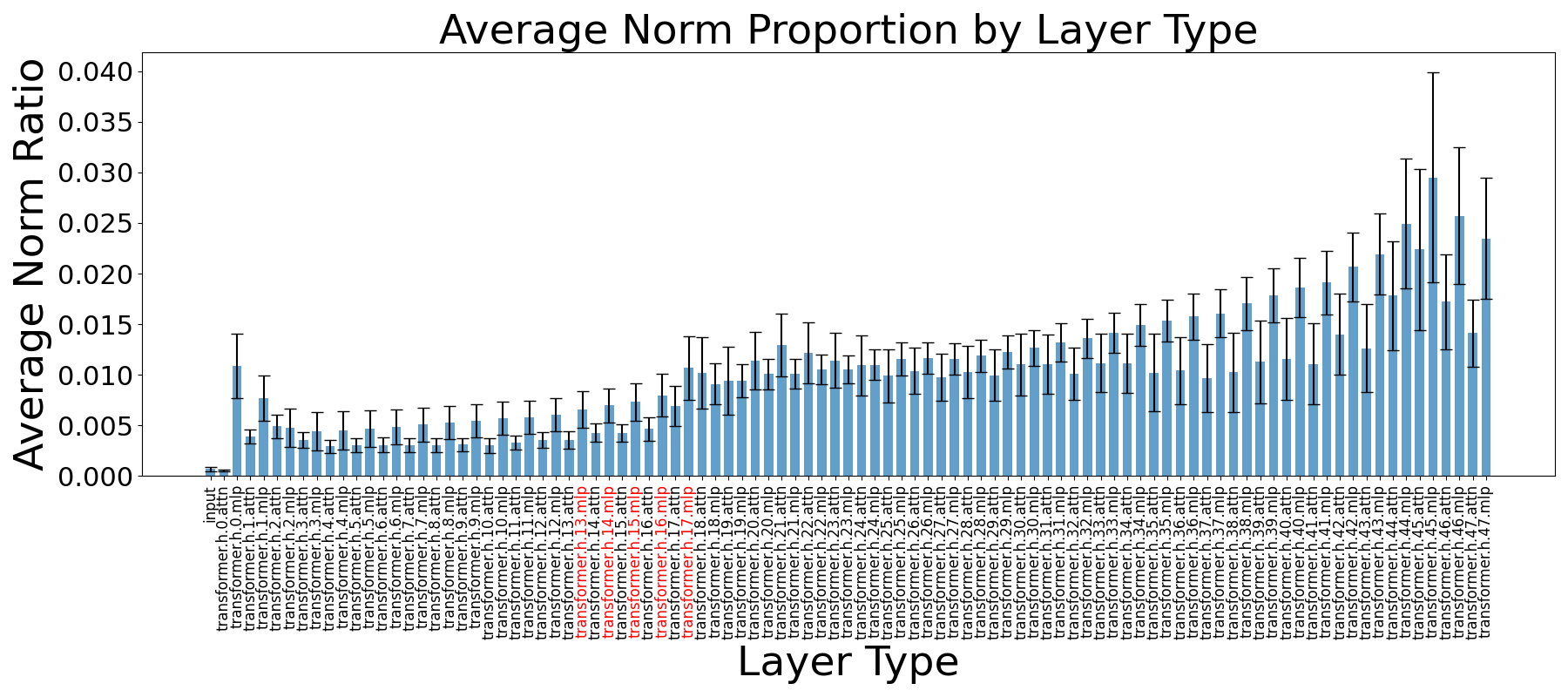} 
        \label{fig:activation-norm-growth-gpt2-xl-memit}
    }
    \caption{Activation norm growth for GPT2-XL.}
    \label{fig:activation-norm-growth-GPT2-XL}
\end{figure}

\begin{figure}[htbp]
    \centering
    \subfigure[Average Norm Proportion For Unedited Llama2-7B]{
        \includegraphics[width=\linewidth]{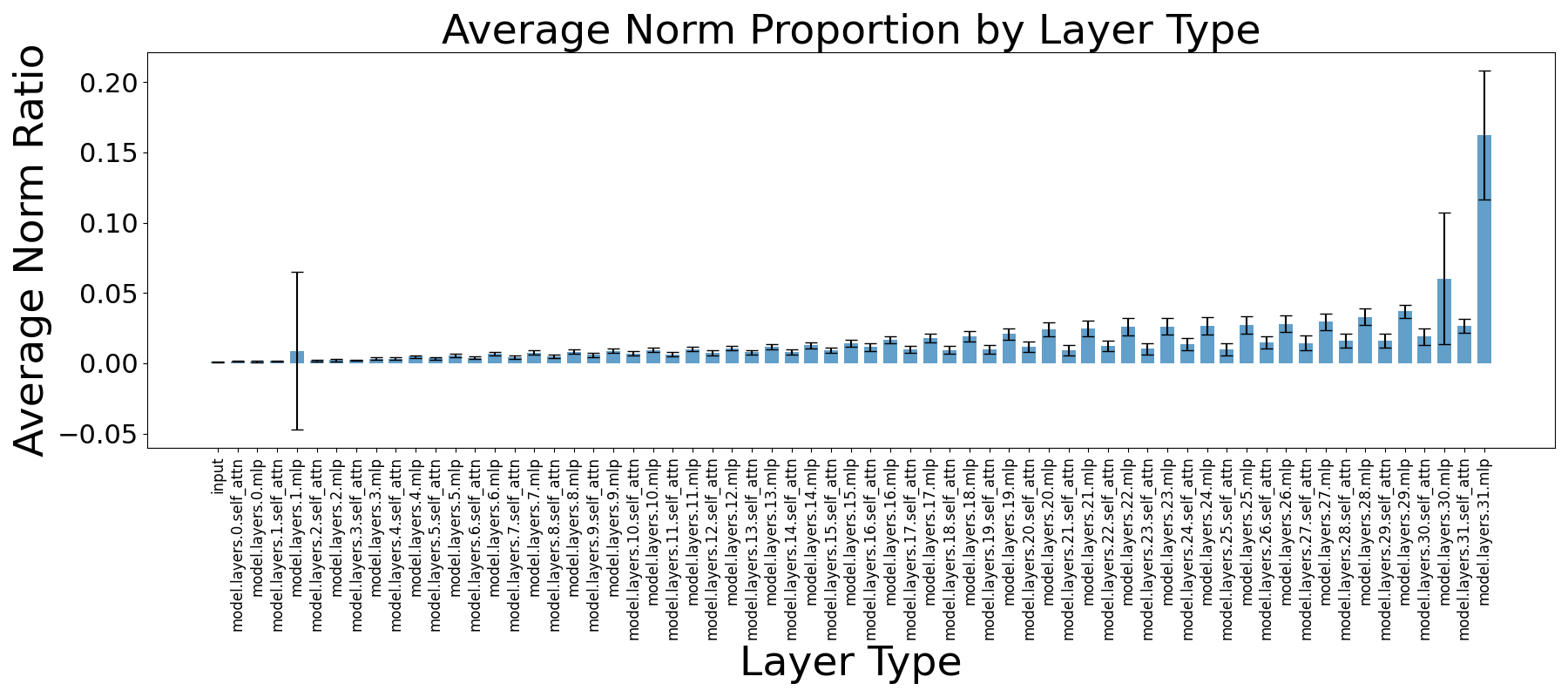} 
        \label{fig:activation-norm-growth-llama2-baseline}
    }
    \subfigure[Average Norm Proportion for Llama2-7B using Alphaedit]{
        \includegraphics[width=\linewidth]{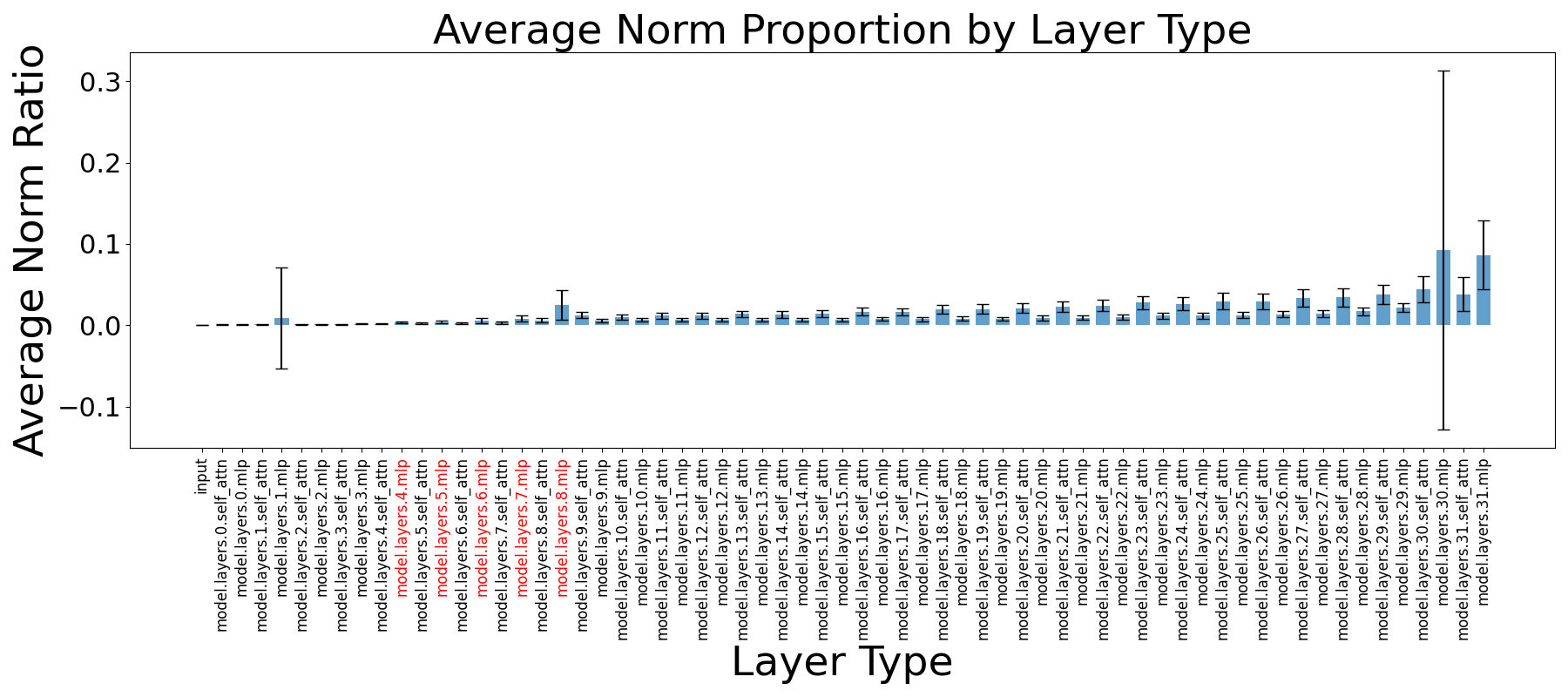} 
        \label{fig:activation-norm-growth-llama2-alphaedit}
    }
    \subfigure[Average Norm Proportion for Llama2-7B using MEMIT]{
        \includegraphics[width=\linewidth]{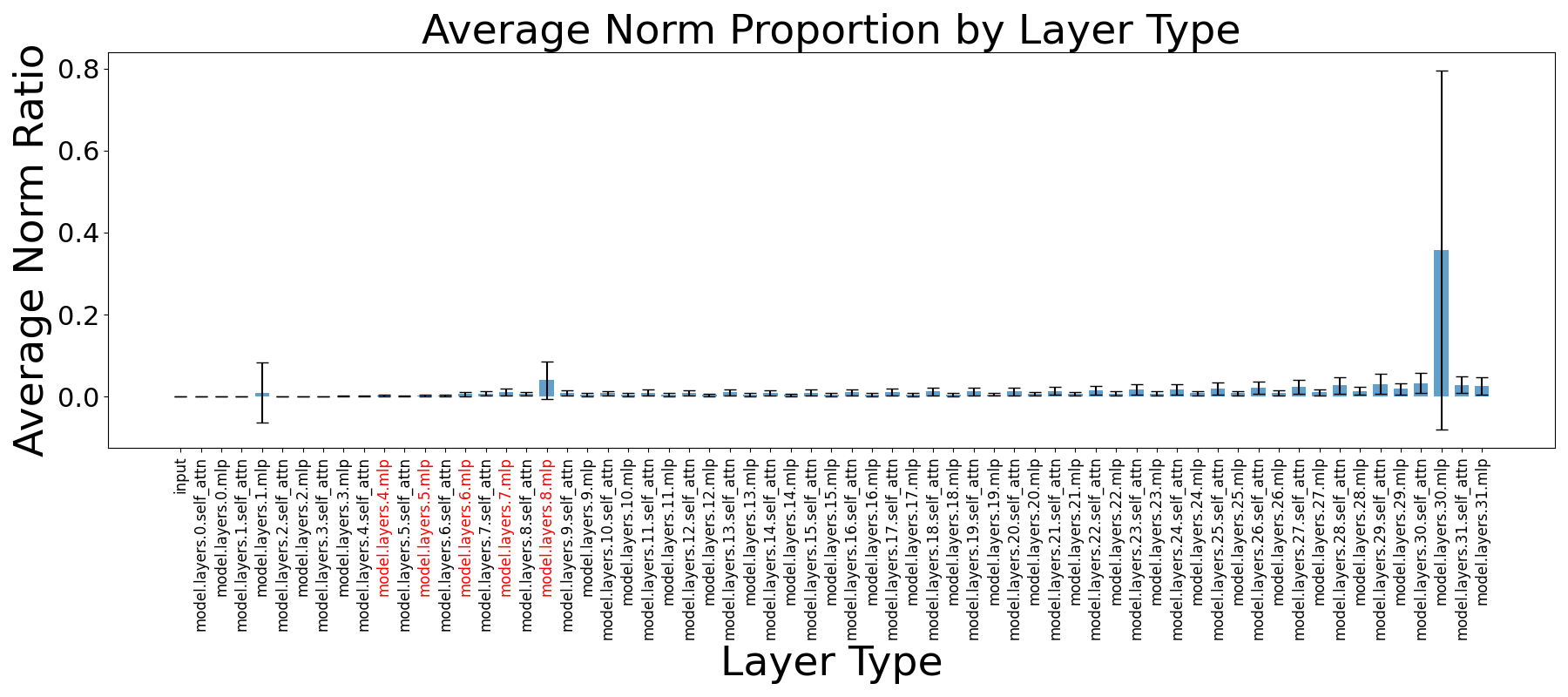} 
        \label{fig:activation-norm-growth-llama2-memit}
    }
    \caption{Activation norm growth for Llama2-7B.}
    \label{fig:activation-norm-growth-llama2-7b}
\end{figure}

\begin{figure}[htbp]
    \centering
    \subfigure[Average Norm Proportion For Unedited Llama3-8B]{
        \includegraphics[width=\linewidth]{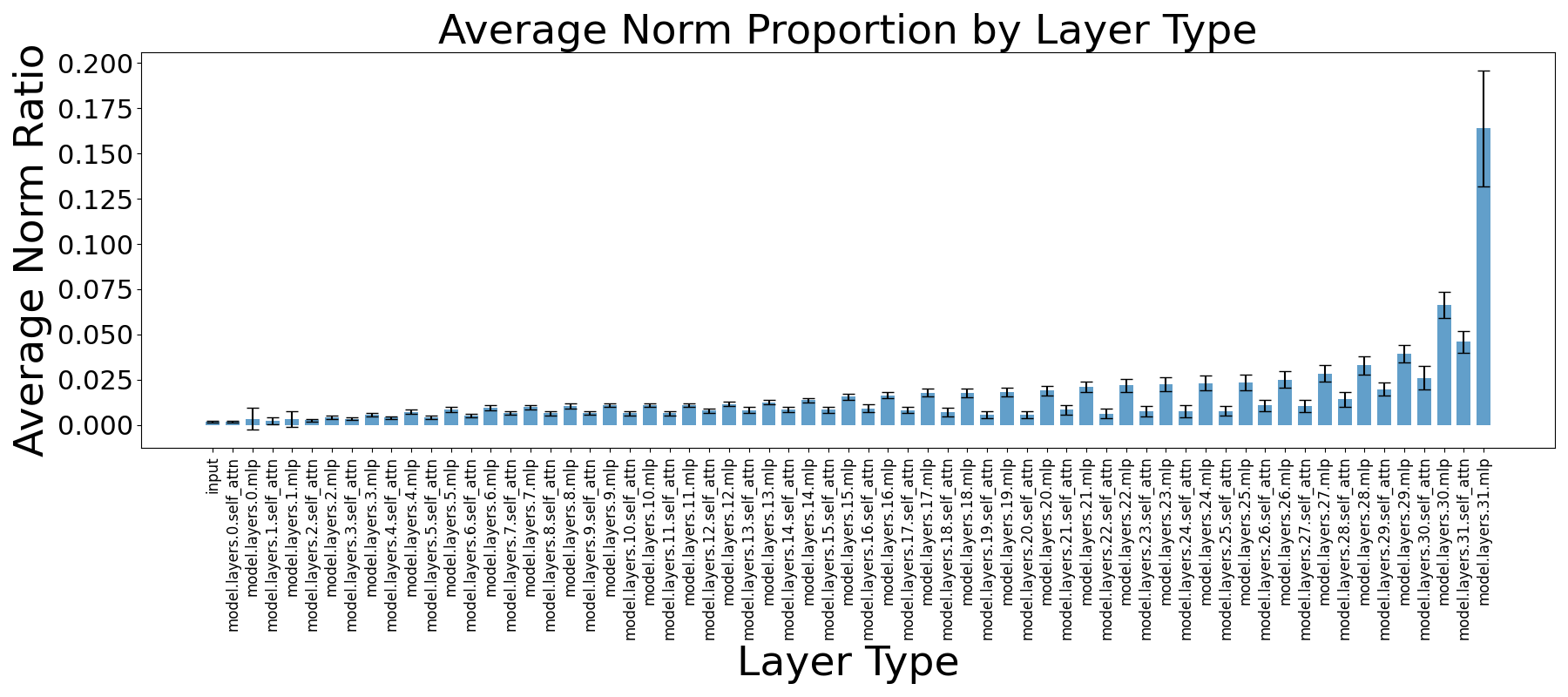} 
        \label{fig:activation-norm-growth-llama3-baseline-appendix}
    }
    \subfigure[Average Norm Proportion for Llama3-8B using Alphaedit]{
        \includegraphics[width=\linewidth]{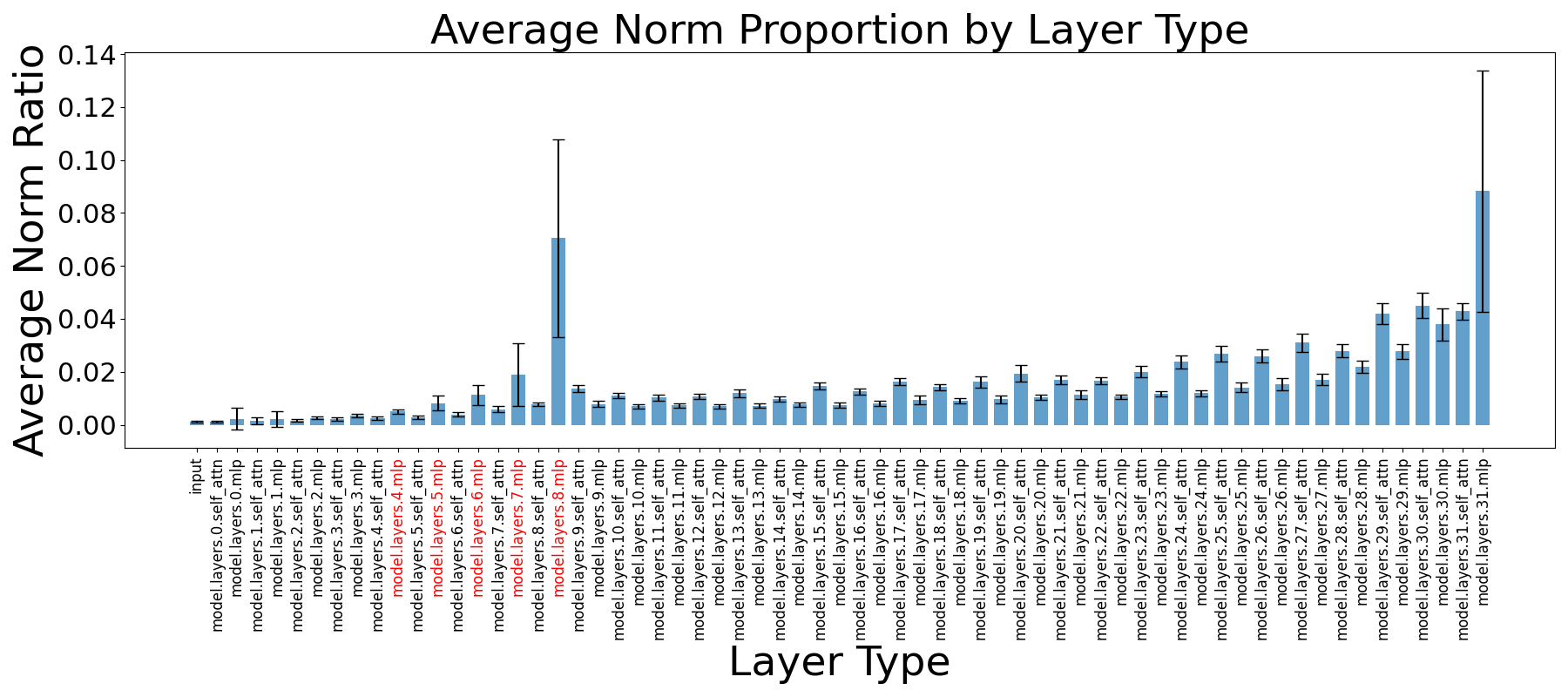} 
        \label{fig:activation-norm-growth-llama3-alphaedit-appendix}
    }
    \subfigure[Average Norm Proportion for Llama3-8B using MEMIT]{
        \includegraphics[width=\linewidth]{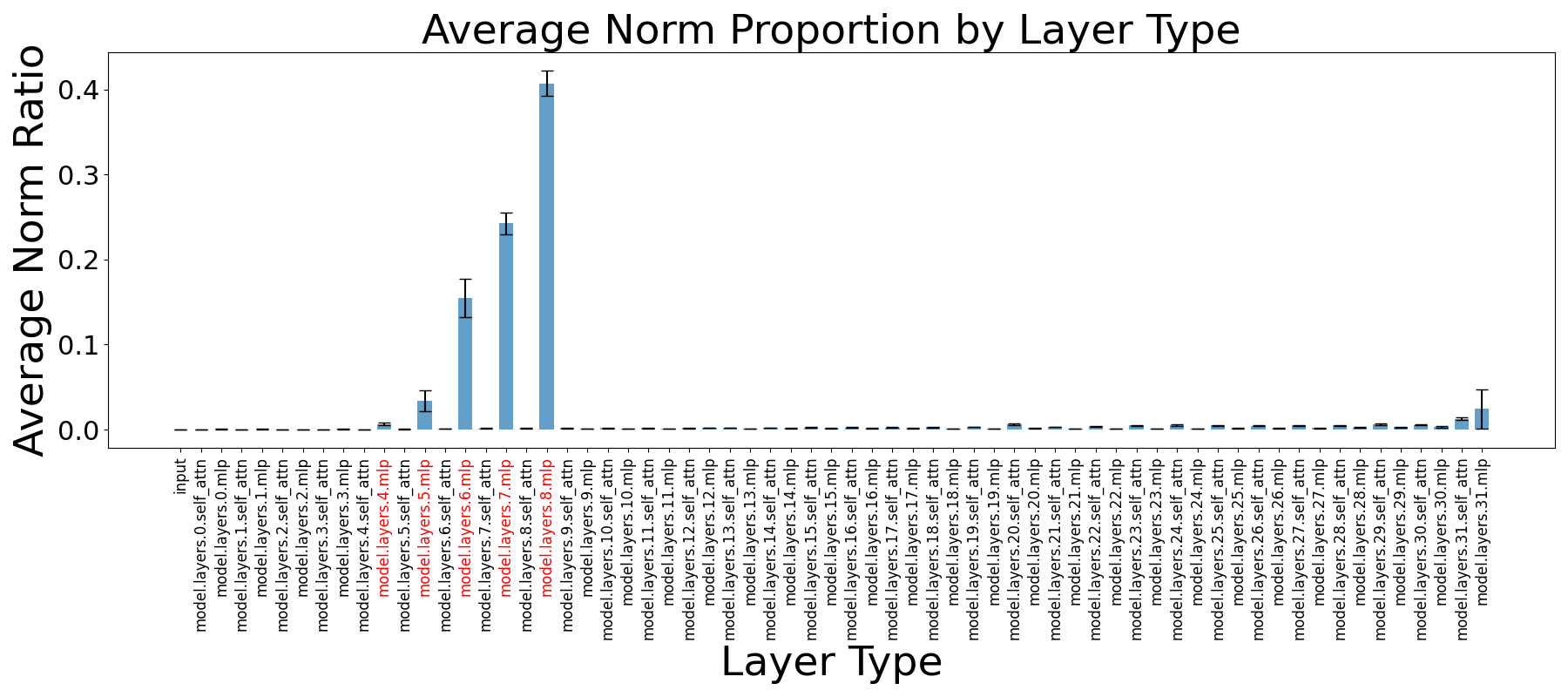} 
        \label{fig:activation-norm-growth-llama3-memit-appendix}
    }
    \caption{Activation norm growth for Llama3-8B.}
    \label{fig:activation-norm-growth-llama3-8b}
\end{figure}

\begin{figure}[htbp]
    \centering
    \subfigure[Average Norm Proportion For GPT2-XL using MEMIT]{
        \includegraphics[width=\linewidth]{figures/activation-norm-growth/average_norm_ratio_memit_gpt2-xl.png} 
        \label{fig:activation-norm-growth-gpt2xl-memit-appendix}
    }
    \subfigure[Average Norm Proportion for GPT2-XL using NC]{
        \includegraphics[width=\linewidth]{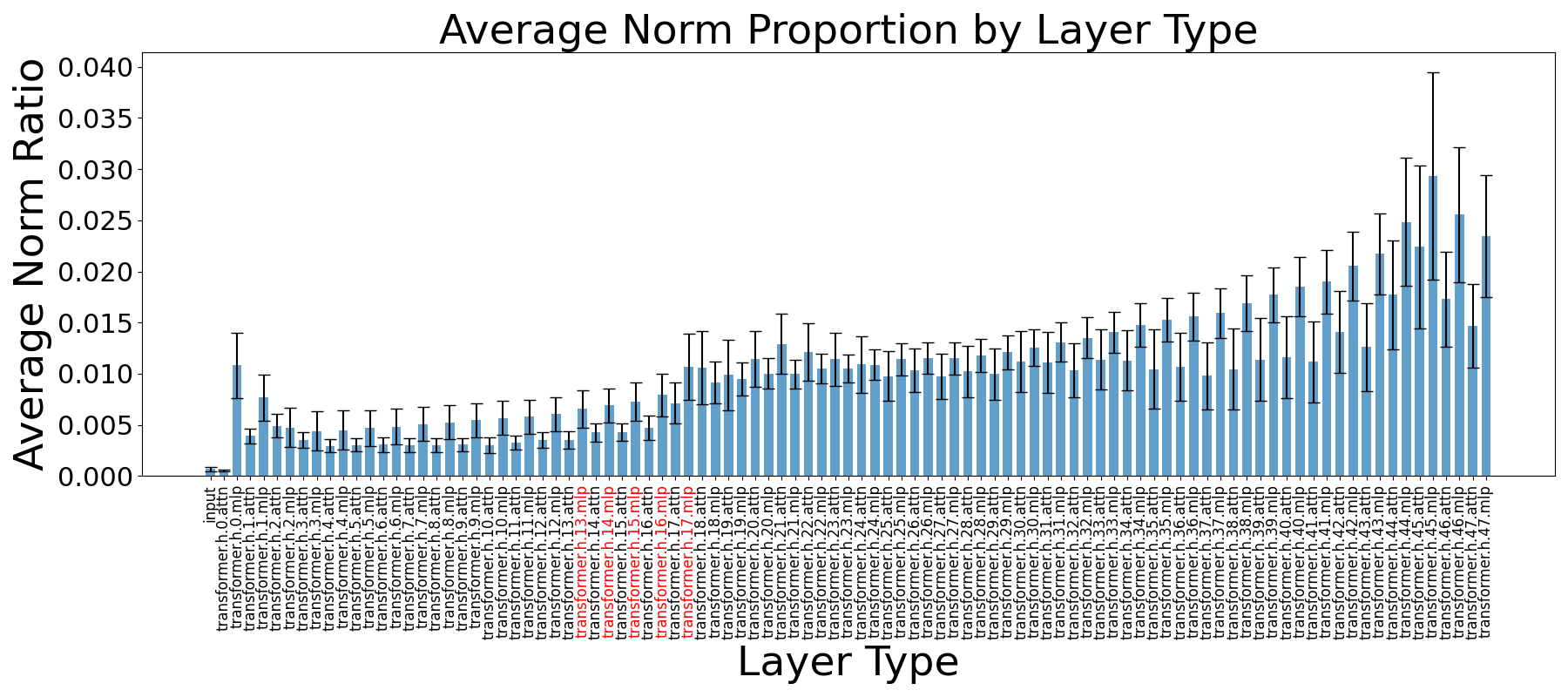} 
        \label{fig:activation-norm-growth-gpt2-xl-norm-appendix}
    }
    \subfigure[Average Norm Proportion for GPT2-XL using MPES + NC]{
        \includegraphics[width=\linewidth]{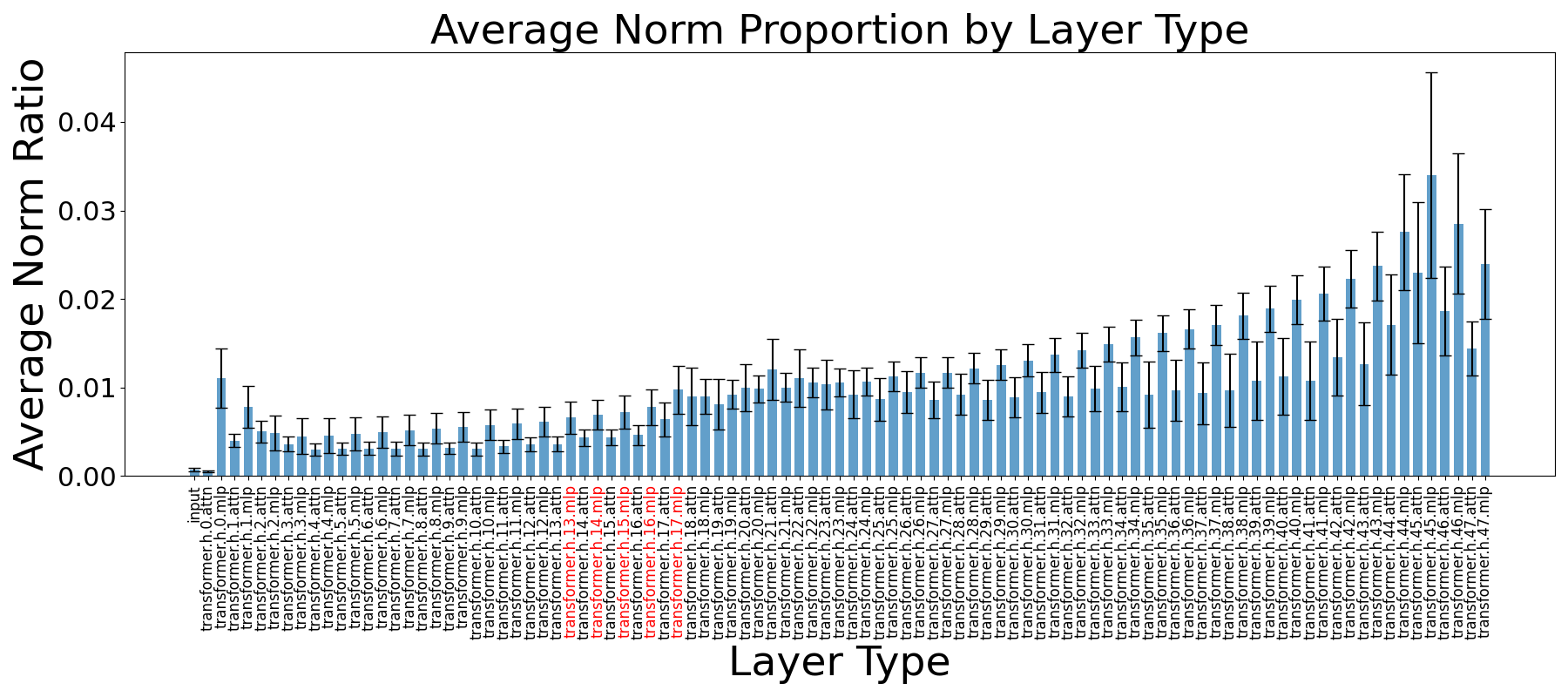} 
        \label{fig:activation-norm-growth-gpt2-xl-combination-apendix}
    }
    \caption{Activation norm growth for GPT2-XL using NC and MPES + NC.}
    \label{fig:activation-norm-growth-gpt2-xl-norm}
\end{figure}

\begin{figure}[htbp]
    \centering
    \subfigure[Average Norm Proportion For Llama2-7B using MEMIT]{
        \includegraphics[width=\linewidth]{figures/activation-norm-growth/average_norm_ratio_memit_llama2-7b.png} 
        \label{fig:activation-norm-growth-llama2-7b-baseline-appendix}
    }
    \subfigure[Average Norm Proportion for Llama2-7B using NC]{
        \includegraphics[width=\linewidth]{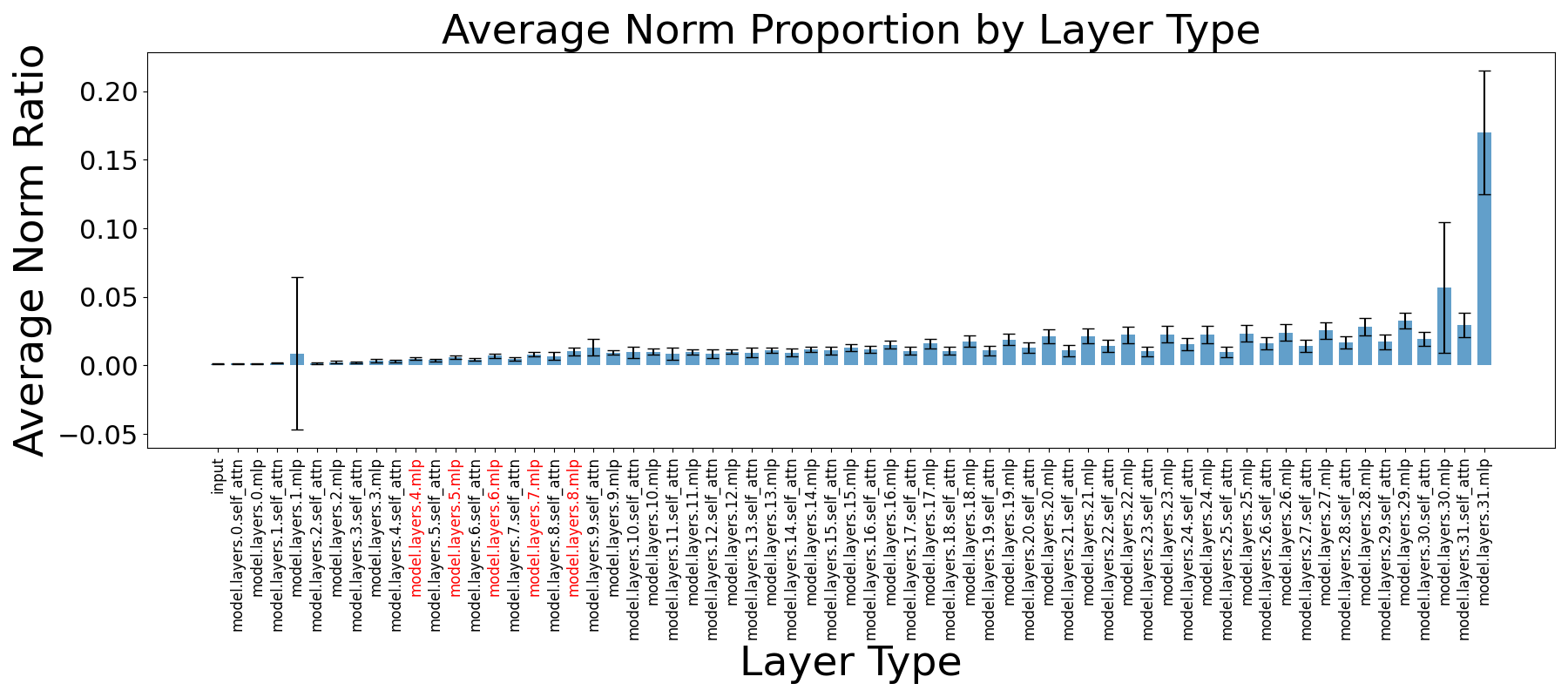} 
        \label{fig:activation-norm-growth-Llama2-7B-norm-appendix}
    }
    \subfigure[Average Norm Proportion for Llama2-7B using MPES + NC]{
        \includegraphics[width=\linewidth]{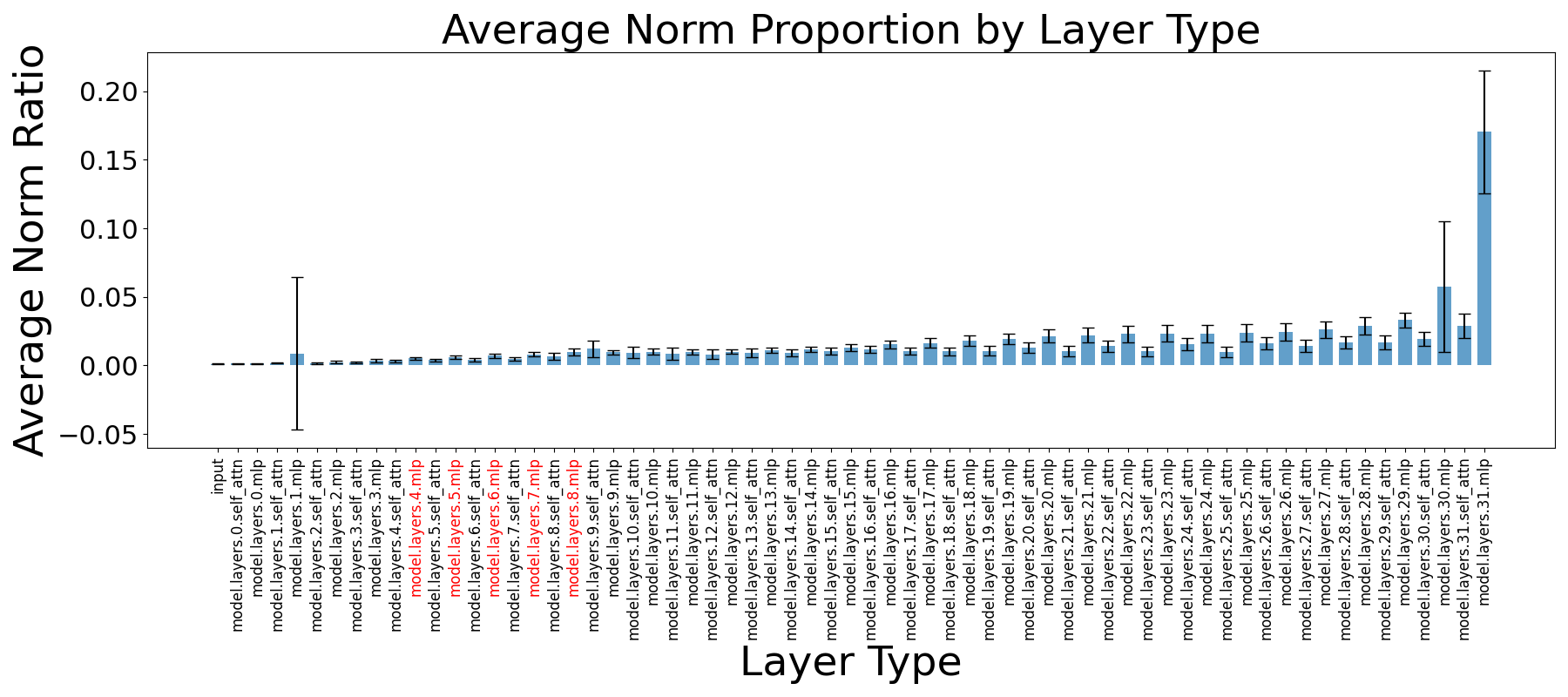} 
        \label{fig:activation-norm-growth-Llama2-7B-combination-apendix}
    }
    \caption{Activation norm growth for Llama2-7B using NC and MPES + NC.}
    \label{fig:activation-norm-growth-Llama2-7B-norm}
\end{figure}

\begin{figure}[htbp]
    \centering
    \subfigure[Average Norm Proportion For Llama3-8B using MEMIT]{
        \includegraphics[width=\linewidth]{figures/activation-norm-growth/average_norm_ratio_memit_llama3-8b.png} 
        \label{fig:activation-norm-growth-llama3-8b-baseline-appendix}
    }
    \subfigure[Average Norm Proportion for Llama3-8B using NC]{
        \includegraphics[width=\linewidth]{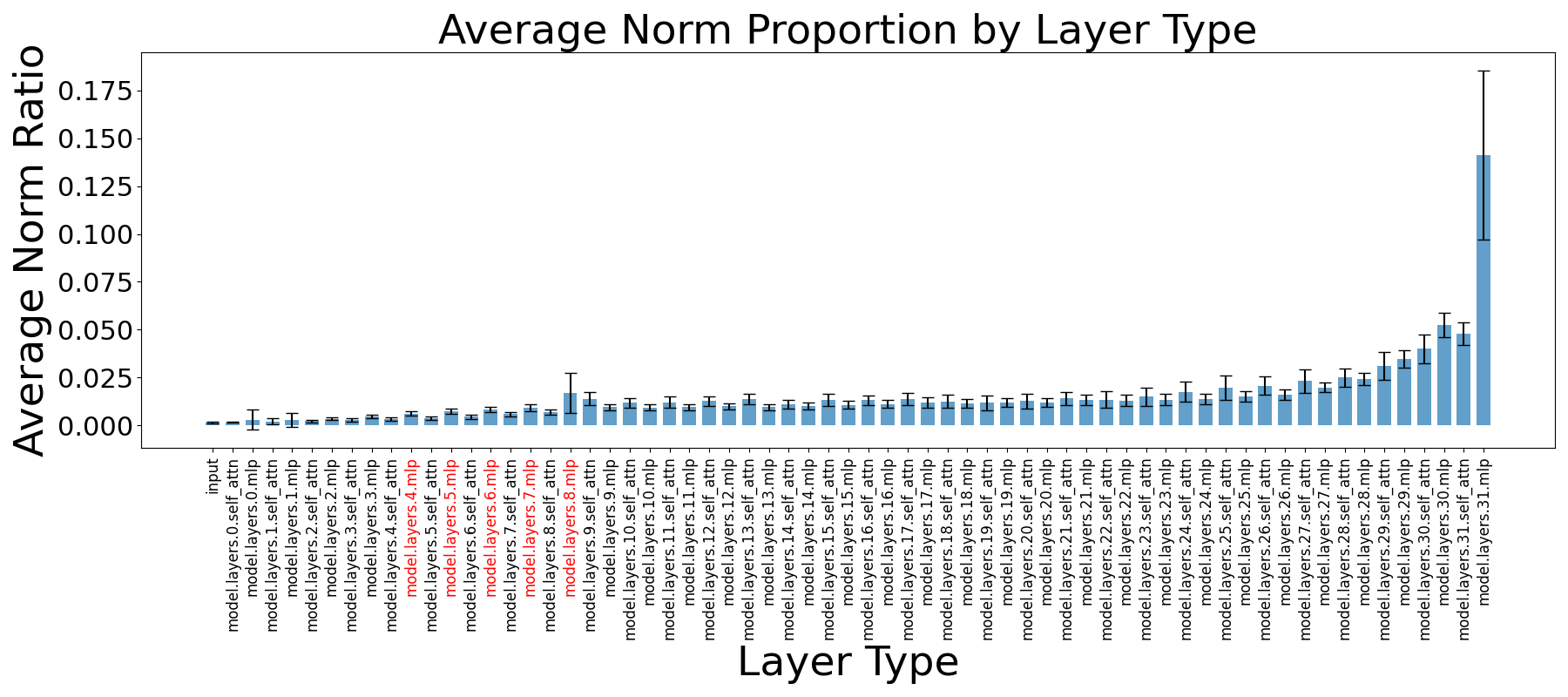} 
        \label{fig:activation-norm-growth-Llama3-8B-norm-appendix}
    }
    \subfigure[Average Norm Proportion for Llama3-8B using MPES + NC]{
        \includegraphics[width=\linewidth]{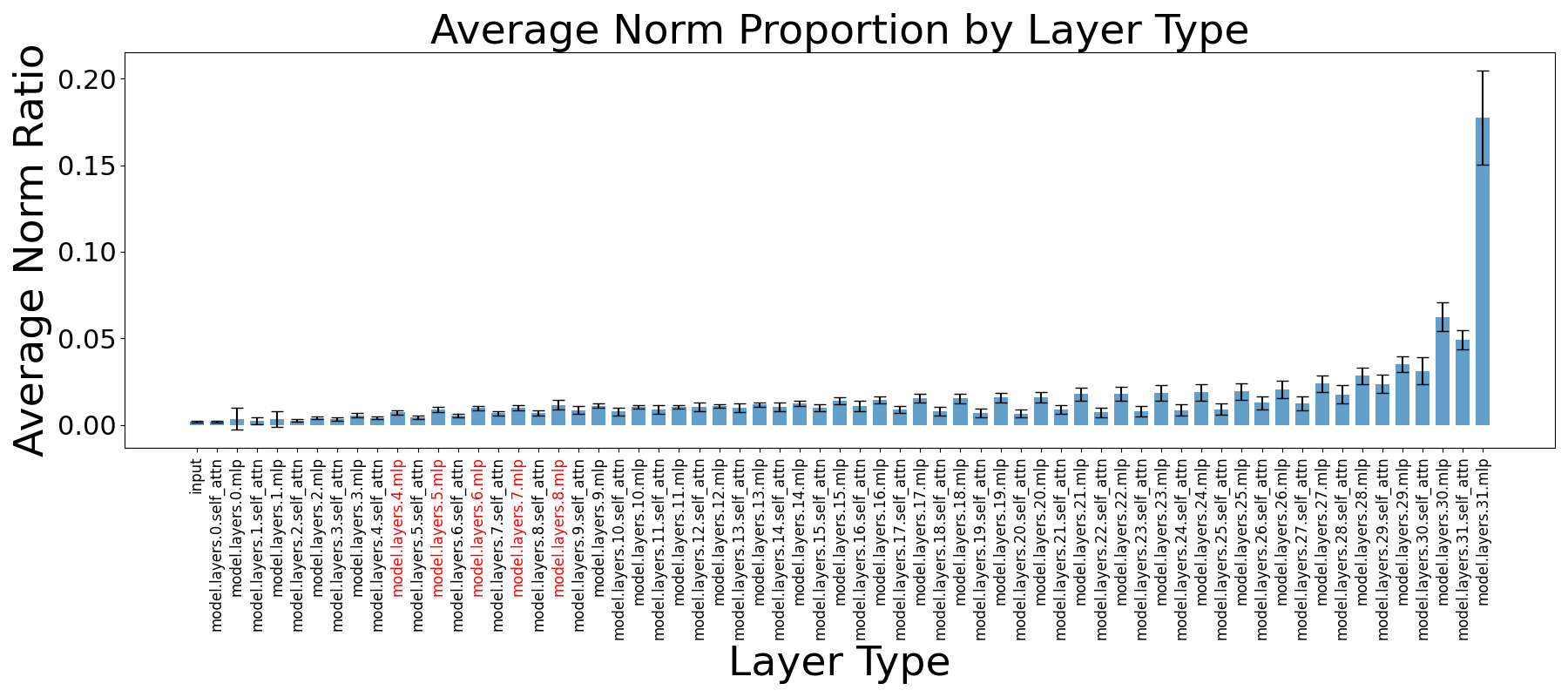} 
        \label{fig:activation-norm-growth-Llama3-8B-combination-apendix}
    }
    \caption{Activation norm growth for Llama3-8B using NC and MPES + NC.}
    \label{fig:activation-norm-growth-Llama3-8B-norm}
\end{figure}

For each of the 30,000 examples, we calculated the proportion of the activation norm at each layer. We then plotted the mean and standard deviation of these proportions for both the edited and unedited models in figures  \ref{fig:activation-norm-growth-GPT2-XL} - \ref{fig:activation-norm-growth-Llama3-8B-norm}. As shown in our results, the proportion of activation norms for the layers that were edited is significantly higher than their neighboring layers. In fact, some of the edited layers show the highest proportions overall. The edited layers are highlighted with \textit{red color} on the x-axis.
\subsection{Theoretical Analaysis of Growth of Vector Norm in a Summation}\label{appendix:activation-norm-growth-proof}
We want to understand the effect of excessive increase in the norm of a vector in a sum of vectors. First, let's start with an easy example where we have a summation of two vectors, $\mathbf{s} = \mathbf{a} + \mathbf{b}$ and then there is excessive increase in the norm of the first vector, that is $\mathbf{s} = k\mathbf{a} + \mathbf{b}$ where $k$ is some positive scalar. To evaluate the effect of this increase, we analyses the tendencies of the sum $\mathbf{s}$ as $k$ increases. 

We first want to understand the norm of $||\mathbf{s}||$. We have the following :

\begin{align*}
    ||\mathbf{s}||^{2} = ||k\mathbf{a} + \mathbf{b}||^{2} = k^{2} ||\mathbf{a}||^{2} + 2k \mathbf{a} \cdot \mathbf{b} + ||\mathbf{b}||^{2}
\end{align*}

From this we can clearly see that as $k$ increases the first term quadratic in $k$ will dominate. This means that as $k \to \infty$, $||\mathbf{s}||^{2} \to k^{2} ||\mathbf{a}||^{2}$, or $||\mathbf{s}|| \to k ||\mathbf{a}||$, which is the norm of the new vector. Thus, as the norm of one of the vectors in the summation increases, the norm of the summation tends to the norm of that vector with increasing norm.

Next, we look at the tendencies of the orientation of the summation as the norm of one vector increases. Let $\theta$ be the angle between $\mathbf{s}$ and $k\mathbf{a}$. Then, the cosine of the angle between the summation and the new vector $k\mathbf{a}$ is as follows (note that angle between $\mathbf{s}$ and $k\mathbf{a}$ is same as the angle between $\mathbf{s}$ and $\mathbf{a}$):

\begin{align*}
    \cos{\theta} = \frac{\mathbf{s} \cdot \mathbf{a}}{||\mathbf{s}|| ||\mathbf{a}||} = \frac{(k\mathbf{a}+\mathbf{b})\cdot \mathbf{a}}{||k\mathbf{a}+\mathbf{b}|| ||\mathbf{a}||}
\end{align*}

In the limit of $k \to \infty$, $||\mathbf{s}|| \to k ||\mathbf{a}||$ as shown above. Thus, the cosine expression in the limit can be written as:

\begin{align*}
    \underset{k \to \infty}{\cos{\theta}} = \frac{(k\mathbf{a}+\mathbf{b})\cdot \mathbf{a}}{k||\mathbf{a}||^2} = \frac{k\mathbf{a}\cdot \mathbf{a}}{k||\mathbf{a}||^2} + \frac{\mathbf{b}\cdot \mathbf{a}}{k||\mathbf{a}||^2} \\
    = 1 +  \frac{\mathbf{b}\cdot \mathbf{a}}{k||\mathbf{a}||^2}
\end{align*}

Thus, as $k \to \infty$, the cosine of angle between the sum and the vector tends to 1, or the angle between the summation and the vector tends to zero. This shows that as the norm of a vector in the summation continues to increase, the both the norm and the orientation of the summation tends towards the vector with increasing norm.

Finally, the same proof can be generalied to a summation of multiple vectors, where $\mathbf{b}$ represents the sum of the remaining vectors.

\newpage

\section{Internal Computations within Decoder-Only LLMs}\label{appendix:internal-computations}

The internal computations happening within a transformer-based decoder-only LLM is shown below:

\begin{align}
    f^l &= \texttt{LN1}(h^{l-1}) \label{eq:ln1} \\
    a^l &= \texttt{Att}(f^l)\\
    g^l &= \texttt{LN2}(h^{l-1} + a^l) \label{eq:ln2}\\
    m^l &= W^l_{proj} \sigma(W^l_{fc}g^{l}  + b^l_{fc}) + b_{proj}\label{eq:MLP-math}\\ 
    h^l &= h^{l-1} + a^l + m^l\label{eq:resiudal-recursive}
\end{align}

The intermediate hidden vector between each layer, $h^l$, is also sometimes referred to as the \textit{residual stream}. \texttt{LN1} represents the first LayerNorm (or equivalently RMSNorm for Llama models) that acts just before the attention module and \texttt{LN2} represents the second LayerNorm just before the MLP module. \texttt{Att} represents the self-attention module in an LLMs whereas the action of a traditional MLP module is written in terms of the individual weight matrices ($W_{fc}, W_{proj}$). As the vectors computed in the attention and MLP modules get added back to the residual stream at each layer, the residual stream represents a summation of an increasing number of vectors as we go deeper into the model.

\section{Proof for MEMIT + MPES + NC Objective}\label{appendix:encore-proof}
First we have that we can write the equation \ref{eq:encore_objective} in term of matrix form where we can stack the $k_{0}^{i}$. Specifically, we define $K_{0} = [k_{0}^{1} \; | k_{0}^{2} \; | \cdots \; | k_{0}^{P}]$, 
$K_{1} = [k_{e}^{1} \; | k_{e}^{2} \; | \cdots \; | k_{e}^{B}]$ and $V_{1} = [v_{e}^{1} \; | v_{e}^{2} \; | \cdots \; | v_{e}^{B}]$

column wise and instead the L2 norm will become the frobenius norm and we have that
\small
\begin{equation*}
    \lambda_p \left\|\hat{W}K_{0}- W_0K_{0}\right\|^2_F + \left\|\hat{W}K_{1}- V_{1}\right\|^2_F + \lambda_n \left\|\hat{W}- W_0\right\|^2_F   
\end{equation*}
We can differentiate this expression with respect to $\hat{W}$ and let it equal to 0, we get the following 
\begin{align*}
    \lambda_p \hat{W}K_{0}K_{0}^{T} - \lambda_p W_{0}K_{0}K_{0}^{T} + \hat{W}K_{1}K_{1}^{T} 
    \\
    - V_{1}K_{1}^{T} + \lambda_n \hat{W} - \lambda_n W_{0} = 0
\end{align*}
\begin{align*}
    \lambda_p \hat{W}K_{0}K_{0}^{T} +  \hat{W}K_{1}K_{1}^{T} +  \lambda_n \hat{W} \\
    = \lambda_p W_{0}K_{0}K_{0}^{T} + V_{1}K_{1}^{T} + \lambda_n W_{0}
\end{align*}
Since $\hat{W} = W_{0} + \Delta$ we have the following
\begin{align*}
    \lambda_p (W_{0} + \Delta)K_{0}K_{0}^{T} +  (W_{0} + \Delta)K_{1}K_{1}^{T} +  \lambda_n (W_{0} + \Delta) \\
    =\lambda_p W_{0}K_{0}K_{0}^{T} + V_{1}K_{1}^{T} + \lambda_n W_{0}
\end{align*}
\begin{align*}
    \lambda_p W_{0} K_{0}K_{0}^{T} + \lambda_p \Delta K_{0}K_{0}^{T} +  W_{0}K_{1}K_{1}^{T} + \Delta K_{1}K_{1}^{T} \\
    +  \lambda_n (W_{0} + \Delta) 
    =\lambda_p W_{0}K_{0}K_{0}^{T} + V_{1}K_{1}^{T} + \lambda_n W_{0}
\end{align*}
\begin{align*}
    &\Delta(\lambda_p K_{0}K_{0}^{T} + K_{1}K_{1}^{T} + \lambda_n I) = (V_{1} - W_{0}K_{1})K_{1}^{T}
    \\
    &\Delta = (V_{1} - W_{0}K_{1})K_{1}^{T} (\lambda_p K_{0}K_{0}^{T} + K_{1}K_{1}^{T} + \lambda_n I)^{-1}
\end{align*}

\section{Editing Hyperparameters and Hardware Details}
Here we present the hyperparameters that we find and used for the MPES, Norm constraint, and MPES + Norm constraint. Tables \ref{tab:hparams-mpes-mcf-appendix}, \ref{tab:hparams-norm-mcf-appendix} and \ref{tab:hparams-encore-mcf-appendix} show the hyperparameters for CounterFact dataset and tables \ref{tab:hparams-norm-zsre-appendix},\ref{tab:hparams-mpes-zsre} and \ref{tab:hparams-encore-zsre-appendix} show the hyperparameters for the zsRE dataset.
Additionally, there is one more hyperparameter that requires tuning, as stopping immediately when the target token becomes the most probable may not always be optimal. We define this hyperparameter as the probability cutoff, which determines how many additional steps we take before stopping. Specifically, a cutoff of $+n$ means that we continue for $n$ more steps after the target token first becomes the most probable.

All experiments in this paper are done on NVIDIA A6000, including experiments where editing speeds of different methods are timed. 

\begin{table}[htbp]
\vskip -0.1in
\begin{center}
\begin{adjustbox}{max width=0.37\textwidth}
\begin{sc}
\begin{tabular}{lccccccr}
\toprule
Method & Model & \multirow{2}{*}{\makecell{$\lambda_{p}$}} & \multirow{2}{*}{\makecell{Probability \\ Cut Off}}  \\
& & & \\
\midrule
AlphaEdit + MPES & GPT2-XL & 20,000 & +1  \\
    & Llama2-7B & 15,000 & +0  \\
    & Llama3-8B & 15,000 & +0 \\

\midrule
MEMIT + MPES & GPT2-XL & 20,000 & +2  \\
    & Llama2-7B & 15,000 & +1  \\
    & Llama3-8B & 15,000 & +2 \\
\bottomrule
\end{tabular}
\end{sc}
\end{adjustbox}
\end{center}
\vskip -0.2in
\caption{Hyperparameters for different algorithms with MPES on CouterFact dataset }\label{tab:hparams-mpes-mcf-appendix}
\vskip -0.1in
\end{table}

\begin{table}[htbp]
\vskip -0.1in
\begin{center}
\begin{adjustbox}{max width=0.37\textwidth}
\begin{sc}
\begin{tabular}{lcccccc}
\toprule
Method & Model & \makecell{$\lambda_{p}$} & \makecell{$\lambda_{n}$}\\
\midrule
Norm Constraint  &  GPT2-XL   & 20,000 &  10 \\
         & Llama2-7B  & 15,000 & 10 \\
         & Llama3-8B  & 15,000  & 20  \\
\bottomrule
\end{tabular}
\end{sc}
\end{adjustbox}
\end{center}
\vskip -0.2in
\caption{Hyperparameters for Norm Constraint on CouterFact dataset }\label{tab:hparams-norm-mcf-appendix}
\vskip -0.1in
\end{table}

\begin{table}[htbp]
\vskip -0.1in
\begin{center}
\begin{adjustbox}{max width=0.37\textwidth}
\begin{sc}
\begin{tabular}{lcccccc}
\toprule
Method & Model & \makecell{$\lambda_{p}$} & \makecell{$\lambda_{n}$} & \makecell{Probability \\ Cut Off}\\
\midrule
MPES + NC  &  GPT2-XL   & 20,000 & 10 & +3  \\
         & Llama2-7B  & 15,000 & 10 & +2  \\
         & Llama3-8B  & 15,000 & 20 & +1  \\
\bottomrule
\end{tabular}
\end{sc}
\end{adjustbox}
\end{center}
\vskip -0.2in
\caption{Hyperparameters for MPES + Norm Constraint on CouterFact dataset }\label{tab:hparams-encore-mcf-appendix}
\vskip -0.1in
\end{table}

\begin{table}[htbp]
\vskip -0.1in
\begin{center}
\begin{adjustbox}{max width=0.37\textwidth}
\begin{sc}
\begin{tabular}{lcccccc}
\toprule
Method & Model & \makecell{$\lambda_{p}$} & \makecell{$\lambda_{n}$}\\
\midrule
Norm Constraint  &  GPT2-XL   & 20,000 &  40 \\
         & Llama2-7B  & 15,000 & 10 \\
         & Llama3-8B  & 15,000  & 20  \\
\bottomrule
\end{tabular}
\end{sc}
\end{adjustbox}
\end{center}
\vskip -0.2in
\caption{Hyperparameters for Norm Constraint on zsRE dataset }\label{tab:hparams-norm-zsre-appendix}
\vskip -0.1in
\end{table}

\begin{table}[htbp]
\vskip -0.1in
\begin{center}
\begin{adjustbox}{max width=0.37\textwidth}
\begin{sc}
\begin{tabular}{lccccccr}
\toprule
Method & Model & \multirow{2}{*}{\makecell{$\lambda_{p}$}} & \multirow{2}{*}{\makecell{Probability \\ Cut Off}}  \\
& & & \\
\midrule
AlphaEdit + MPEs & GPT2-XL & 20,000 & +1  \\
    & Llama2-7B & 15,000 & +1  \\
    & Llama3-8B & 15,000 & +3 \\

\midrule
MEMIT + MPEs & GPT2-XL & 20,000 & +5  \\
    & Llama2-7B & 15,000 & +1  \\
    & Llama3-8B & 15,000 & +4 \\
\bottomrule
\end{tabular}
\end{sc}
\end{adjustbox}
\end{center}
\vskip -0.2in
\caption{Hyperparameters for different algorithms with MPES on zsRE dataset }\label{tab:hparams-mpes-zsre}
\vskip -0.1in
\end{table}

\begin{table}[htbp]
\vskip -0.1in
\begin{center}
\begin{adjustbox}{max width=0.37\textwidth}
\begin{sc}
\begin{tabular}{lcccccc}
\toprule
Method & Model & \makecell{$\lambda_{p}$} & \makecell{$\lambda_{n}$} & \makecell{Probability \\ Cut Off}\\
\midrule
MPES + NC  &  GPT2-XL   & 20,000 & 40 & +4  \\
         & Llama2-7B  & 15,000 & 10 & +2  \\
         & Llama3-8B  & 15,000 & 10 & +3  \\
\bottomrule
\end{tabular}
\end{sc}
\end{adjustbox}
\end{center}
\vskip -0.2in
\caption{Hyperparameters for MEMIT+ Norm Constraint on zsRE dataset }\label{tab:hparams-encore-zsre-appendix}
\vskip -0.1in
\end{table}


\section{Norm growth Result}
\begin{figure}[htbp]
    \centering
    \subfigure[Norm growth of GPT2-XL using Alphaedit]{
        \includegraphics[width=0.45\linewidth]{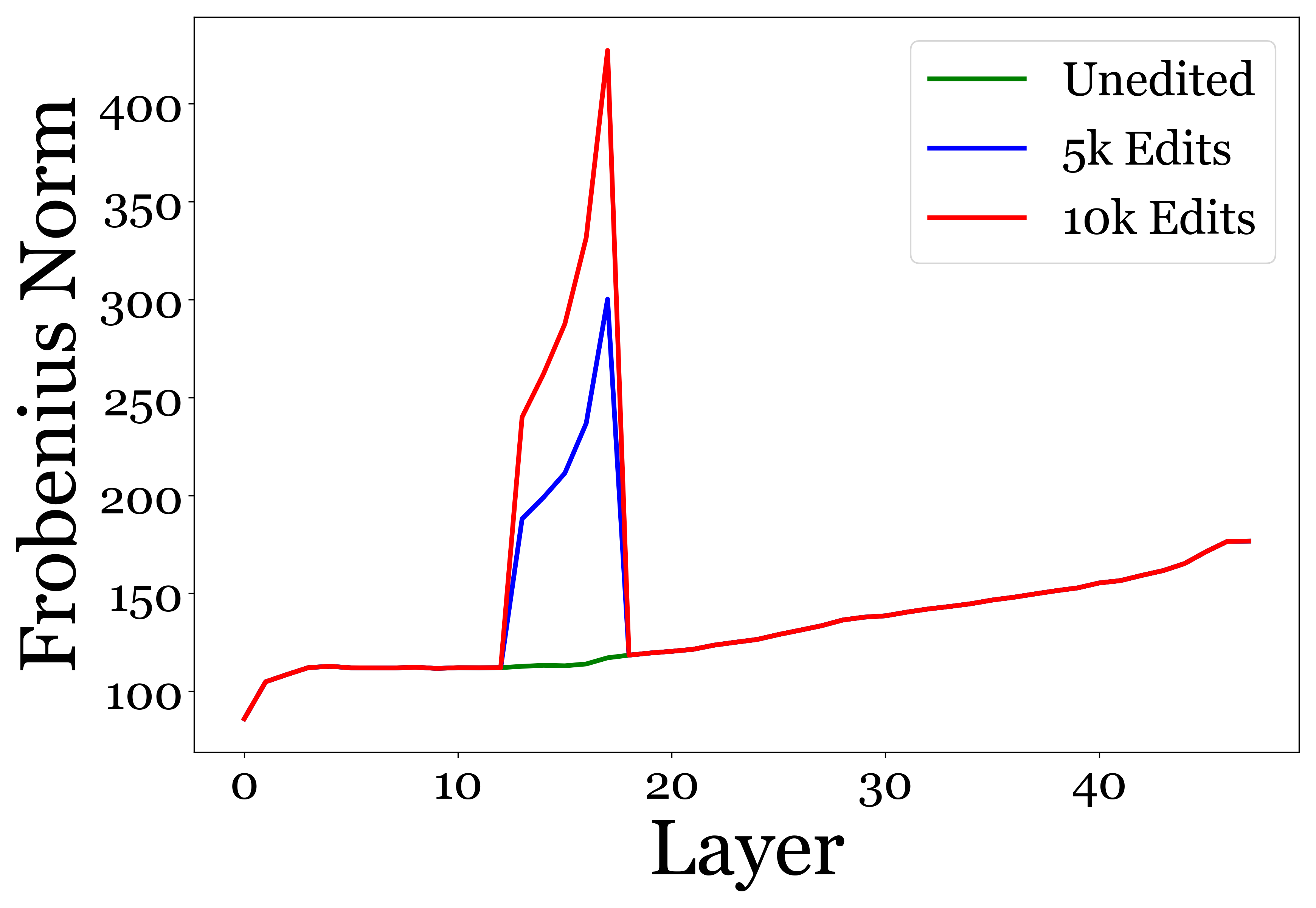} 
        \label{fig:norm-growth-gpt2-xl-alphaedit}
    }
    \subfigure[Norm growth of GPT2-XL using MEMIT]{
        \includegraphics[width=0.45\linewidth]{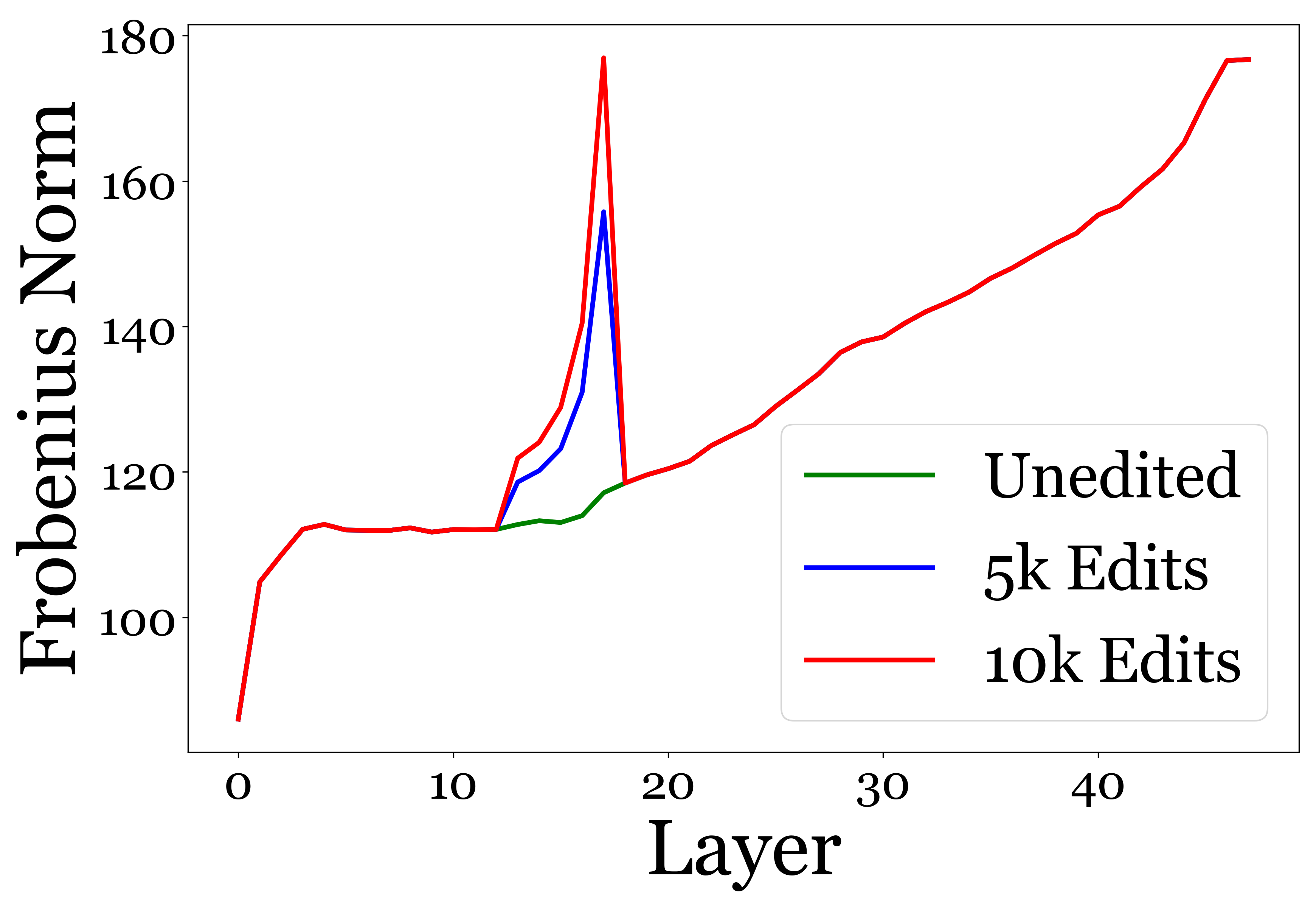} 
        \label{fig:norm-growth-gpt2-xl-memit}
    }
    \caption{Norm growth of GPT2-XL using AlphaEdit and MEMIT} 
    \label{fig:norm-growth-gpt2-xl}
\end{figure}

\begin{figure}[htbp]
    \centering
    \subfigure[Norm growth of GPT2-XL using NC]{
        \includegraphics[width=0.45\linewidth]{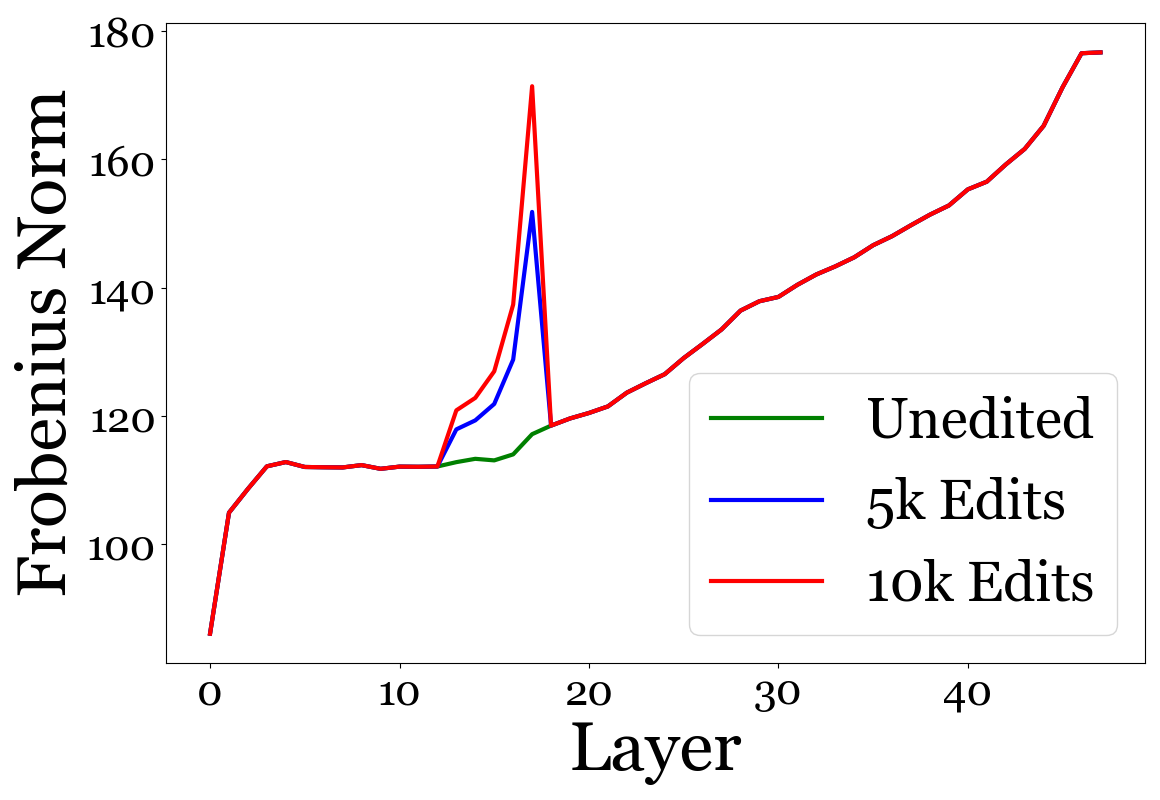} 
        \label{fig:norm-growth-gpt2-xl-memit-nc}
    }
    \subfigure[Norm growth of GPT2-XL using MPES + NC]{
        \includegraphics[width=0.45\linewidth]{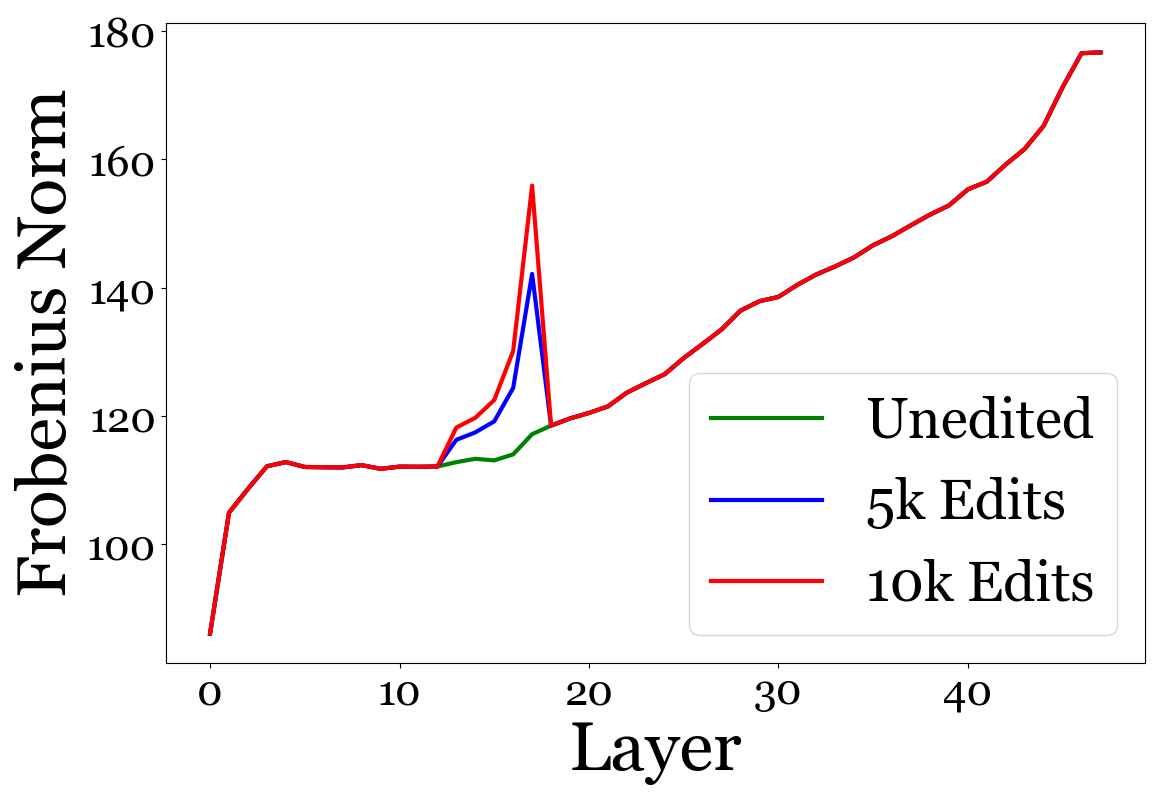} 
        \label{fig:norm-growth-gpt2-xl-memit-mpes+nc}
    }
    \caption{Norm growth of GPT2-XL using NC and MPES + NC} 
    \label{fig:norm-growth-gpt2-xl-encore}
\end{figure}

\begin{figure}[htbp]
    \centering
    \subfigure[Norm growth of Llama2-7B using Alphaedit]{
        \includegraphics[width=0.45\linewidth]{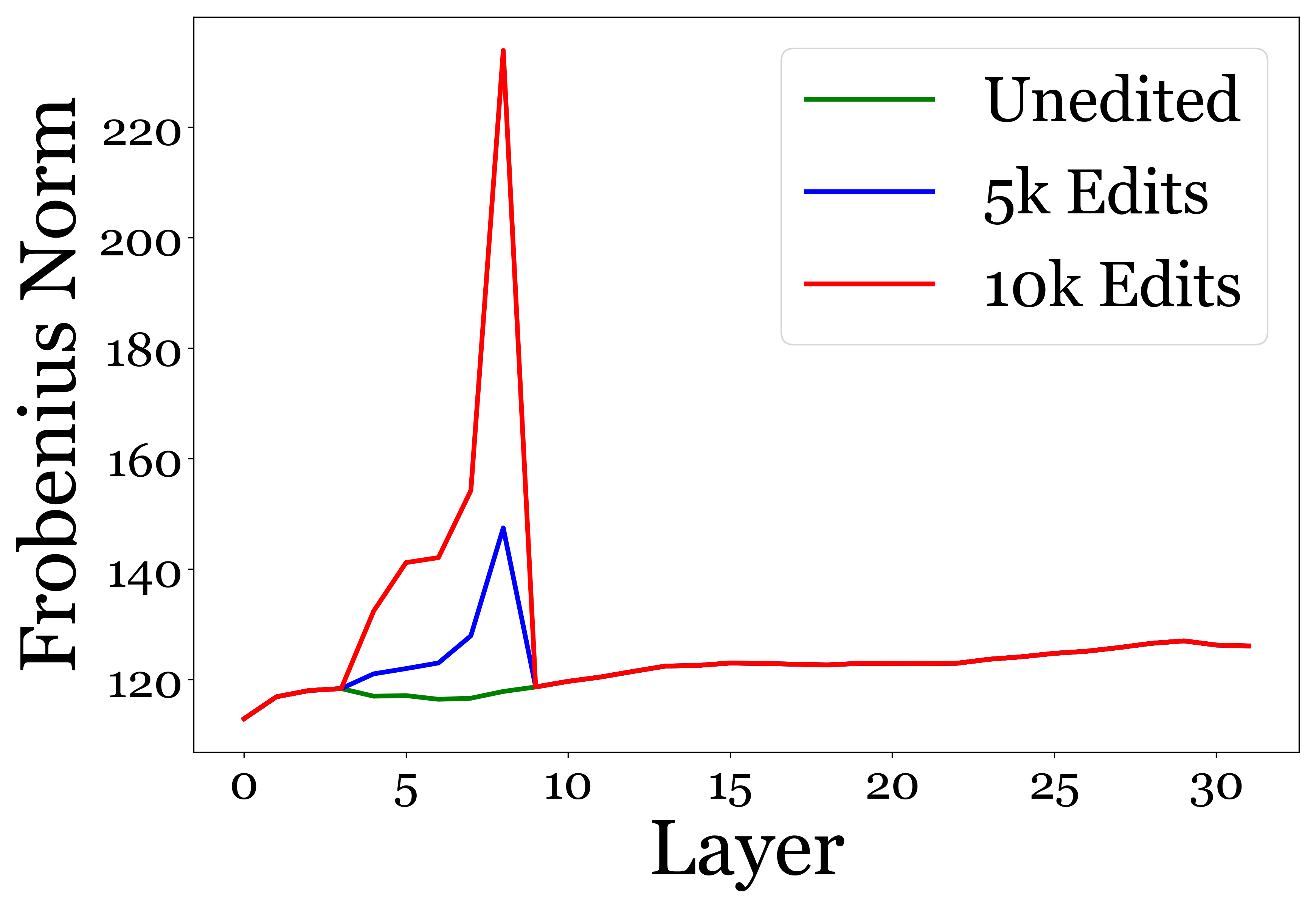} 
        \label{fig:norm-growth-llama2-7b-alphaedit}
    }
    \subfigure[Norm growth of Llama2-7B using MEMIT]{
        \includegraphics[width=0.45\linewidth]{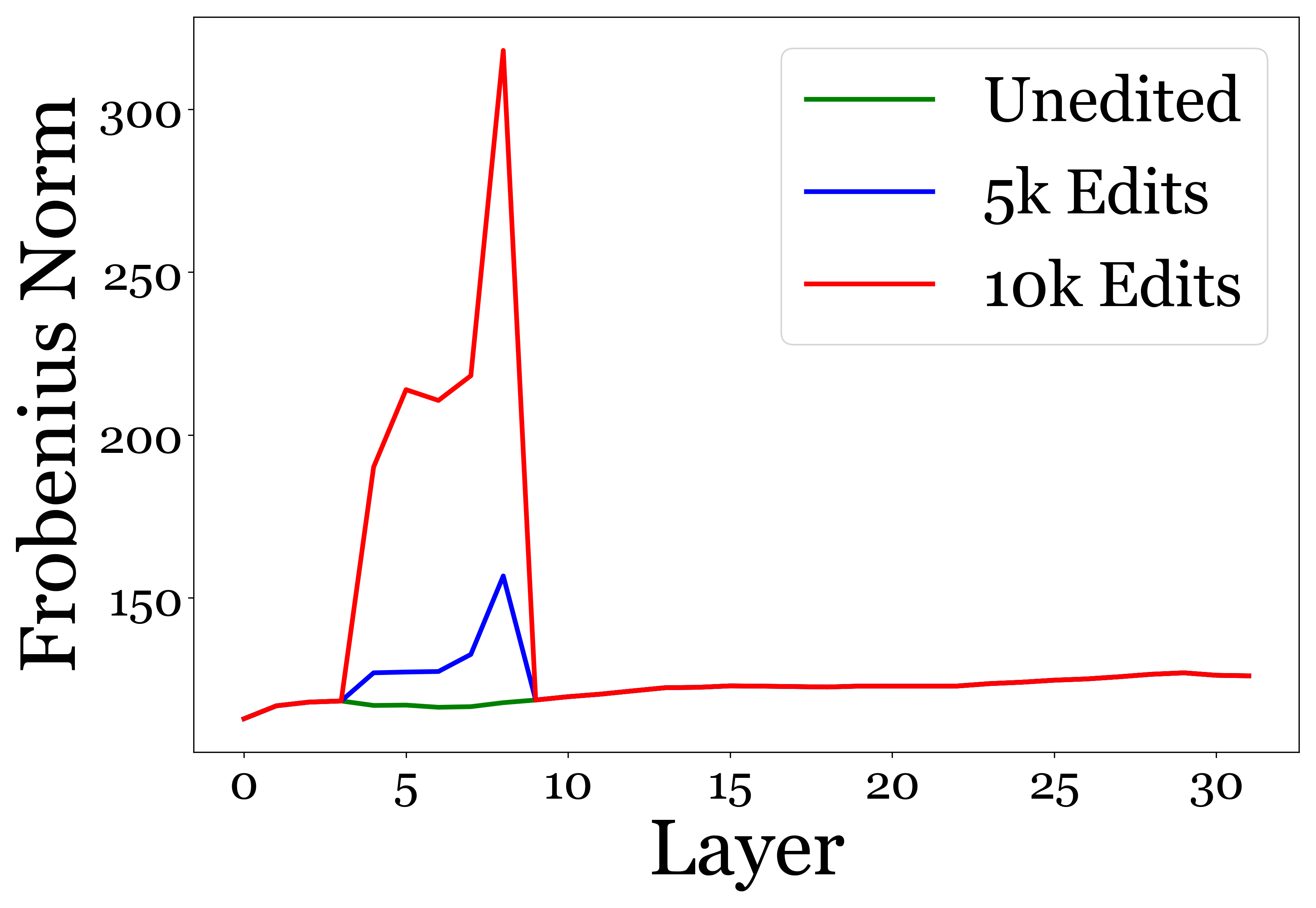} 
        \label{fig:norm-growth-llama2-7b-memit}
    }
    \caption{Norm growth of Llama2-7B using AlphaEdit and MEMIT}
    \label{fig:norm-growth-llama2-7b}
\end{figure}

\begin{figure}[htbp]
    \centering

    \subfigure[Norm growth of Llama2-7B using NC]{
        \includegraphics[width=0.45\linewidth]{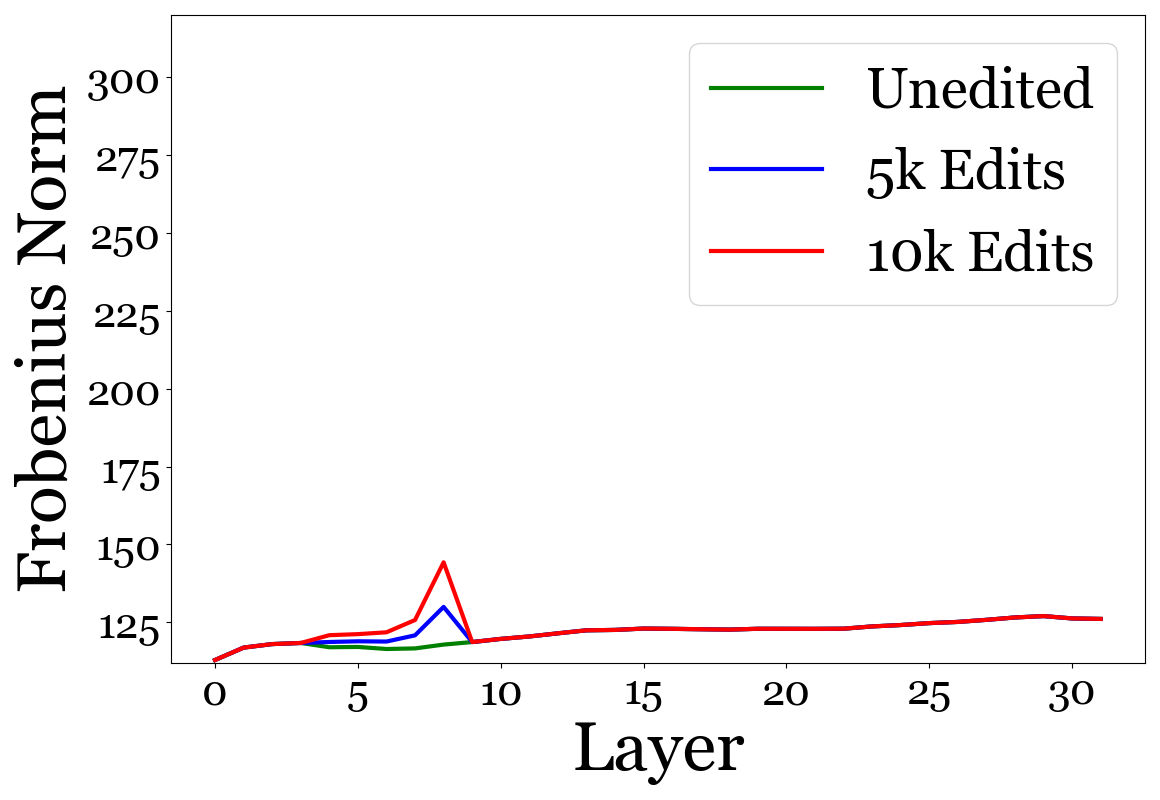} 
        \label{fig:norm-growth-llama2-7b-norm}
    }
    \subfigure[Norm growth of Llama2-7B using MPES + NC]{
        \includegraphics[width=0.45\linewidth]{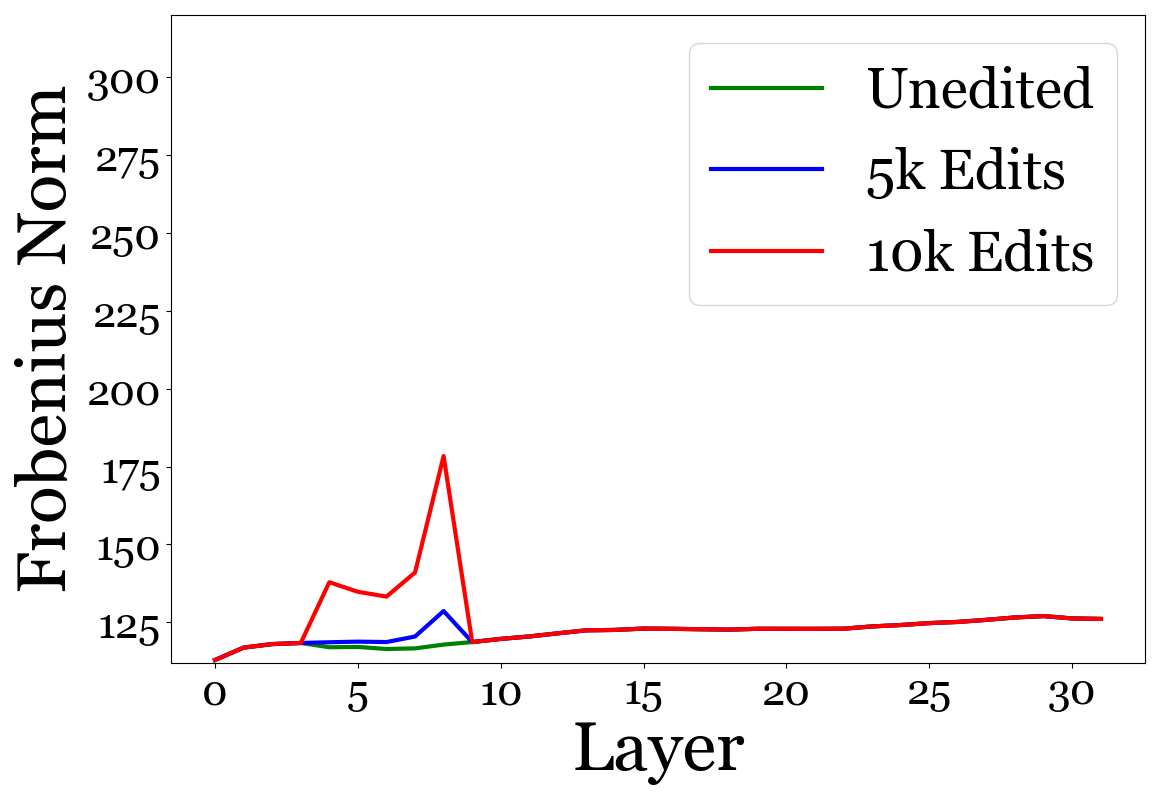} 
        \label{fig:norm-growth-llama2-7b-encore-sub}
    }
    \caption{Norm growth of Llama2-7B using NC and MPES + NC}
    \label{fig:norm-growth-llama2-7b-encore}
\end{figure}

\begin{figure}[htbp]
    \centering

    \subfigure[Norm growth of Llama3-8B using Alphaedit]{
        \includegraphics[width=0.45\linewidth]{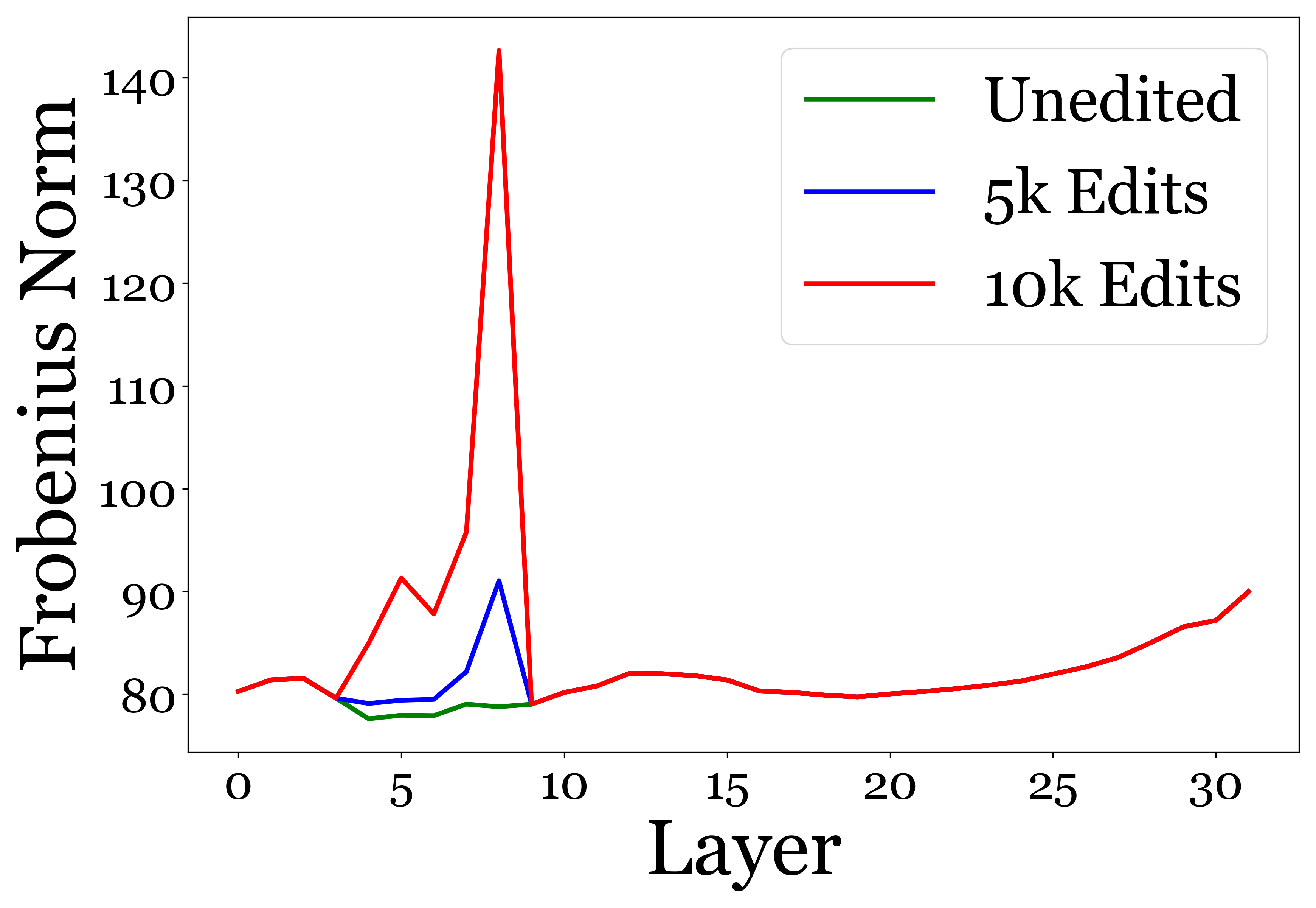} 
        \label{fig:norm-growth-llama3-8b-alphaedit}
    }
    \subfigure[Norm growth of Llama3-8B using MEMIT]{
        \includegraphics[width=0.45\linewidth]{figures/norm-growth/llama-3-8b_memit.png} 
        \label{fig:norm-growth-llama3-8b-MEMIT}
    }
    \caption{Norm growth of Llama3-8B across different methods}
    \label{fig:norm-growth-llama3-8b}
\end{figure}

\begin{figure}[htbp]
    \centering
    \subfigure[Norm growth of Llama3-8B using NC]{
        \includegraphics[width=0.45\linewidth]{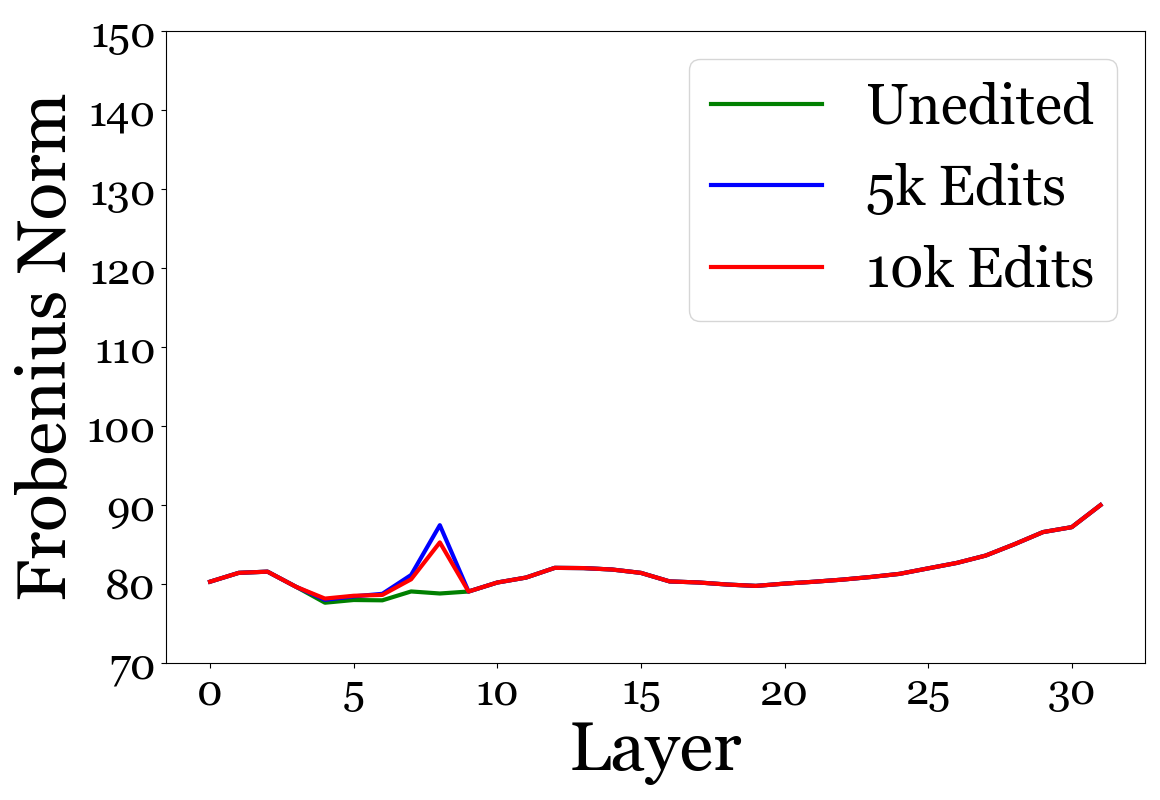} 
        \label{fig:norm-growth-llama3-8b-norm}
    }
    \subfigure[Norm growth of Llama3-8B using MPES + NC]{
        \includegraphics[width=0.45\linewidth]{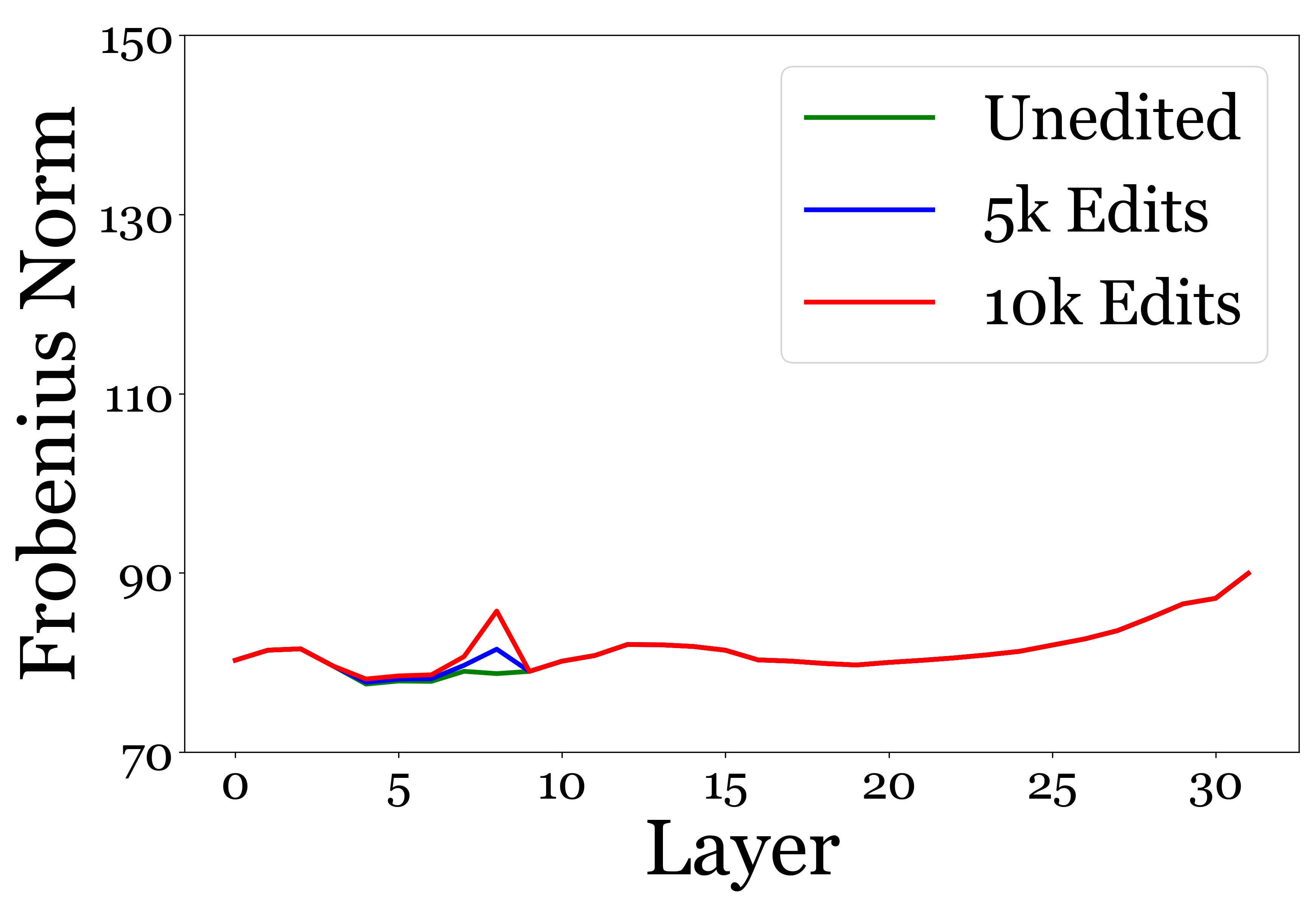} 
        \label{fig:norm-growth-llama3-8b-encore-sub}
    }
    \caption{Norm growth of Llama3-8B using NC and MPES + NC}
    \label{fig:norm-growth-llama3-8b-encore}
\end{figure}

In this section we present the observation of norm growth during editing for different methods (including Norm Constraint and MPES + Norm Constraint) with different models. Figures \ref{fig:norm-growth-gpt2-xl} and \ref{fig:norm-growth-gpt2-xl-encore} show the result for different methods for GPT2-XL. Figures \ref{fig:norm-growth-llama2-7b} and \ref{fig:norm-growth-llama2-7b-encore} show the result for Llama2-7B. Lastly figures \ref{fig:norm-growth-llama3-8b} and \ref{fig:norm-growth-llama3-8b-encore} show the result for Llama3-8B.

\newpage

\section{Editing Performance on zsRE dataset}
Tables \ref{tab:editing-performance-zsre-gpt2xl} - \ref{tab:editing-performance-zsre-llama3} show the editing scores for the sequential editing experiments on zsRE.

\begin{table}[htbp]
\vskip 0.15in
\begin{center}
\begin{adjustbox}{max width=0.5\textwidth}
\begin{sc}
\begin{tabular}{lccccccr}
\toprule
Model & Method  & \multirow{2}{*}{\makecell{Edit \\ Score}} & \multirow{2}{*}{\makecell{Paraphrase \\ Score}} & \multirow{2}{*}{\makecell{Neighborhood \\ Score}} \\
& & & \\
\midrule
GPT2-XL 
& AlphaEdit    & 42.1 & 33.61 & 14.61  \\
& AlphaEdit + MPES    & 54.99 & 43.18 & 18.40 \\
& MEMIT    & 74.60 & 61.77 & 22.40  \\
& MEMIT + MPES   & \textbf{75.09} & 61.58 & 23.37  \\
& MEMIT + NC &  74.51 & \textbf{61.90} & 23.39 \\
& MEMIT + MPES + NC     &   74.46 & 61.79 & \textbf{23.41}  \\
\bottomrule
\end{tabular}
\end{sc}
\end{adjustbox}
\end{center}
\vskip -0.1in
\caption{Editing performance for GPT2-XL on zsre dataset}
\label{tab:editing-performance-zsre-gpt2xl}
\end{table}

\begin{table}[htbp]
\vskip 0.15in
\begin{center}
\begin{adjustbox}{max width=0.5\textwidth}
\begin{sc}
\begin{tabular}{lccccccr}
\toprule
Model & Method  & \multirow{2}{*}{\makecell{Edit \\ Score}} & \multirow{2}{*}{\makecell{Paraphrase \\ Score}} & \multirow{2}{*}{\makecell{Neighborhood \\ Score}} \\
& & & \\
\midrule
Llama2-7B 
& AlphaEdit    & 83.77 & 77.12 & 41.96  \\
& AlphaEdit + MPES    & 83.80 & 77.64 & 41.97 \\
& MEMIT    & 79.49 & 74.29 & 41.80  \\
& MEMIT + MPES   & 83.01 & 77.45 & 44.64  \\
& MEMIT + NC  &  88.73 & 84.05 & 47.98 \\
& MEMIT + MPES + NC     &   \textbf{89.10} & \textbf{84.28} & \textbf{48.51}  \\
\bottomrule
\end{tabular}
\end{sc}
\end{adjustbox}
\end{center}
\vskip -0.1in
\caption{Editing performance for Llama2-7B on zsre dataset}
\label{tab:editing-performance-zsre-llama2}
\end{table}

\begin{table}[htbp]
\vskip 0.15in
\begin{center}
\begin{adjustbox}{max width=0.5\textwidth}
\begin{sc}
\begin{tabular}{lccccccr}
\toprule
Model & Method  & \multirow{2}{*}{\makecell{Edit \\ Score}} & \multirow{2}{*}{\makecell{Paraphrase \\ Score}} & \multirow{2}{*}{\makecell{Neighborhood \\ Score}} \\
& & & \\
\midrule
Llama3-8B 
& AlphaEdit    & 89.27 & 82.19 & 45.23  \\
& AlphaEdit + MPES    & 93.54 & 85.93 & 47.32 \\
& MEMIT    & 96.45 & 90.30 & 48.91  \\
& MEMIT + MPES   & \textbf{96.85} & \textbf{90.76} & 47.34  \\
& MEMIT + NC  & 90.40  & 84.58 & 49.09 \\
& MEMIT + MPES + NC     & 93.15   & 86.19 & \textbf{49.81}  \\
\bottomrule
\end{tabular}
\end{sc}
\end{adjustbox}
\end{center}
\vskip -0.1in
\caption{Editing performance for Llama3-8B on zsre dataset}
\label{tab:editing-performance-zsre-llama3}
\end{table}

\section{Downstream Performance}\label{appendix:downstream-performance}

\begin{figure}[ht]
    \centering
    \subfigure[AlphaEdit and MPES]{
        \includegraphics[width=0.6\linewidth]{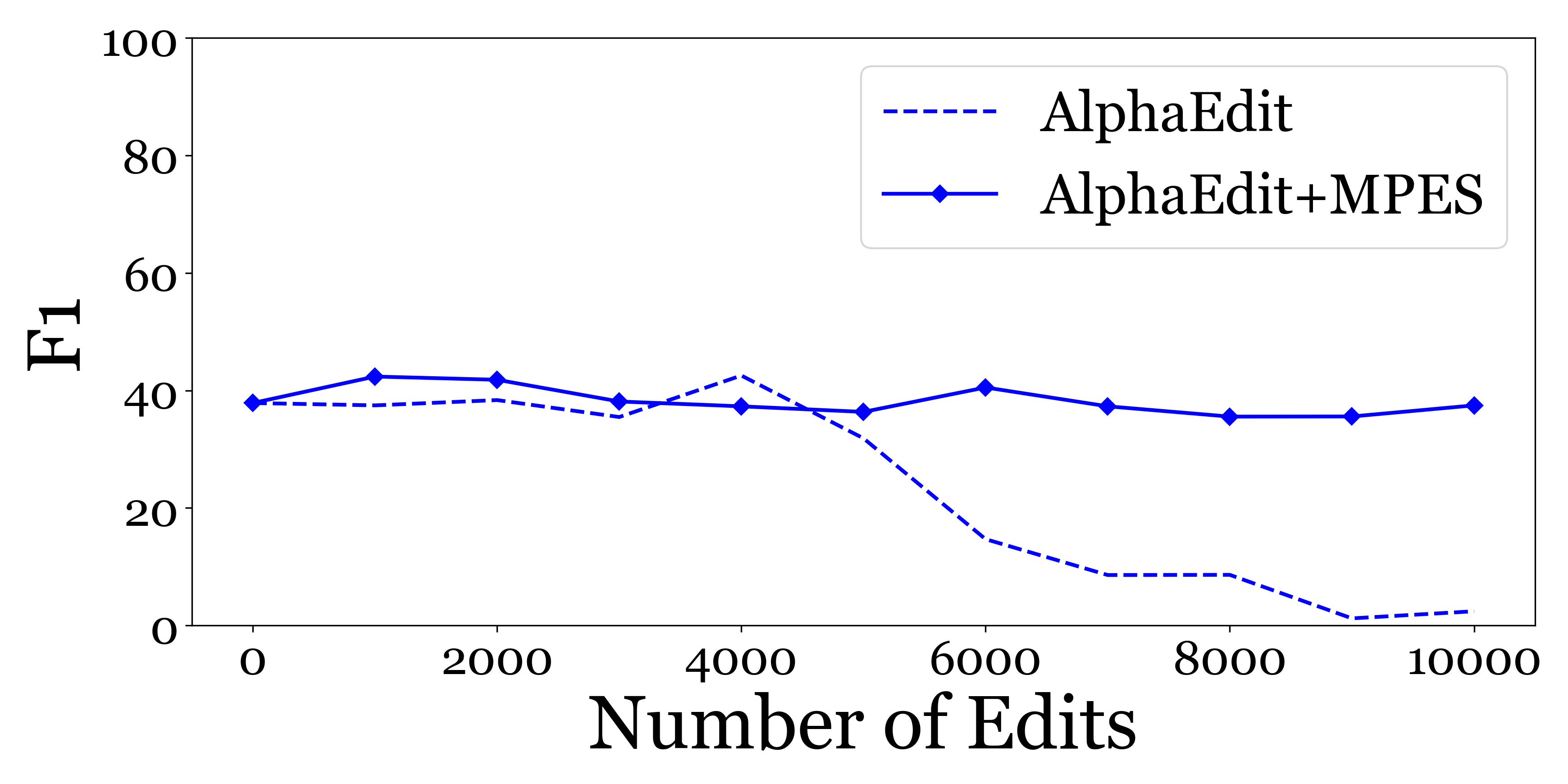}
        \label{fig:downstream-gpt2-xl-alphaedit}
    }
    \subfigure[MEMIT, MPES, NC, MPES+NC]{
        \includegraphics[width=0.6\linewidth]{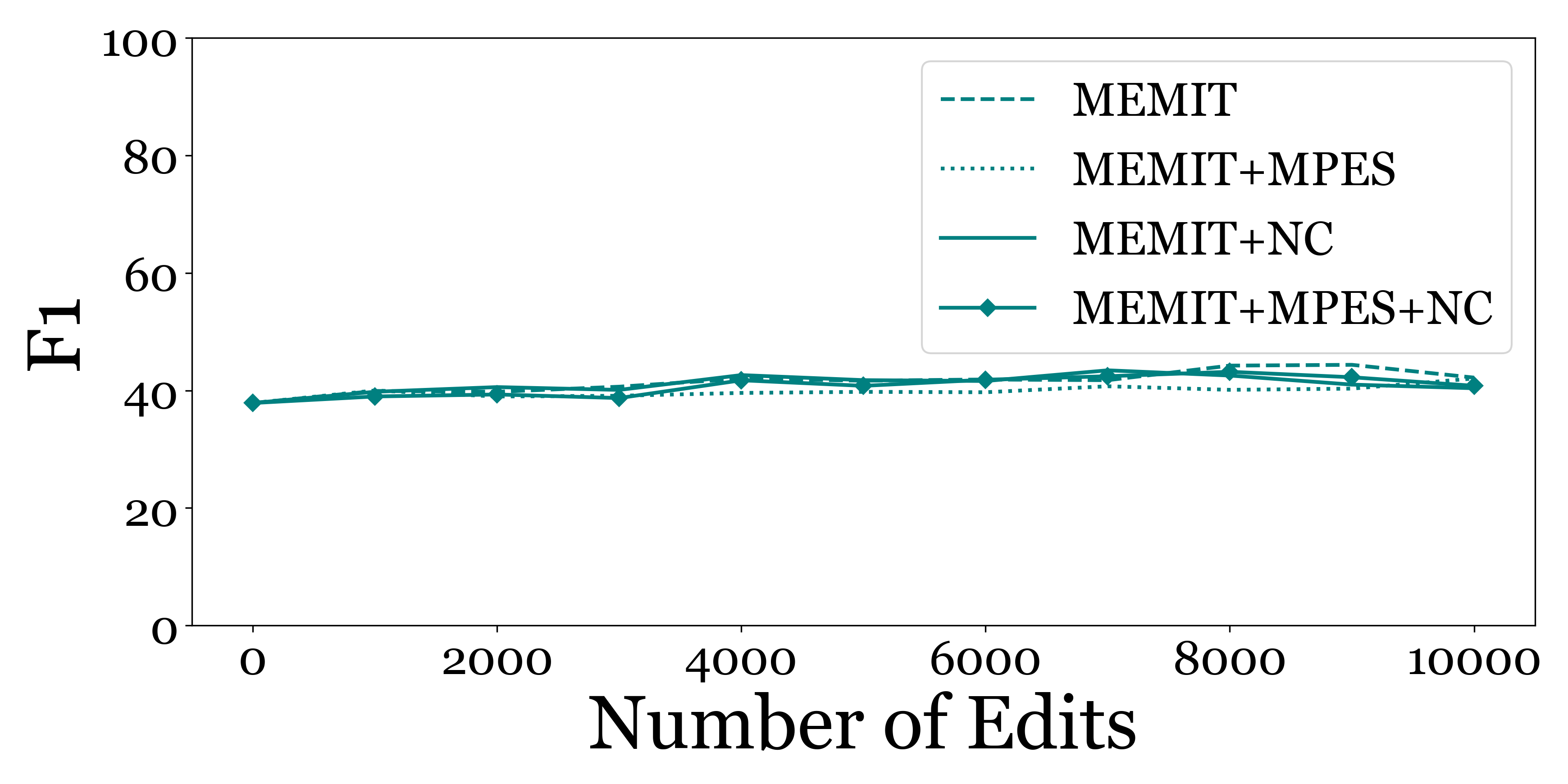}
        \label{fig:downstream-gpt2-xl-memit}
    }
    \caption{Downstream Performance for GPT2-XL using different editing methods with CounterFact dataset}
    \label{fig:downstream-gpt2-xl-comparison}
\end{figure}

\begin{figure}[ht]
    \centering
    \subfigure[AlphaEdit and MPES]{
        \includegraphics[width=0.6\linewidth]{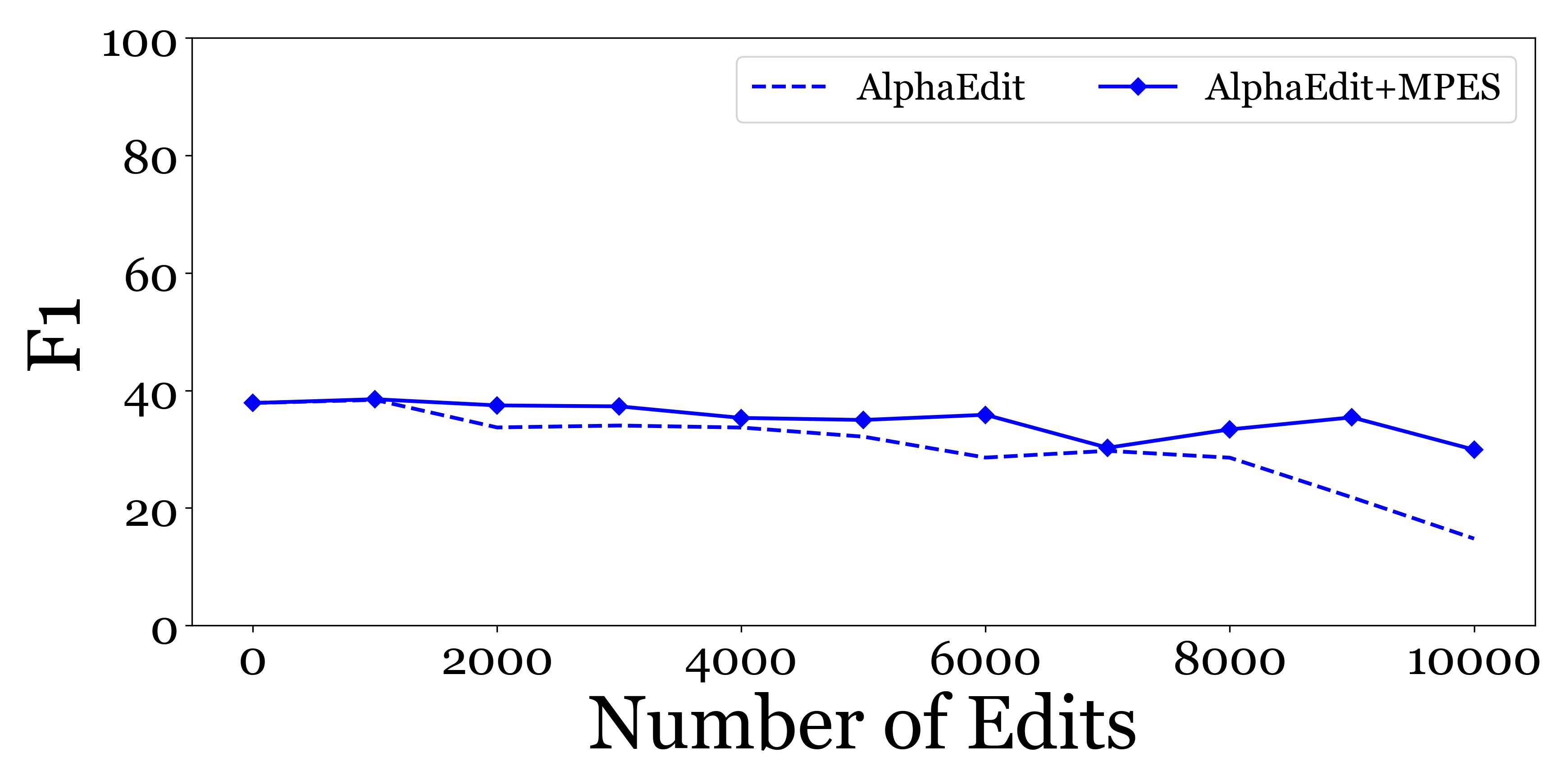}
        \label{fig:downstream-gpt2-xl-alphaedit-zsre}
    }
    \subfigure[MEMIT, MPES, NC, MPES+NC]{
        \includegraphics[width=0.6\linewidth]{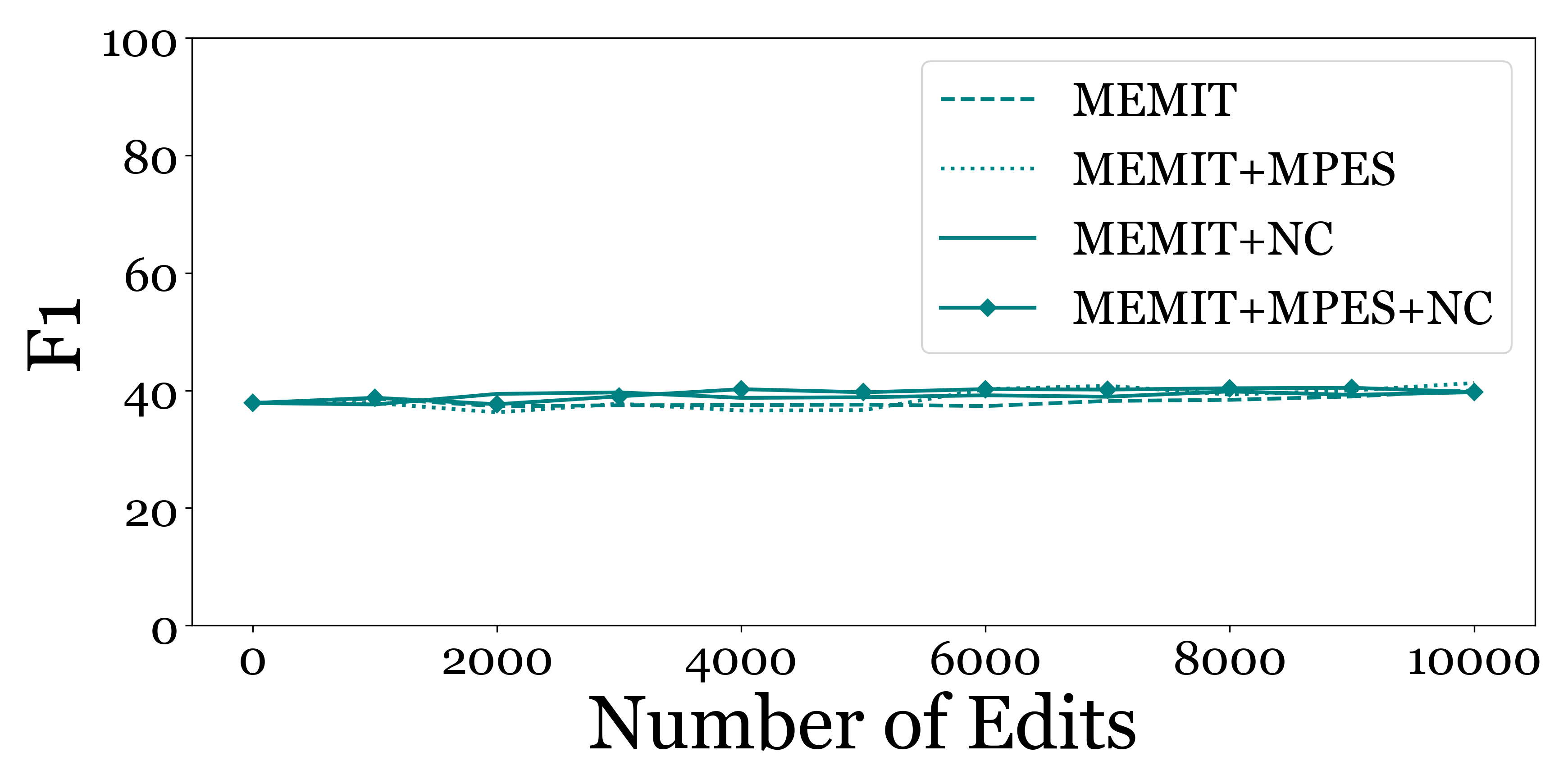}
        \label{fig:downstream-gpt2-xl-memit-zsre}
    }
    \caption{Downstream Performance for GPT2-XL using different editing methods with zsRE dataset}
    \label{fig:downstream-gpt2-xl-comparison-zsre}
\end{figure}

\begin{figure}[ht]
    \centering
    \subfigure[AlphaEdit and MPES]{
        \includegraphics[width=0.6\linewidth]{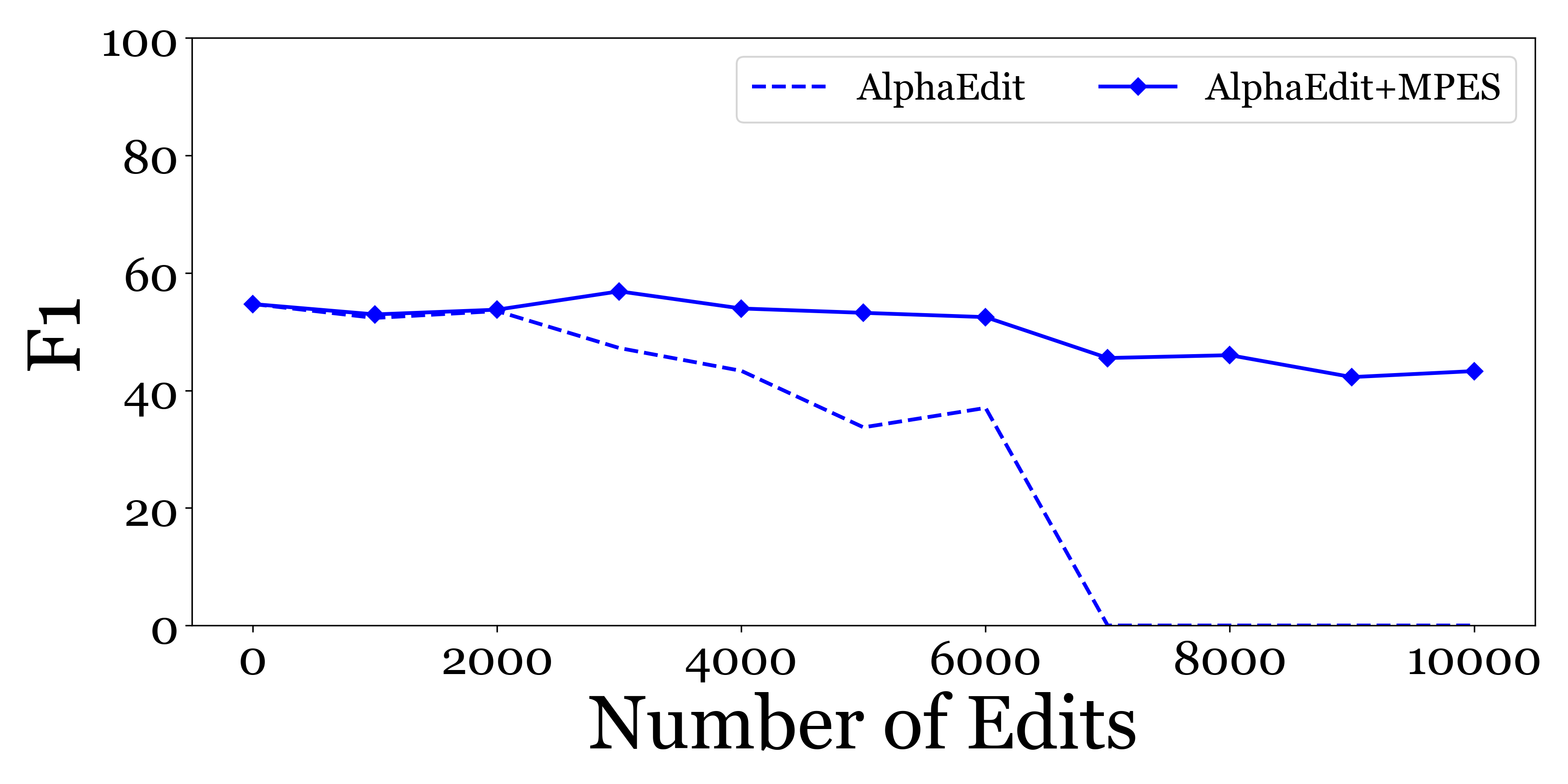}
        \label{fig:downstream-llama2-alphaedit}
    }
    \subfigure[MEMIT, MPES, NC, MPES+NC]{
        \includegraphics[width=0.6\linewidth]{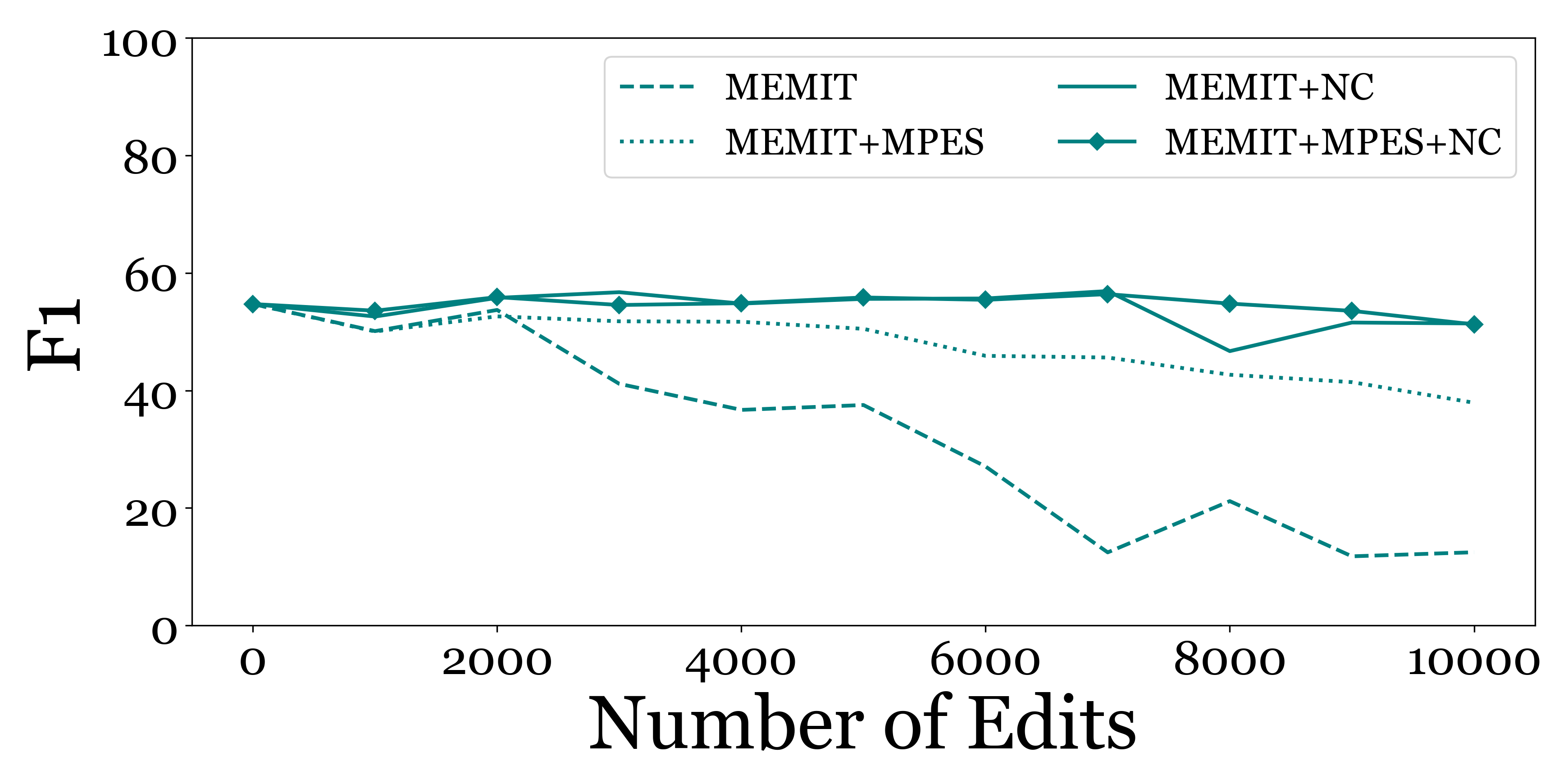}
        \label{fig:downstream-llama2-memit}
    }
    \caption{Downstream Performance for Llama2-7B using different editing methods with CounterFact dataset}
    \label{fig:downstream-llama2-7b-comparison}
\end{figure}

\begin{figure}[ht]
    \centering
    \subfigure[AlphaEdit and MPES]{
        \includegraphics[width=0.6\linewidth]{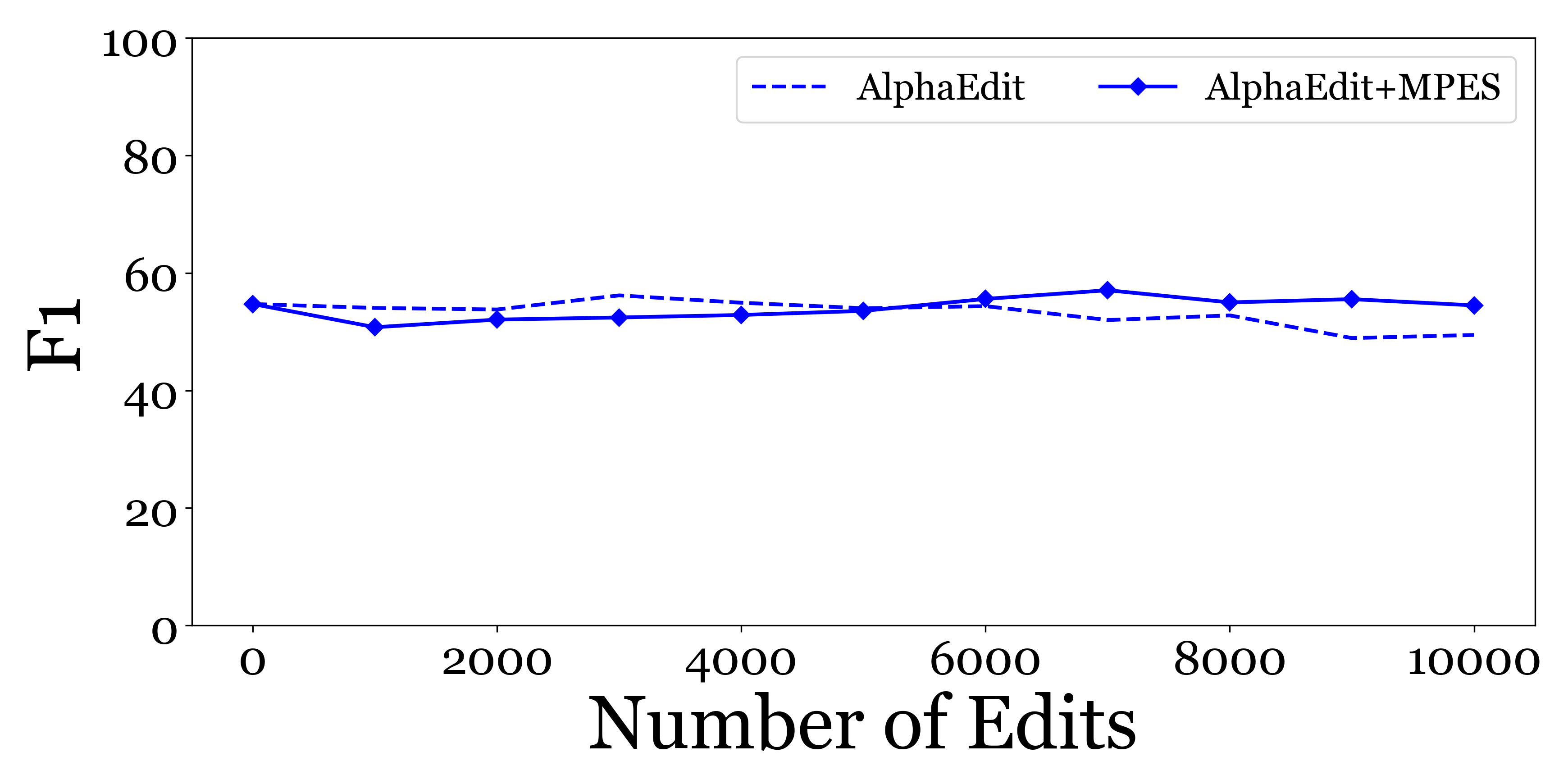}
        \label{fig:downstream-llama2-alphaedit-zsre}
    }
    \subfigure[MEMIT, MPES, NC, MPES+NC]{
        \includegraphics[width=0.6\linewidth]{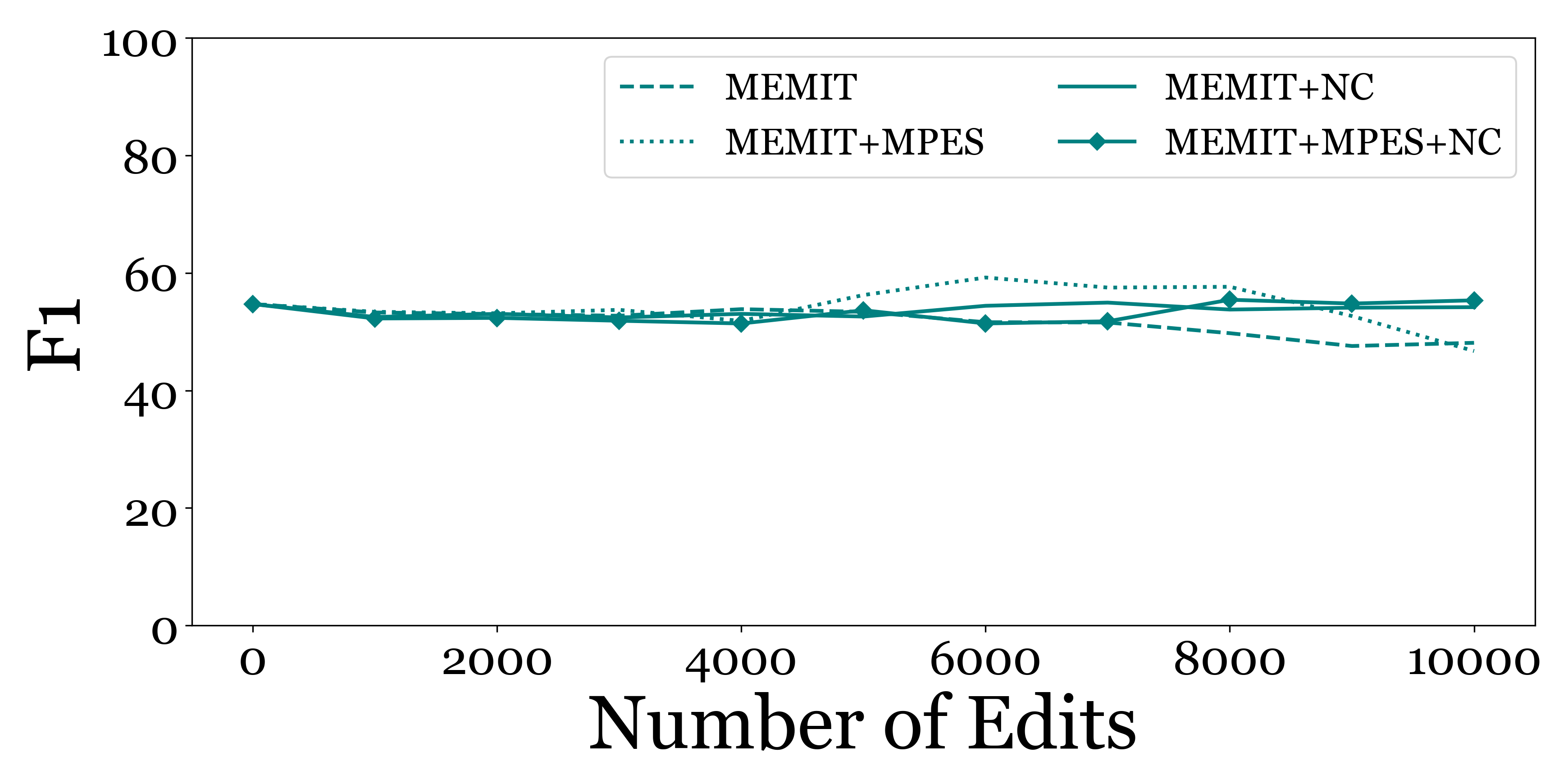}
        \label{fig:downstream-llama2-memit-zsre}
    }
    \caption{Downstream Performance for Llama2-7B using different editing methods with zsRE dataset}
    \label{fig:downstream-llama2-7b-comparison-zsre}
\end{figure}

\begin{figure}[ht]
    \centering
    \subfigure[AlphaEdit and MPES]{
        \includegraphics[width=0.6\linewidth]{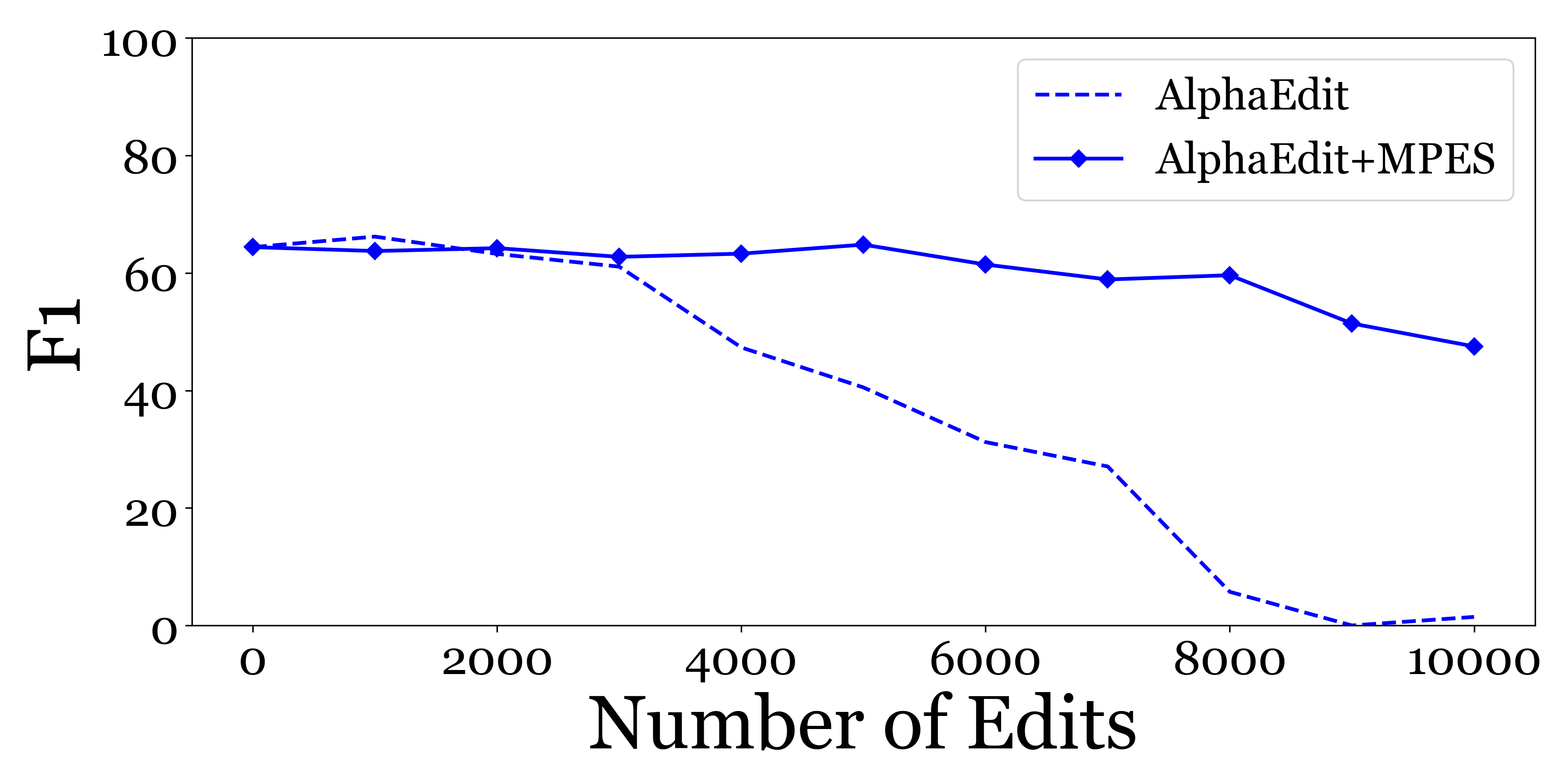}
        \label{fig:downstream-llama3-alphaedit}
    }
    \subfigure[MEMIT, MPES, NC, MPES+NC]{
        \includegraphics[width=0.6\linewidth]{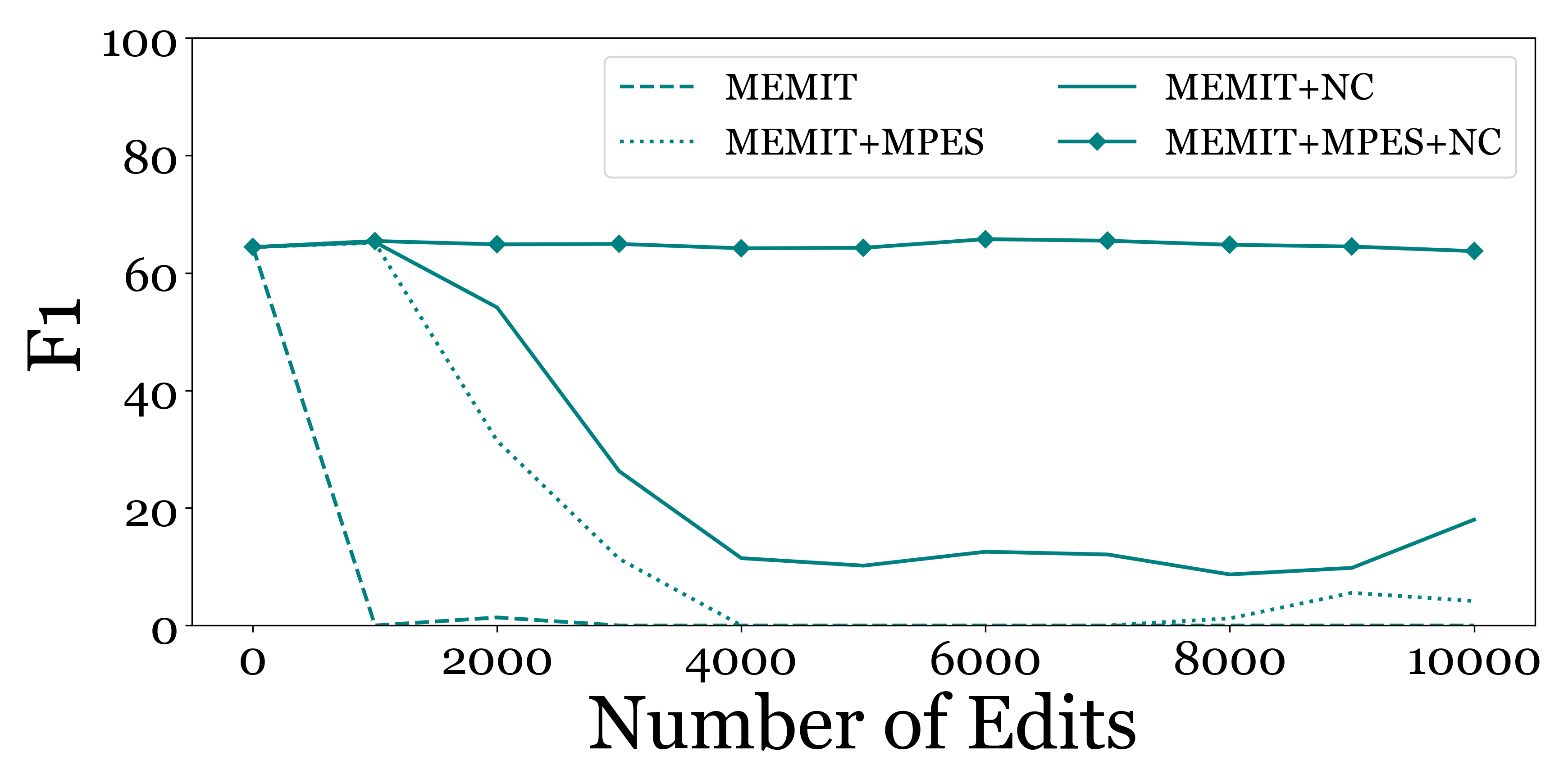}
        \label{fig:downstream-llama3-memit}
    }
    \caption{Downstream Performance for Llama3-8B using different editing methods with CounterFact dataset}
    \label{fig:downstream-llama3-8b-comparison}
\end{figure}

\begin{figure}[ht]
    \centering
    \subfigure[AlphaEdit and MPES]{
        \includegraphics[width=0.6\linewidth]{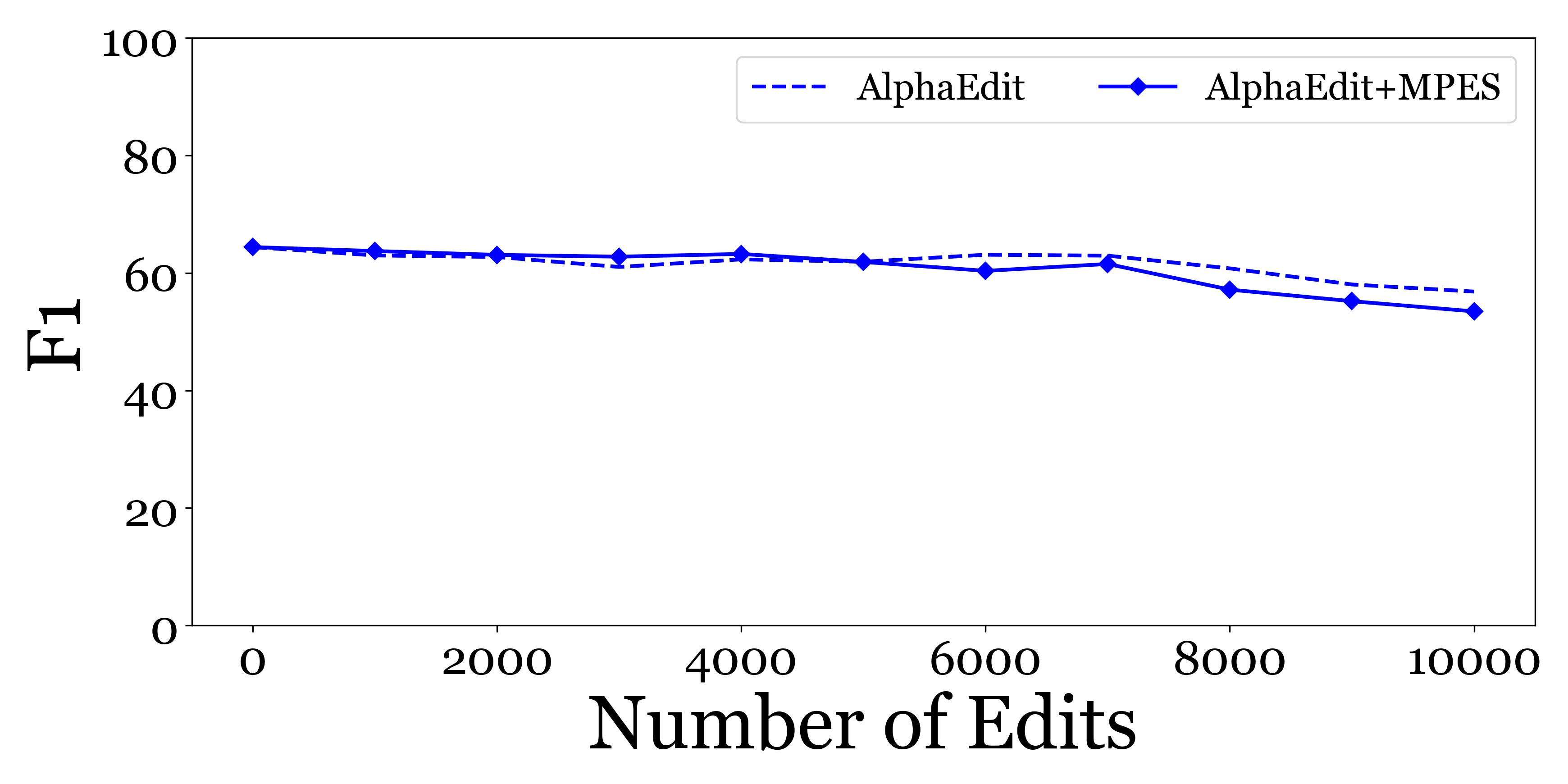}
        \label{fig:downstream-llama3-alphaedit-zsre}
    }
    \subfigure[MEMIT, MPES, NC, MPES+NC]{
        \includegraphics[width=0.6\linewidth]{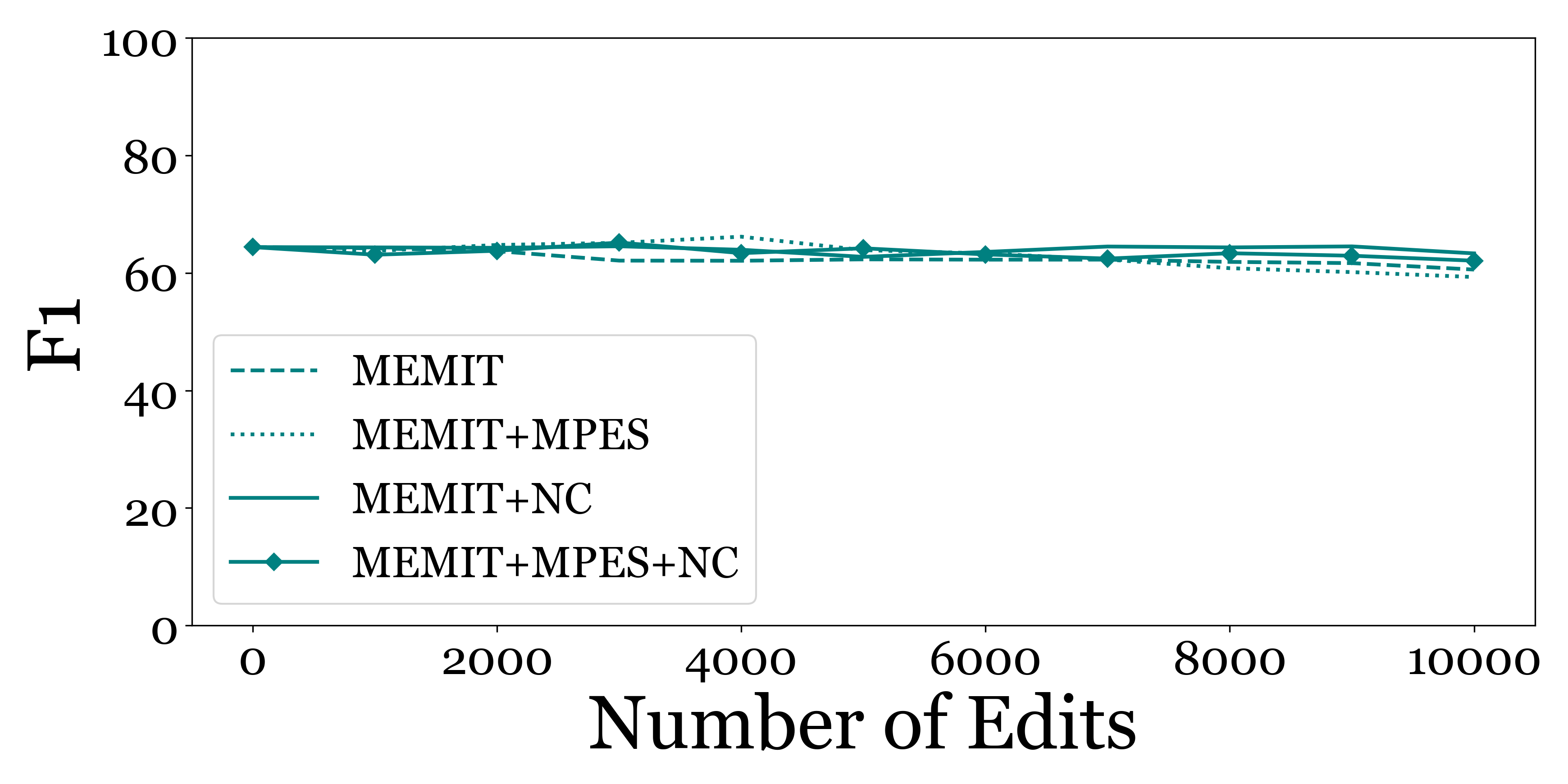}
        \label{fig:downstream-llama3-memit-zsre}
    }
    \caption{Downstream Performance for Llama3-8B using different editing methods with zsRE dataset}
    \label{fig:downstream-llama3-8b-comparison-zsre}
\end{figure}

Figures \ref{fig:downstream-gpt2-xl-alphaedit} - \ref{fig:downstream-gpt2-xl-memit-zsre} show the result for the downstream performance for GPT2-XL on both CounterFact and zsRE datasets. Figures \ref{fig:downstream-llama2-alphaedit} - \ref{fig:downstream-llama2-memit-zsre} show the result for Llama2-7B on both datasets.
Lastly, figures \ref{fig:downstream-llama3-alphaedit} - \ref{fig:downstream-llama3-memit-zsre} show the result for Llama3-8B on both datasets.

\section{Norm Decrease From MPES}
\begin{figure}[ht]
    \centering

    \includegraphics[width=0.7\linewidth]{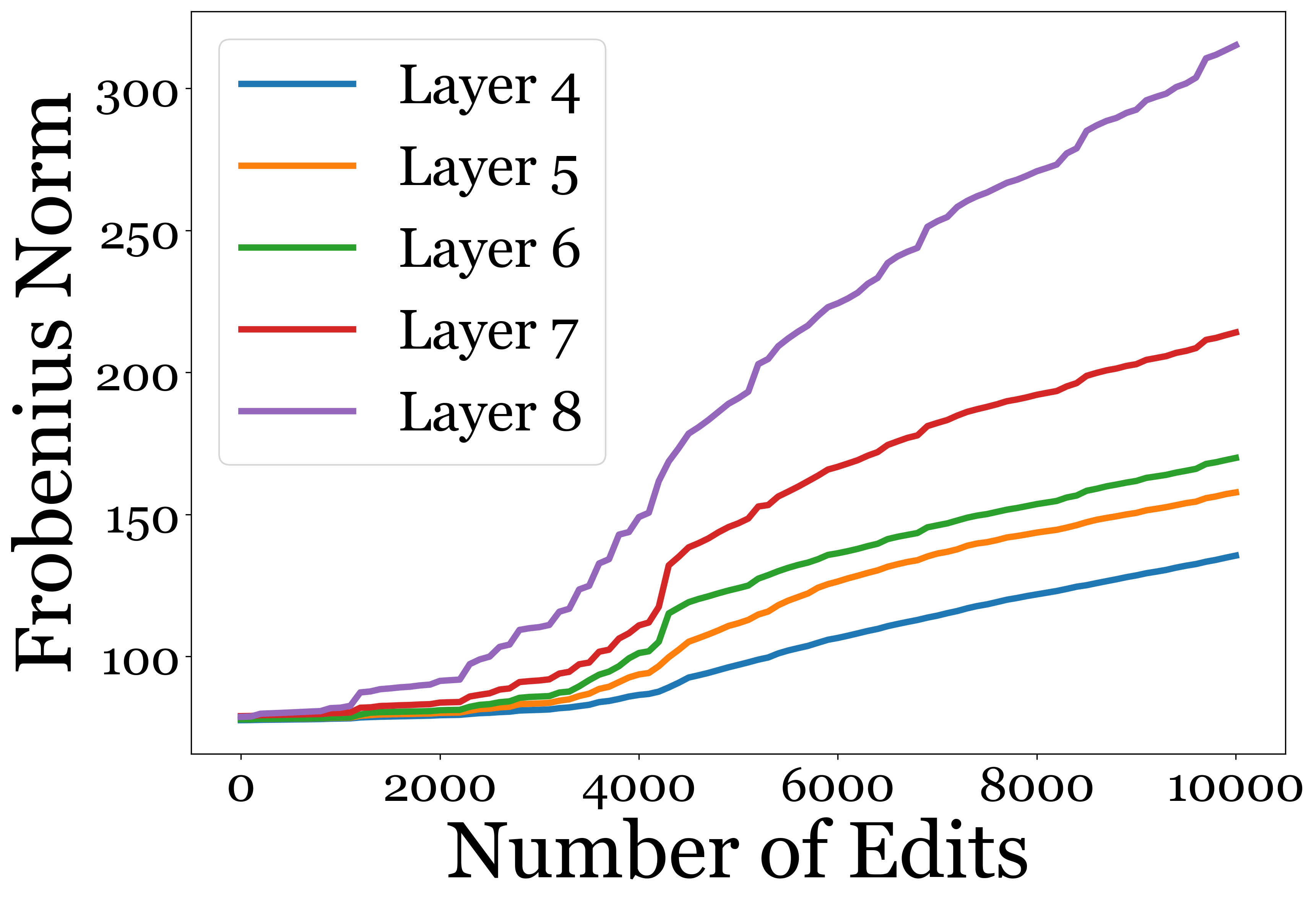}
    \caption{The figure shows the Norm-growth as function of number edits in MEMIT Llama3-8B with MPES.}
    \label{fig:memit-mpes-llama3}
\end{figure}

In this section we want to highlight one observation that we have is that while MPES does reduce the norm of the edited layer shown in Figure \ref{fig:memit-mpes-llama3} by quite a lot compared to just MEMIT alone in Figure \ref{fig:norm-growth-MEMIT}. But we still note that the norm growth is still 3 times higher which is the reason why we think that explicit norm control is still needed. 

\section{Unbound growth of norm in AlphaEdit}\label{appendix:unbound-growth}
Here we provide the result if we remove the norm constraint from the objective function of AlphaEdit. As we can see, once we remove the norm constraint, the norm growth of AlphaEdit just becomes unbound. Note that the reason why the x-axis only goes up to 300 edits is that after that the norm value becomes overflow in Python and it therefore it cannot be plot. 

\begin{figure}[ht]
    \centering
    \subfigure[AlphaEdit with Llama2-7B]{
        \includegraphics[width=0.8\linewidth]{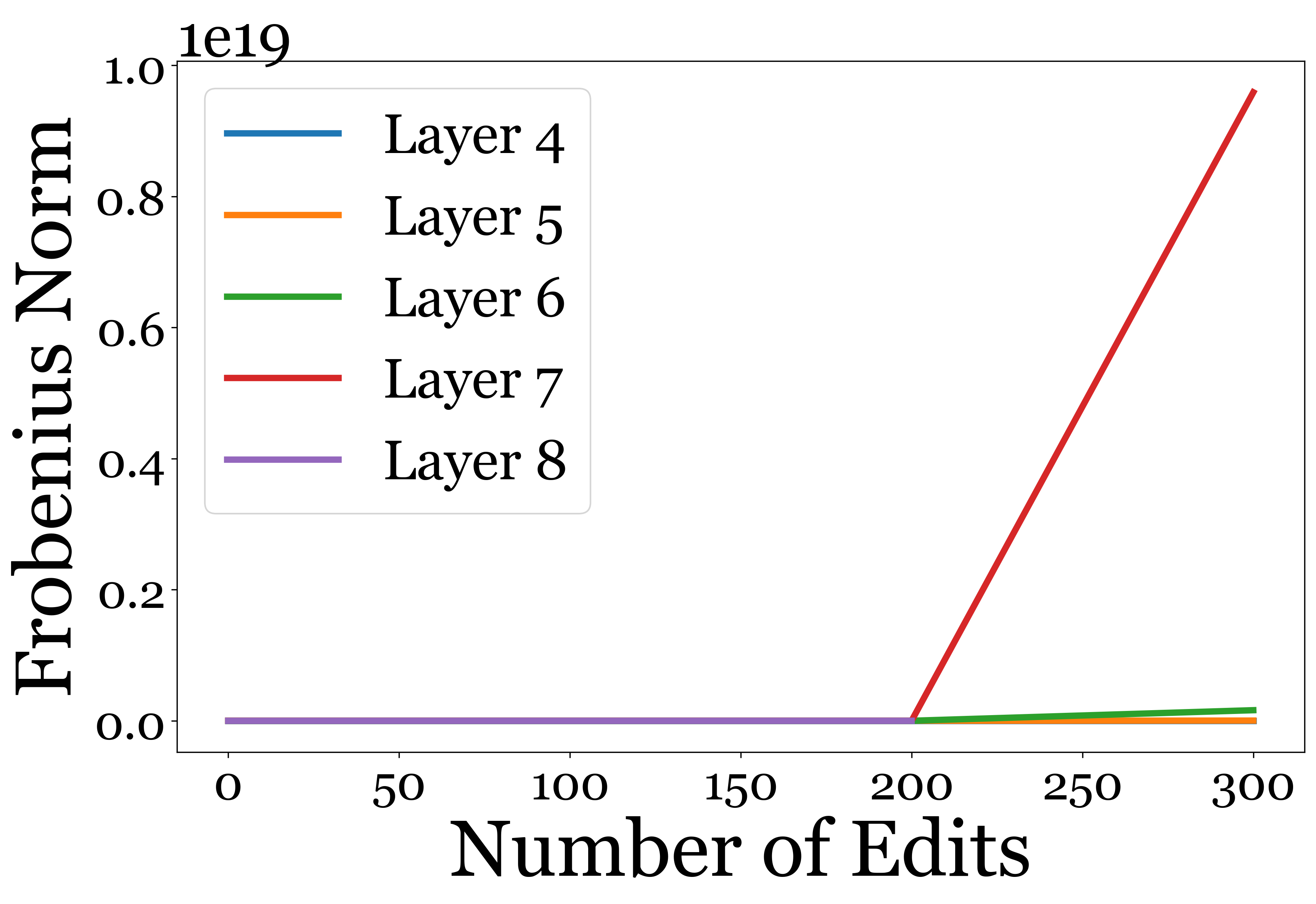}
        \label{fig:alphaedit-llama2-without-norm}
    }
    \subfigure[AlphaEdit with Llama3-8B]{
        \includegraphics[width=0.8\linewidth]{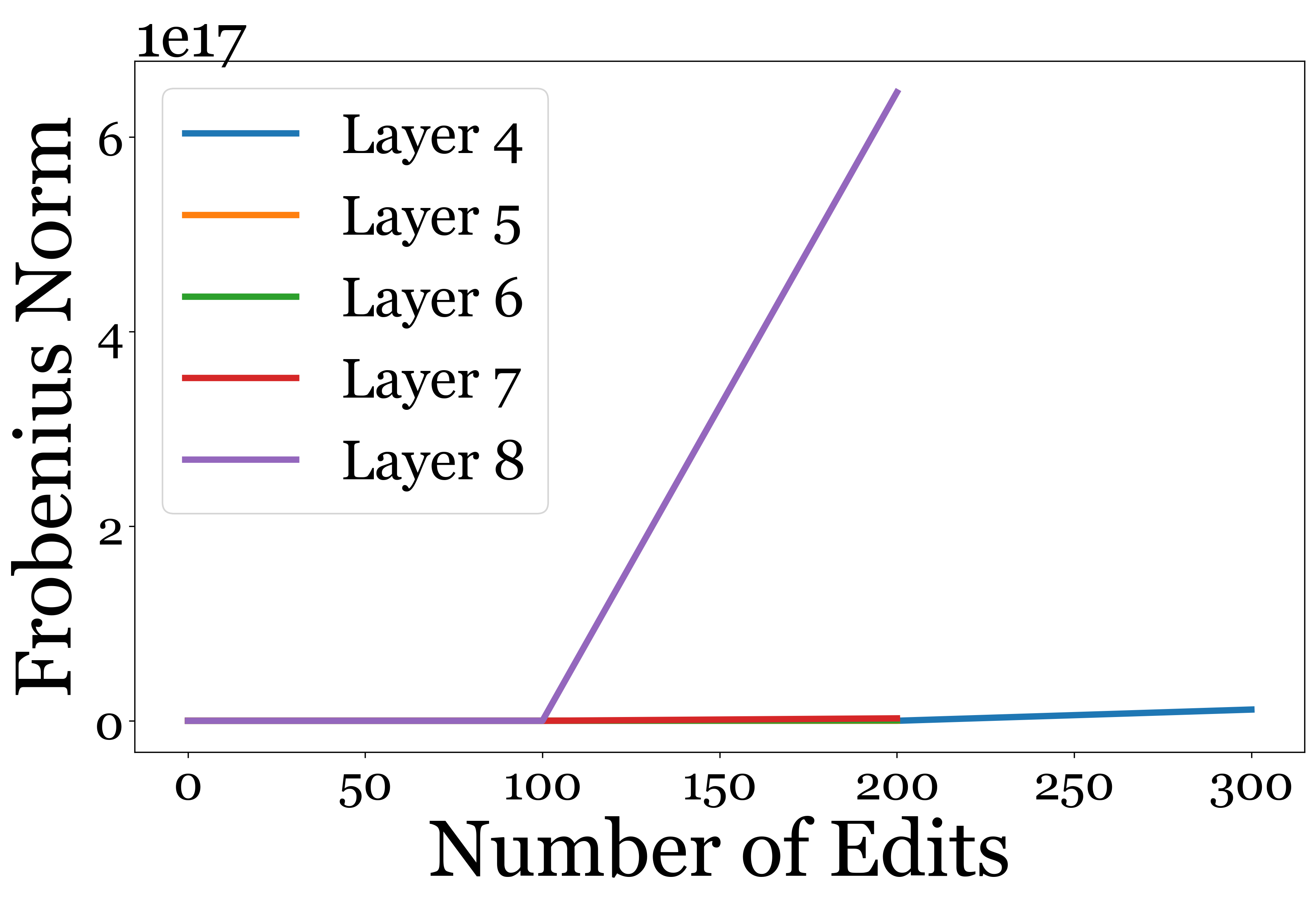}
        \label{fig:alphaedit-llama3-without-norm}
    }
    \caption{The figures show the unbound Norm-growth as function of number edits in AlphaEdit}
    \label{fig:alphaedit-without-norm}
\end{figure}

\section{Batch Size Effect}
For this section we show the result for different batch size we see that our method MPES + NC is still effective for all batch size and for all model 

\begin{table*}[t]
\vskip 0.1in
\begin{center}
\setlength{\tabcolsep}{4pt}
\begin{adjustbox}{max width=\textwidth}
\begin{tabular}{l ccc ccc ccc ccc ccc}
\toprule
\textbf{Method} & 
\multicolumn{3}{c}{\textbf{Edit Score}} & 
\multicolumn{3}{c}{\textbf{Paraphrase Score}} & 
\multicolumn{3}{c}{\textbf{Neighborhood Score}} & 
\multicolumn{3}{c}{\textbf{Overall Score}} & 
\multicolumn{3}{c}{\textbf{Generation Entropy}} \\
\cmidrule(lr){2-4} \cmidrule(lr){5-7} \cmidrule(lr){8-10} \cmidrule(lr){11-13} \cmidrule(lr){14-16}
& GPT2-XL & Llama2-7B & Llama3-8B & GPT2-XL & Llama2-7B & Llama3-8B & GPT2-XL & Llama2-7B & Llama3-8B & GPT2-XL & Llama2-7B & Llama3-8B & GPT2-XL & Llama2-7B & Llama3-8B \\
\midrule
MEMIT                         & 91.97 & 81.73 & 50.32 & 77.67 & 65.58 & 49.56 & 58.00 & \textbf{66.22} & 50.59 &73.19 & 70.45 & 50.15 & 503.98 & 522.72 & 281.88  \\
PRUNE                         & \textbf{92.76} & 51.80 & 48.50 & \textbf{78.43} & 50.15 & 49.21 & 57.25 & 50.17 & 51.15 & 73.17 & 50.70 & 49.60 & 505.02 & 247.88 & 297.80 \\
 RECT                          & 51.01 & 51.85 & 48.94 & 48.92 & 50.47 & 49.98 & 52.57 & 49.34 & 50.14 & 50.79 & 50.53 & 49.68 & 539.58 & 323.46 & 192.70\\
AlphaEdit                     & 91.24 & 56.92 & 57.80 & 73.71 & 51.38 & 56.55 & 56.39 & 53.73 & 49.70 & 70.99 & 53.92 & 54.44 & \textbf{586.77} & 494.64 & 430.37  \\
MEMIT + MPES + NC     & 92.57 & \textbf{90.79} & \textbf{89.71} & 78.19 & \textbf{79.76} & \textbf{80.48} & \textbf{60.36} & 59.70 & \textbf{57.63} & \textbf{74.70} & \textbf{74.44} & \textbf{73.31} & 510.81 & \textbf{555.18} & \textbf{539.91} \\
\bottomrule
\end{tabular}
\end{adjustbox}
\end{center}
\caption{\centering Editing performance of our approach when compared to baseline MEMIT, AlphaEdit and MEMIT regularization method such as PRUNE and RECT using batchsize 10. }\label{tab:editing-performance-encore-main-batchsize-10}
\vskip -0.1in
\end{table*}

\begin{figure*}[ht]
    \centering
    \subfigure[Downstream Performance for GPT2-XL with batch size 10]{
        \includegraphics[width=0.3\linewidth]{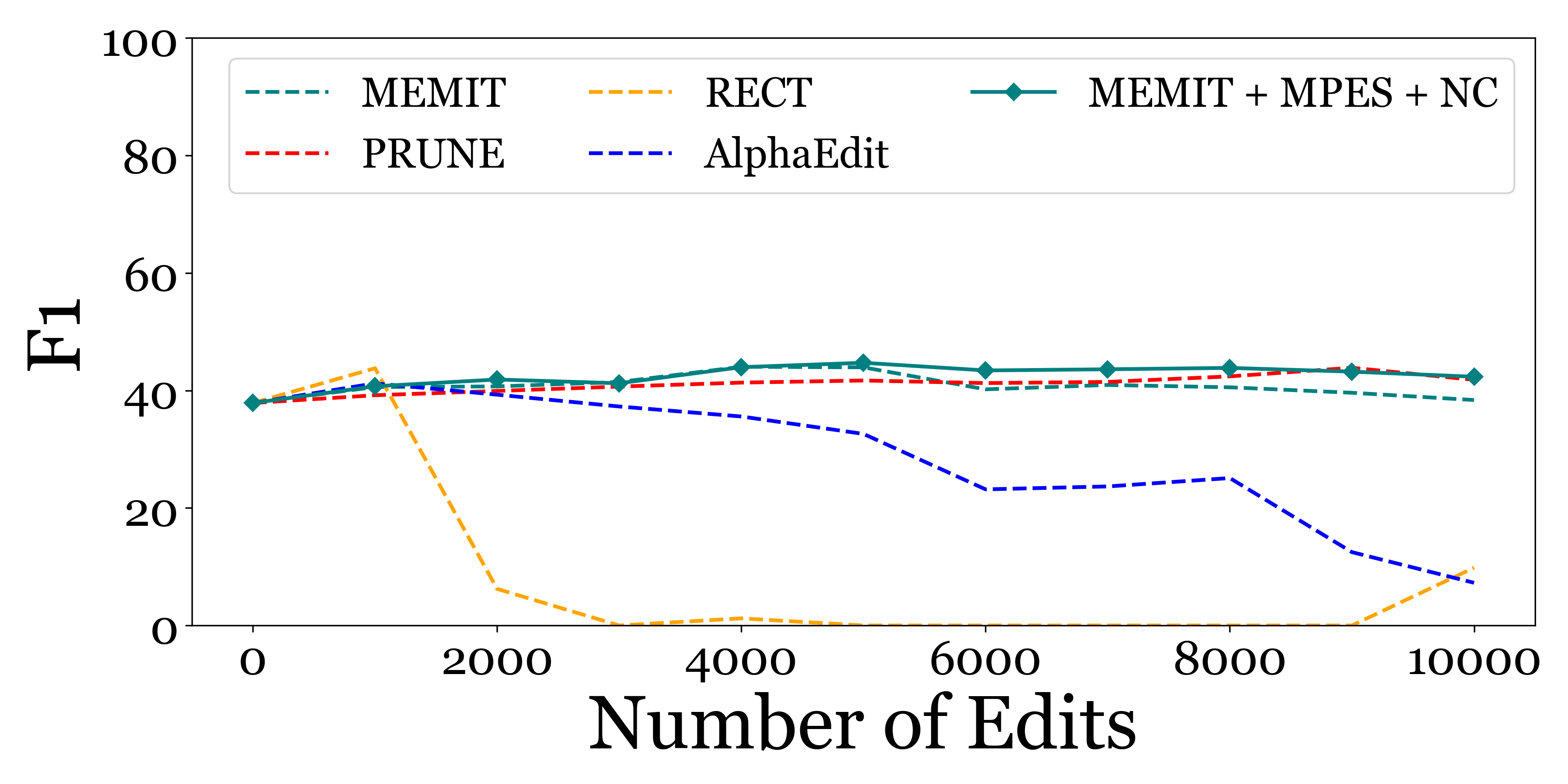} 
        \label{fig:gpt2xl-batch-10}
    }
    \subfigure[Downstream Performance for Llama2-7B with batch size 10]{
        \includegraphics[width=0.3\linewidth]{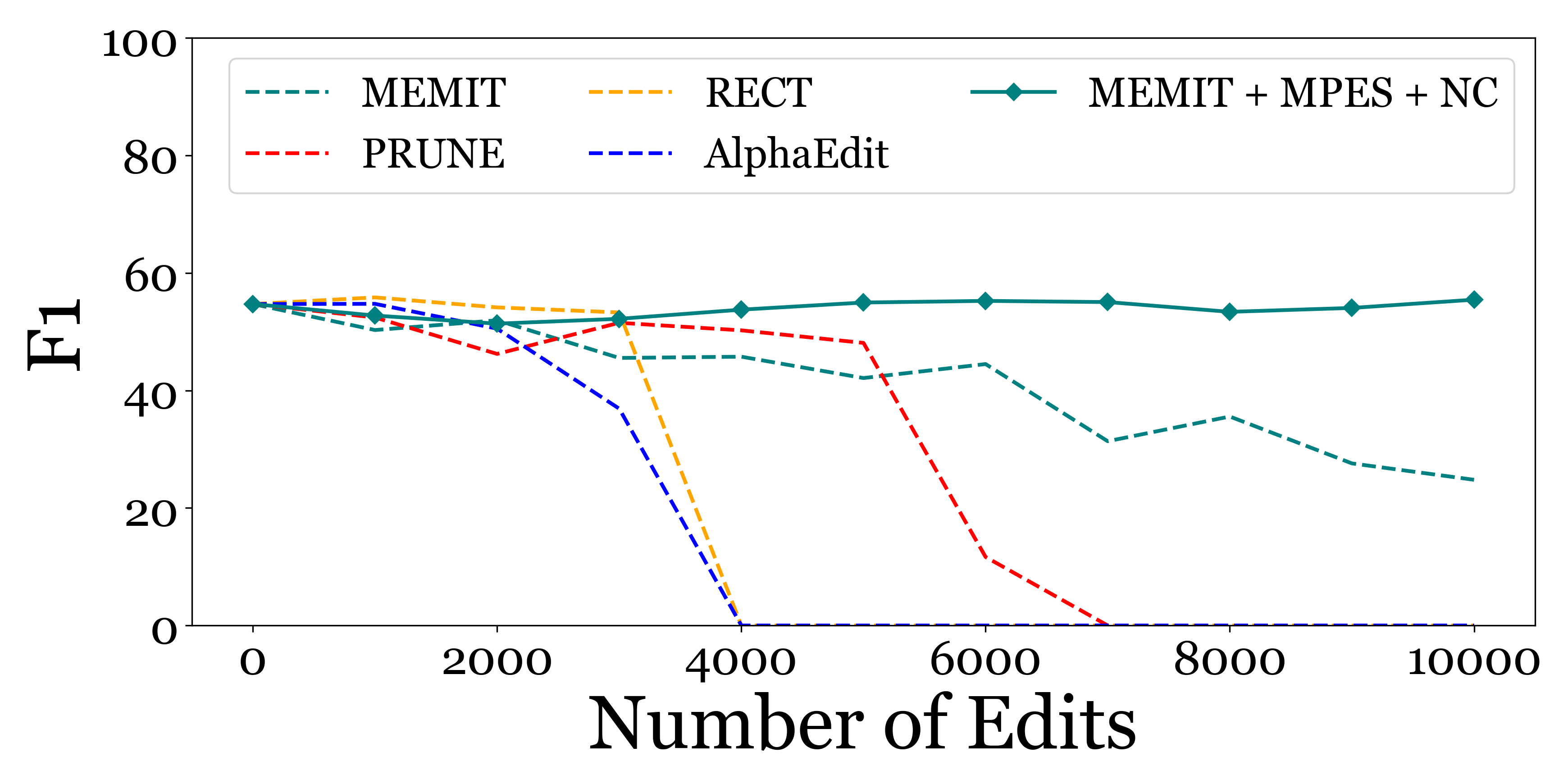} 
        \label{fig:llama2-batch-10}
    }
    \subfigure[Downstream Performance for Llama3-8B with batch size 10]{
        \includegraphics[width=0.3\linewidth]{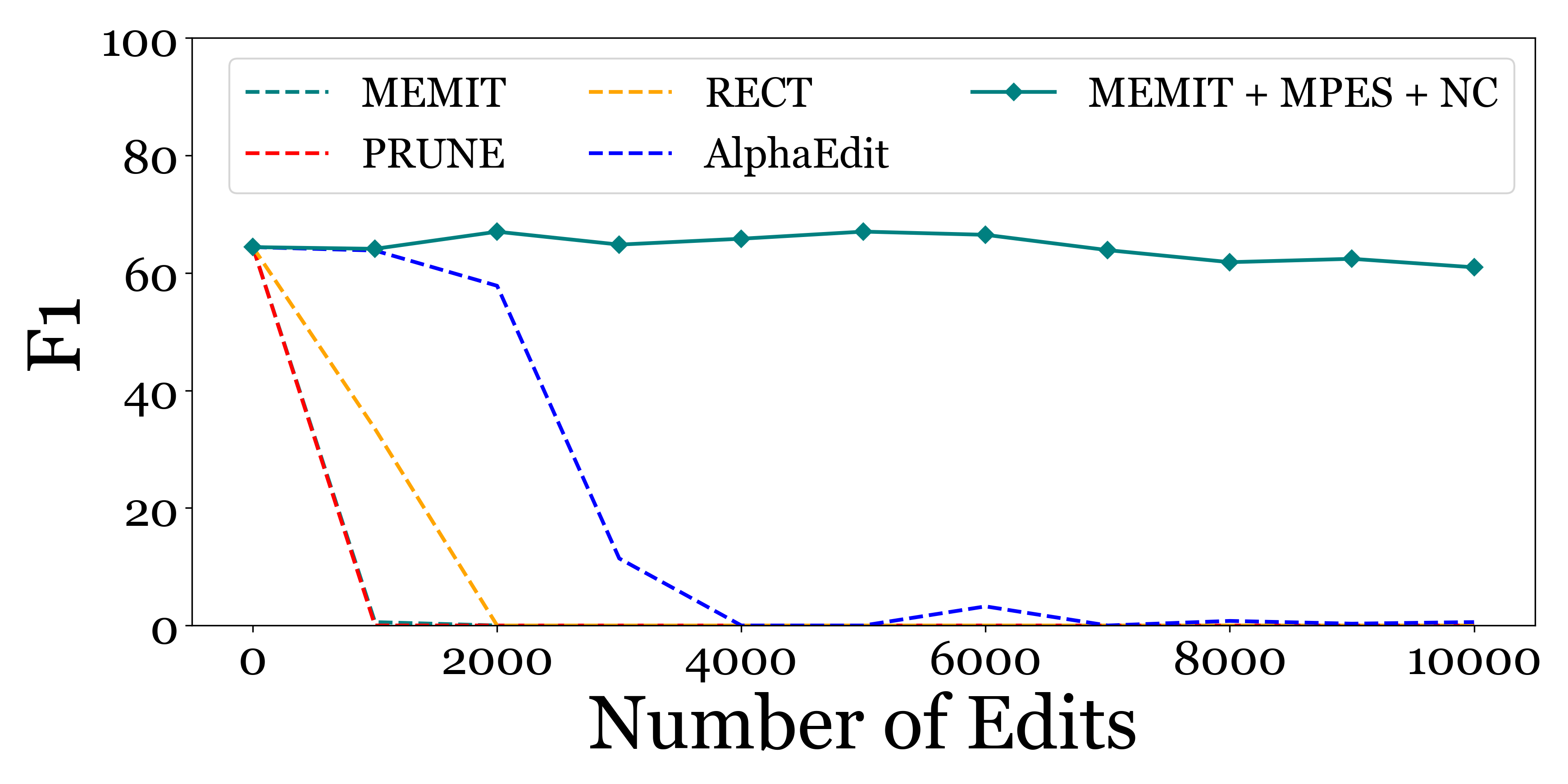} 
        \label{fig:llama3-batch-10}
    }
    \caption{Different Downstream Performance for different models with batch size 10}
    \label{fig:downstream-batch-10}
\end{figure*}

\begin{table*}[t]
\vskip 0.1in
\begin{center}
\setlength{\tabcolsep}{4pt}
\begin{adjustbox}{max width=\textwidth}
\begin{tabular}{l ccc ccc ccc ccc ccc}
\toprule
\textbf{Method} & 
\multicolumn{3}{c}{\textbf{Edit Score}} & 
\multicolumn{3}{c}{\textbf{Paraphrase Score}} & 
\multicolumn{3}{c}{\textbf{Neighborhood Score}} & 
\multicolumn{3}{c}{\textbf{Overall Score}} & 
\multicolumn{3}{c}{\textbf{Generation Entropy}} \\
\cmidrule(lr){2-4} \cmidrule(lr){5-7} \cmidrule(lr){8-10} \cmidrule(lr){11-13} \cmidrule(lr){14-16}
& GPT2-XL & Llama2-7B & Llama3-8B & GPT2-XL & Llama2-7B & Llama3-8B & GPT2-XL & Llama2-7B & Llama3-8B & GPT2-XL & Llama2-7B & Llama3-8B & GPT2-XL & Llama2-7B & Llama3-8B \\
\midrule
MEMIT                         & \textbf{94.04} & 81.04 & 49.68 & \textbf{79.91} & 64.67 & 49.29 & 57.90 & 60.95 & 51.31 & 74.22 & 67.86 & 50.08 & 517.37 & 442.59 & 373.48 \\
PRUNE                         &  61.05 & 70.80 & 49.38 & 58.05 & 62.11 & 49.63 & 50.00 & 51.86 & 51.09 & 55.96 & 60.60 & 50.02 & 579.69 & 280.83 & 340.22\\
RECT                          & 51.40 & 82.42 & 63.17 & 49.83 & 66.84 & 56.92 & 52.17 & \textbf{67.39} & 52.89 & 51.12 & 71.54 & 57.36 & 409.42 & 549.35 & \textbf{588.39}\\
AlphaEdit                     & 88.58 & 61.10 & 72.67 & 70.33 & 55.86 & 63.44 & 56.04 & 53.75 & 52.90 & 69.20 & 56.74 & 61.95 & \textbf{580.27} & 540.92 & 465.81\\
MEMIT + MPES + NC     &  93.35 & \textbf{92.57} & \textbf{88.77} & 78.66 & \textbf{82.64} & \textbf{78.19} & \textbf{59.84} & 60.43 & \textbf{60.07} & \textbf{74.75} & \textbf{76.04} & \textbf{73.71} & 523.49 & \textbf{560.16} & 523.61 \\
\bottomrule
\end{tabular}
\end{adjustbox}
\end{center}
\caption{Editing performance of our approach when compared to baseline MEMIT, AlphaEdit and MEMIT regularization method such as PRUNE and RECT using batchsize 100. }\label{tab:editing-performance-encore-main-batchsize-100}
\vskip -0.1in
\end{table*}

\begin{figure*}[ht]
    \centering
    \subfigure[Downstream Performance for GPT2-XL with batch size 100]{
        \includegraphics[width=0.3\linewidth]{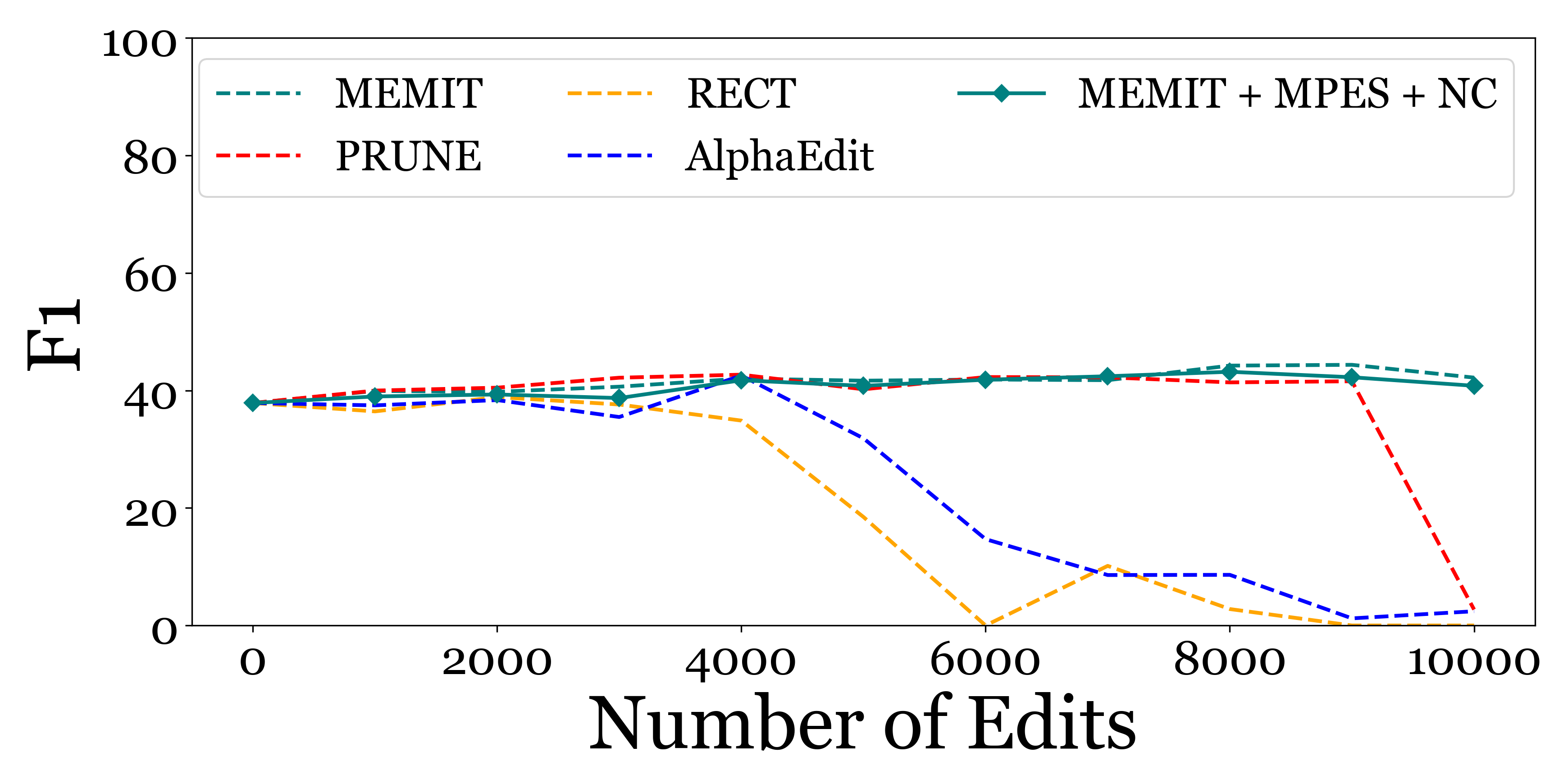} 
        \label{fig:gpt2xl-batch-100}
    }
    \subfigure[Downstream Performance for Llama2-7B with batch size 100]{
        \includegraphics[width=0.3\linewidth]{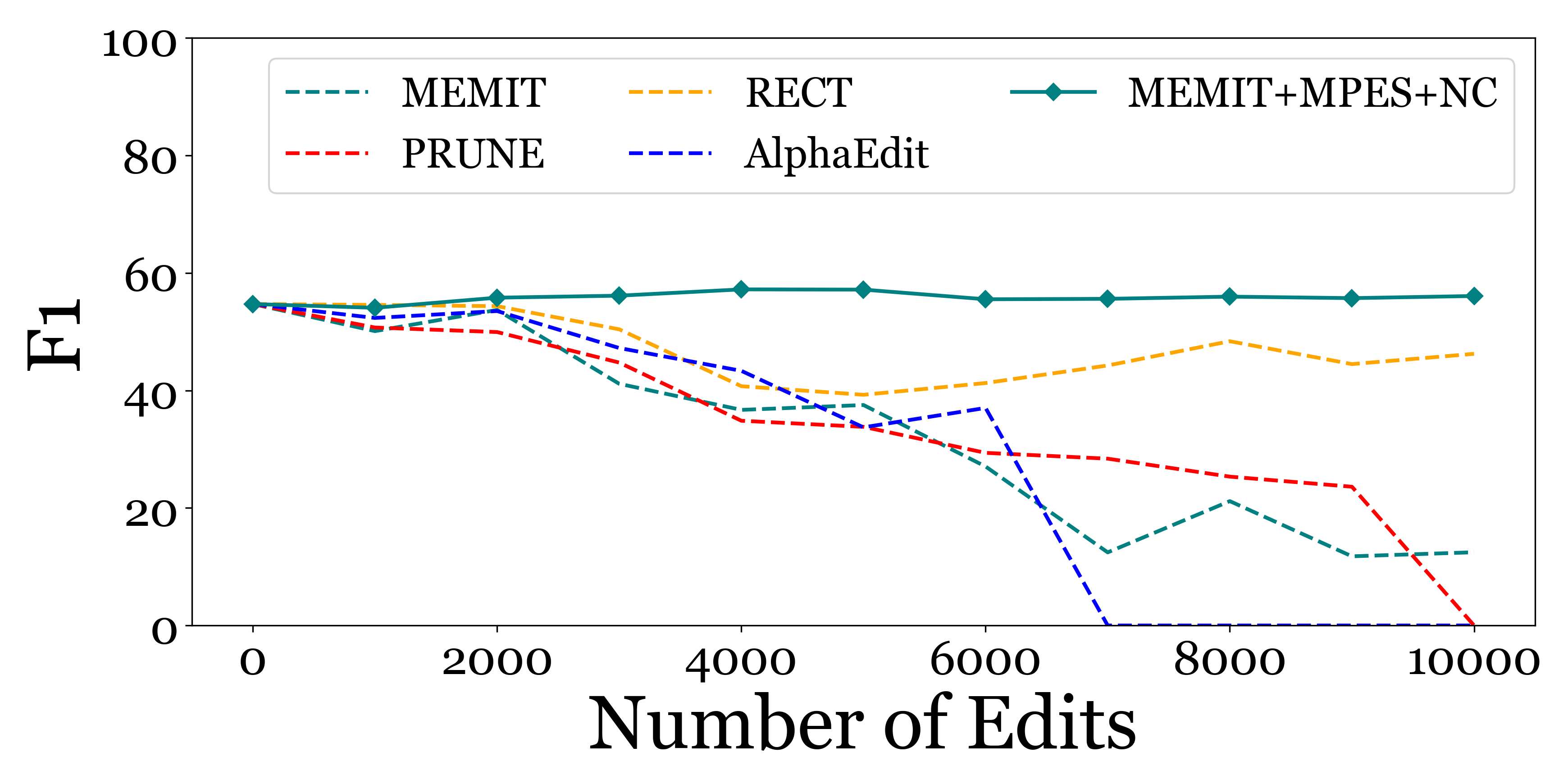} 
        \label{fig:llama2-batch-100}
    }
    \subfigure[Downstream Performance for Llama3-8B with batch size 100]{
        \includegraphics[width=0.3\linewidth]{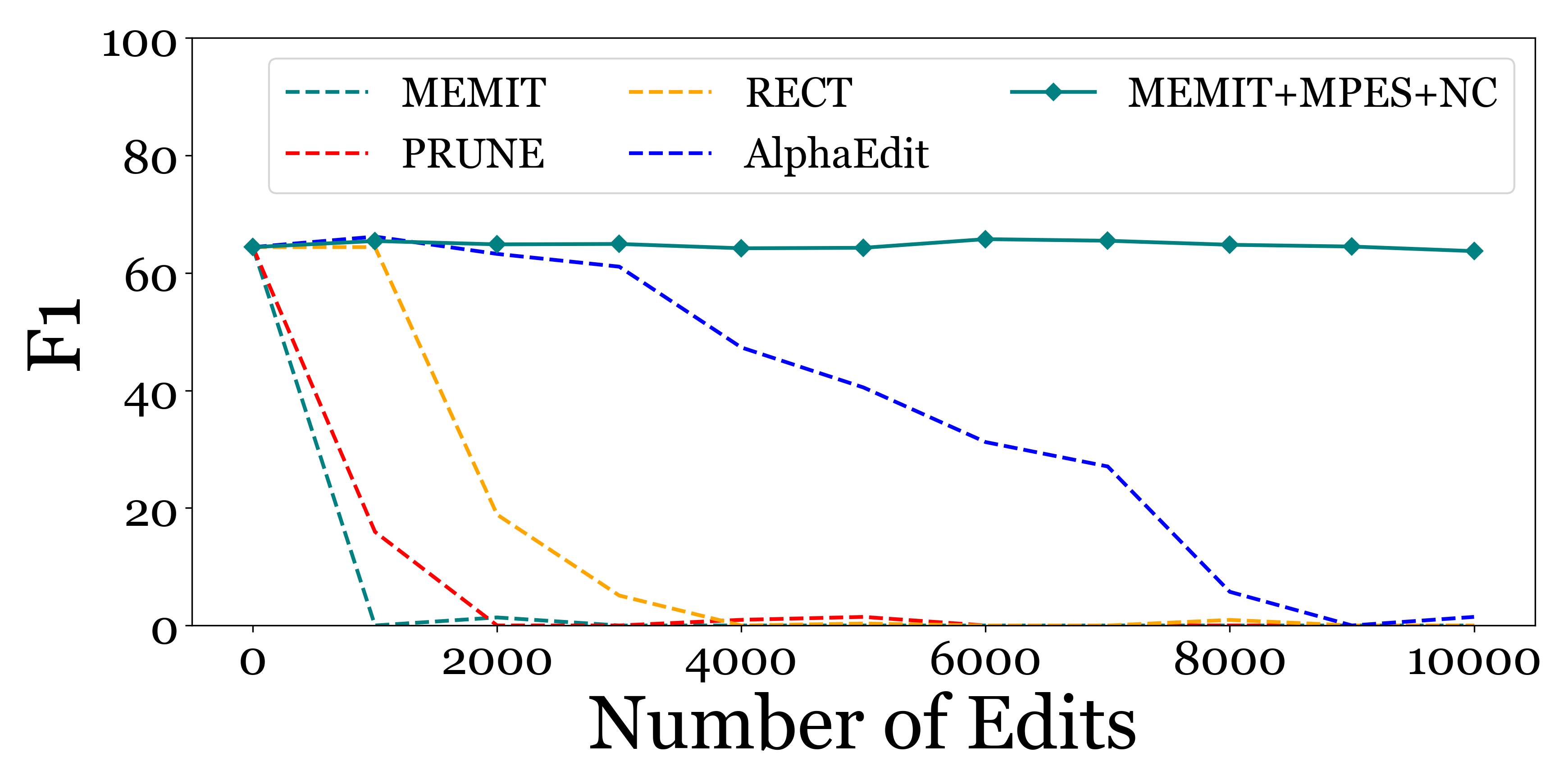} 
        \label{fig:llama3-batch-100}
    }
    \caption{Different Downstream Performance for different models with batch size 100}
    \label{fig:downstream-batch-100}
\end{figure*}

\begin{table*}[t]
\vskip 0.1in
\begin{center}
\setlength{\tabcolsep}{4pt}
\begin{adjustbox}{max width=\textwidth}
\begin{tabular}{l ccc ccc ccc ccc ccc}
\toprule
\textbf{Method} & 
\multicolumn{3}{c}{\textbf{Edit Score}} & 
\multicolumn{3}{c}{\textbf{Paraphrase Score}} & 
\multicolumn{3}{c}{\textbf{Neighborhood Score}} & 
\multicolumn{3}{c}{\textbf{Overall Score}} & 
\multicolumn{3}{c}{\textbf{Generation Entropy}} \\
\cmidrule(lr){2-4} \cmidrule(lr){5-7} \cmidrule(lr){8-10} \cmidrule(lr){11-13} \cmidrule(lr){14-16}
& GPT2-XL & Llama2-7B & Llama3-8B & GPT2-XL & Llama2-7B & Llama3-8B & GPT2-XL & Llama2-7B & Llama3-8B & GPT2-XL & Llama2-7B & Llama3-8B & GPT2-XL & Llama2-7B & Llama3-8B \\
\midrule
MEMIT                         & \textbf{93.55} & 93.94 & 74.45 & \textbf{81.31} & 76.83 & 61.25 & 59.66 & 67.64 & 57.33 & 75.47 & 78.03 & 63.56 & 558.93 & 577.69 & 457.97 \\
PRUNE                         & 92.42 & 88.58 & 62.83 & 81.27 & 70.43 & 54.11 & 58.79 & 64.64 & 52.29 & 74.75 & 73.25 & 56.05 & 548.27 & 527.05 & 461.71 \\
RECT                          & 87.49 & 87.35 & 57.36 & 71.94 & 60.86 & 52.31 & 64.9 & \textbf{73.67} & 64.34 & 73.64 & 72.37 & 57.59 & \textbf{603.91} & 566.67 & 215.44 \\
AlphaEdit                     & 92.61 & \textbf{96.11} & \textbf{91.29} & 76.78 & \textbf{87.27} & \textbf{76.88} & 56.09 & 61.63 & 68.95 & 72.03 & \textbf{78.76} & \textbf{78.00} & 587.19 & 588.09 & 593.35\\
MEMIT + MPES + NC     & 92.57 & 93.75 & 84.29 & 80.01 & 80.45 & 70.51 & \textbf{61.26} & 65.71 & \textbf{69.48} &\textbf{75.71} & 78.30 & 74.19 & 566.94 & \textbf{593.12} & \textbf{598.86} \\
\bottomrule
\end{tabular}
\end{adjustbox}
\end{center}
\caption{Editing performance of our approach when compared to baseline MEMIT, AlphaEdit and MEMIT regularization method such as PRUNE and RECT using batchsize 1000. }\label{tab:editing-performance-encore-main-batchsize-1000}
\vskip -0.1in
\end{table*}

\begin{figure*}[ht]
    \centering
    \subfigure[Downstream Performance for GPT2-XL with batch size 1000]{
        \includegraphics[width=0.3\linewidth]{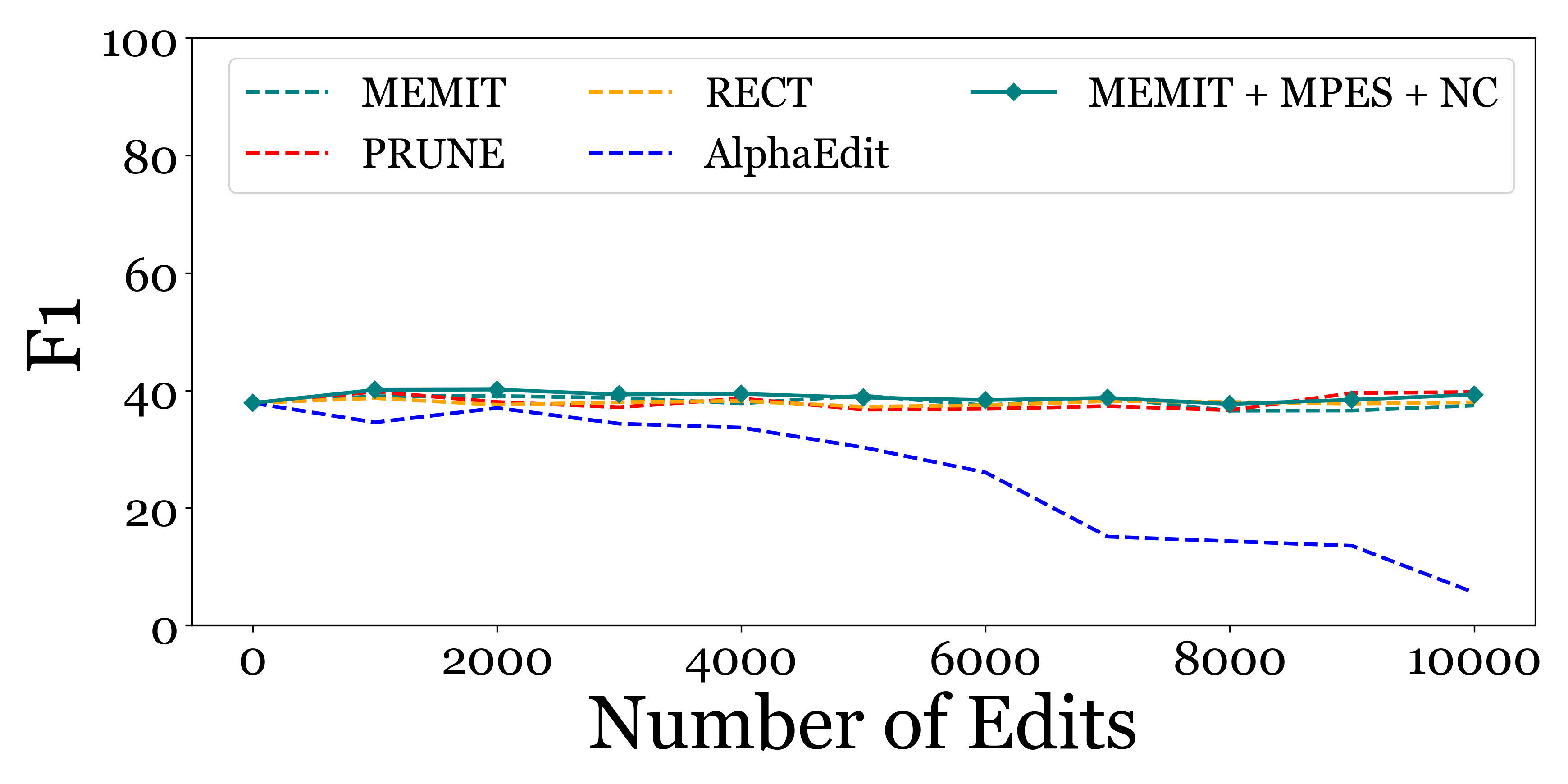} 
        \label{fig:gpt2xl-batch-1000}
    }
    \subfigure[Downstream Performance for Llama2-7B with batch size 1000]{
        \includegraphics[width=0.3\linewidth]{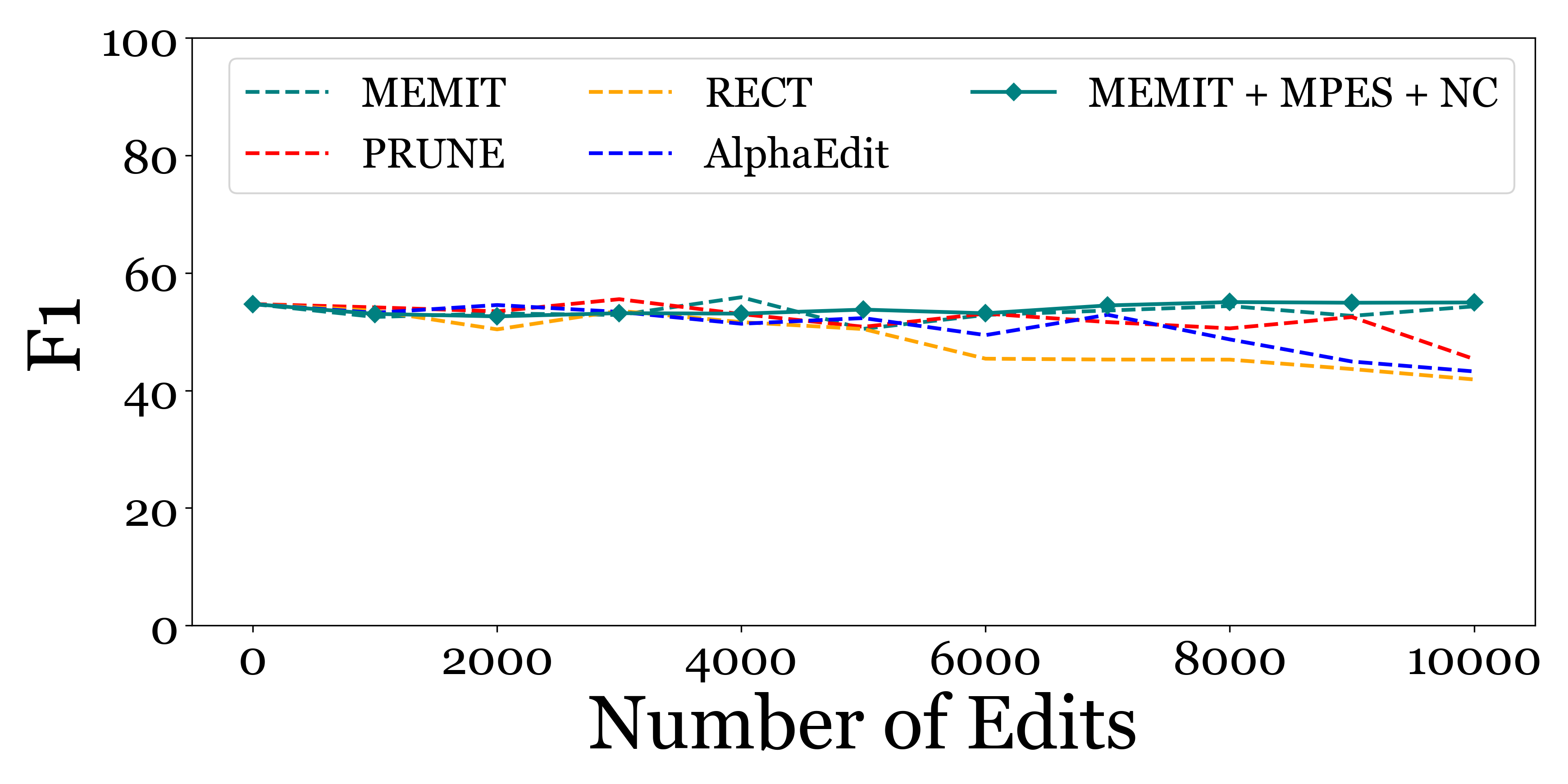} 
        \label{fig:llama2-batch-1000}
    }
    \subfigure[Downstream Performance for Llama3-8B with batch size 1000]{
        \includegraphics[width=0.3\linewidth]{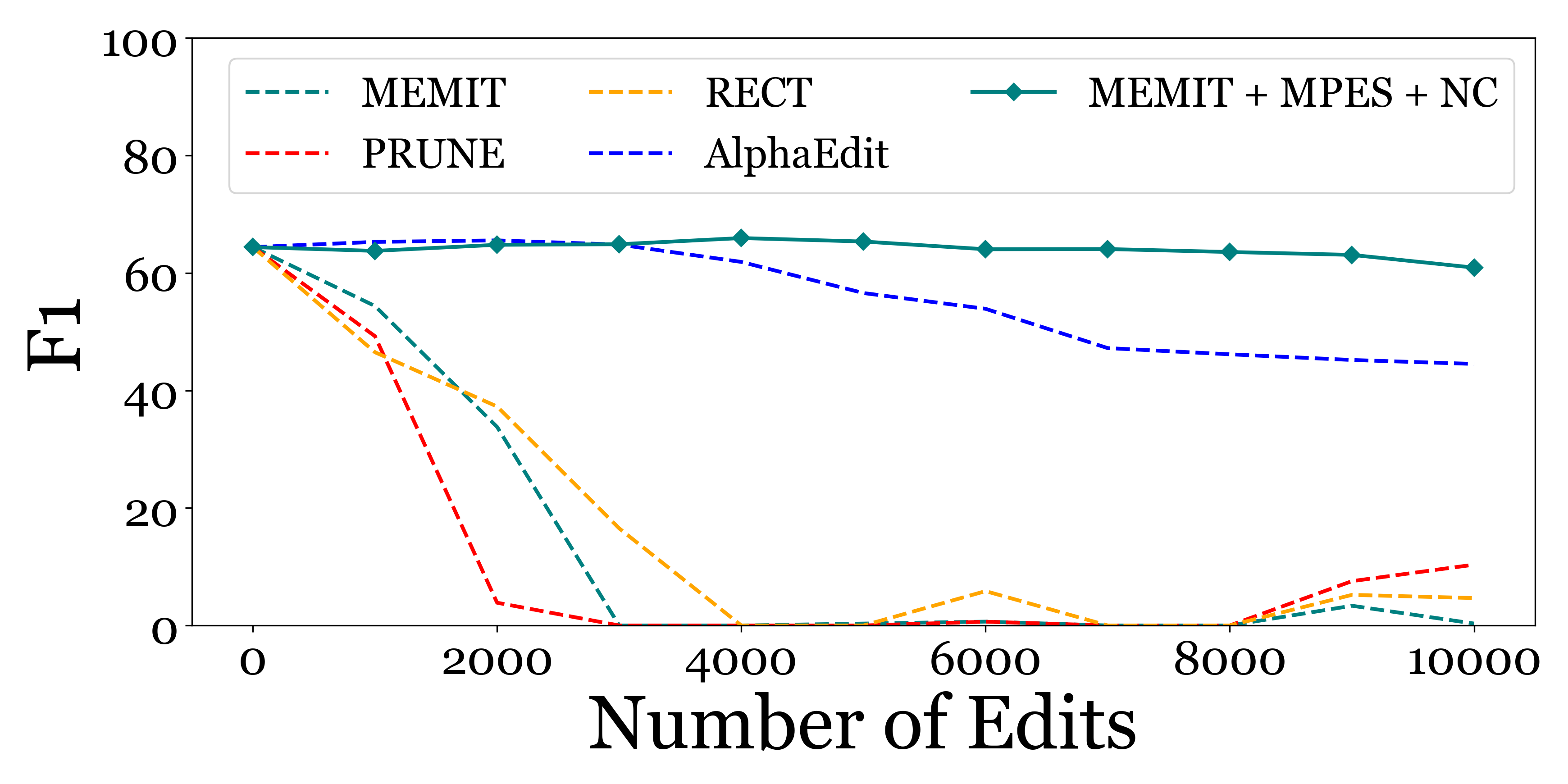} 
        \label{fig:llama3-batch-1000}
    }
    \caption{Different Downstream Performance for different models with batch size 1000}
    \label{fig:downstream-batch-1000}
\end{figure*}


\clearpage
\section{Evaluation of Downstream Performance}\label{appendix:downstream}

In this paper, we assess model degradation by measuring downstream performance at regular intervals of edits. Our evaluation suite is wide-ranging and consists of the following 6 tasks – sentiment analysis (SST2) \cite{sst2}, paraphrase detection (MRPC) \cite{mrpc}, natural language inference (NLI, RTE) \cite{nli1, nli2, nli3, nli4}, linguistic acceptability classification (CoLA) \cite{cola}, and massive multitask language understanding (MMLU) \cite{MMLU}.

For each task, we created a subset of 100 examples balanced across all multiple-choice options. The models were evaluated on the tasks above, and the accuracy score was measured every 1000 edits. In order to improve the models' initial performance and achieve meaningful signals, we provided few-shot examples. The few-shot prompt templates used for each task are shown in Figures \ref{fig:sst-prompt}-\ref{fig:nli-prompt}.


\begin{figure}[ht]
    \centering
    \scalebox{0.7}{
    \fbox{
        \parbox{0.65\textwidth}{
            Review : an exhilarating futuristic thriller-noir , minority report twists the best of technology around a gripping story , delivering a riveting , pulse intensifying escapist adventure of the first order \newline
            Sentiment : positive \newline \newline
            Review : try as i may , i ca n't think of a single good reason to see this movie , even though everyone in my group extemporaneously shouted , ` thank you ! ' \newline
            Sentiment : negative \newline \newline
            Review : the film 's performances are thrilling .  \newline
            Sentiment : positive \newline \newline
            Review : vera 's technical prowess ends up selling his film short ; he smoothes over hard truths even as he uncovers them . \newline
            Sentiment : negative \newline \newline
            Review : [input] \newline
            Sentiment :
        }
    }
    }
    \caption{Few shot prompt template used for SST-2 }
    \label{fig:sst-prompt}
\end{figure}

\begin{figure}[htbp]
    \centering
    \scalebox{0.7}{
    \fbox{
        \parbox{0.65\textwidth}{
            Question: Which expression is equivalent to 4 x 9? \newline
            (A) (4x 4) + (4x5) \newline
            (B) (4+4) x (4+5) \newline
            (C) (4+4)+(4+5) \newline
            (D) (4x 4) x (4x5) \newline
            Answer: A\newline\newline
            Question: A marketing researcher is conducting a survey in a large selling area by contacting a small group of people that is representative of all people in that area. The small, representative group is known as the \newline
            (A) population\newline
            (B) sample\newline
            (C) stratification\newline
            (D) universe\newline
            Answer: B\newline\newline
            Question: A research participant eats half a bowl of M\&M candies, and then stops eating. How would a motivation researcher using drive reduction theory explain this participant's behavior?\newline
            (A) Humans are instinctively driven to eat sugar and fat when presented to them.\newline
            (B) The Yerkes-Dodson law explains that people will eat food when presented to them, but usually in moderate amounts in order to avoid being perceived as selfish.\newline
            (C) The primary drive of hunger motivated the person to eat, and then stop when she/he regained homeostasis.\newline
            (D) The research participant was satisfying the second step on the hierarchy of needs: Food needs.\newline
            Answer: C\newline\newline
            Question: In a deductively valid argument\newline
            (A) If all the premises are true, the conclusion must be true\newline
            (B) The conclusion has already been stated in its premises\newline
            (C) If all the premises are true, the conclusion may still be false\newline
            (D) Both A and B\newline
            Answer: D\newline\newline
            Question: [input] \newline
            Answer:
        }
    }
    }
    \caption{Few shot prompt template used for MMLU}
    \label{fig:mmlu-prompt}
\end{figure}

\begin{figure}[htbp]
    \centering
    \scalebox{0.7}{
    \fbox{
        \parbox{0.65\textwidth}{
            Are the sentences paraphrases of each other. \newline
            Sentence 1: Federal regulators have turned from sour to sweet on a proposed \$2.8 billion merger of ice cream giants Nestle Holdings Inc. and Dreyer 's Grand Ice Cream Inc .\newline
            Sentence 2: Federal regulators have changed their minds on a proposed \$2.8 billion merger of ice cream giants Nestle Holdings and Dreyer 's Grand Ice Cream .\newline
            Answer: Yes\newline\newline
            Are the sentences paraphrases of each other.\newline
            Sentence 1: In the year-ago quarter , the steelmaker recorded a profit of \$16.2 million , or 15 cents per share , on sales of \$1.14 billion .\newline
            Sentence 2: In the second quarter last year , AK Steel reported a profit of \$16.2 million , or 15 cents a share .\newline
            Answer: No\newline\newline
            Are the sentences paraphrases of each other.\newline
            Sentence 1: He added : ``I 've never heard of more reprehensiblebehaviour by a doctor .\newline
            Sentence 2: The Harrisons ’ lawyer Paul LiCalsi said : “ I ’ ve never heard of more reprehensible behaviour by a doctor .\newline
            Answer: Yes\newline\newline
            Are the sentences paraphrases of each other.\newline
            Sentence 1: While dioxin levels in the environment were up last year , they have dropped by 75 percent since the 1970s , said Caswell .\newline
            Sentence 2: The Institute said dioxin levels in the environment have fallen by as much as 76 percent since the 1970s .\newline
            Answer: No\newline\newline
            Are the sentences paraphrases of each other.\newline
            Sentence 1: [input 1]\newline
            Sentence 2: [input 2]\newline
            Answer:
        }
    }
    }
    \caption{Few shot prompt template used for MRPC}
    \label{fig:mrpc-prompt}
\end{figure}

\begin{figure}[htbp]
    \centering
    \scalebox{0.7}{
    \fbox{
        \parbox{0.65\textwidth}{
            Is this sentence linguistically acceptable?\newline
            Sentence: The evidence assembled by the prosecution convinced the jury.\newline
            Answer: Yes\newline\newline
            Is this sentence linguistically acceptable?\newline
            Sentence: I live at the place where Route 150 crosses the Hudson River and my dad lives at it too.\newline
            Answer: No\newline\newline
            Is this sentence linguistically acceptable?\newline
            Sentence: The government's imposition of a fine.\newline
            Answer: Yes\newline\newline
            Is this sentence linguistically acceptable?\newline
            Sentence: Sam gave the ball out of the basket.\newline
            Answer: No\newline\newline
            Is this sentence linguistically acceptable?\newline
            Sentence: [input] \newline
            Answer: 
        }
    }
    }
    \caption{Few shot prompt template used for RTE}
    \label{fig:rte-prompt}
\end{figure}

\begin{figure}[htbp]
    \centering
    \scalebox{0.7}{
    \fbox{
        \parbox{0.65\textwidth}{
            The town is also home to the Dalai Lama and to more than 10,000 Tibetans living in exile. \newline
            Question: The Dalai Lama has been living in exile since 10,000. True or False? \newline
            Answer: True \newline\newline
            P. Prayong, who like Kevala belongs to the Theravada sect of Buddhism, chose India over other Buddhist majority nations as it is the birthplace of Gautama Buddha. \newline
            Question: P. Prayong is a member of Theravada. True or False? \newline
            Answer: False \newline\newline
            The medical student accused of murdering an erotic masseuse he met on Craigslist is drowning in more than \$100,000 in student loan debt and is so broke he can't afford to pay an attorney, according to court papers. Philip Markoff, a 23-year-old suspended Boston University medical school student, owes \$130,000 in student loans and does not get money from his parents, leaving him to lean on a taxpayer-funded attorney for his defense, according to a court document in Boston Municipal Court that labels him indigent. Markoff graduated from the State University of New York-Albany and was a second-year medical student at BU.\newline
            Question: The medical student Philip Markoff was engaged. True or False?\newline
            Answer: True\newline\newline
            Traditionally, the Brahui of the Raisani tribe are in charge of the law and order situation through the Pass area. This tribe is still living in present day Balochistan in Pakistan. \newline
            Question: The Raisani tribe resides in Pakistan. True or False? \newline
            Answer: False \newline\newline
            The latest attacks targeted the U-S embassy and a top prosecutor's office in the Uzbek capital.\newline
            Question: [input]. True or False?\newline
            Answer: 
        }
    }
    }
    \caption{Few shot prompt template used for CoLA}
    \label{fig:cola-prompt}
\end{figure}

\begin{figure}[htbp]
    \centering
    \scalebox{0.7}{
    \fbox{
        \parbox{0.65\textwidth}{
            Turkey is unlikely to become involved in, or allow U.S. forces to use Turkish territory in a Middle East war that does not threaten her territory directly. entails the U.S. to use Turkish military bases. \newline True or False? \newline Answer: False \newline \newline
            Brooklyn Borough Hall featured a Who's Who in New York's literary community during the second annual Brooklyn Book Festival. According to Brooklyn Borough President Marty Markowitz, the borough's zip code 11215 boasts more authors than anywhere else in the country. It appeared to be the case on Sunday. More than 100 authors were featured at the day-long event, including The Basketball Diaries writer Jim Carroll, former M*A*S*H star Mike Farrell, author and illustrator Mo Willems, Jack Kerouac's sometime lover and National Book Critics Circle Award recipient Joyce Johnson and PEN American Center President Francine Prose. entails the The Brooklyn Book Festival is held in Brooklyn Borough every year. \newline True or False? \newline Answer: True \newline\newline
            NASA's Saturn exploration spacecraft, Cassini , has discovered an atmosphere about the moon Enceladus . This is the first such discovery by Cassini, other than Titan , of the presence of an atmosphere around a Saturn moon. entails the Titan is the fifteenth of Saturn's known satellites.\newline True or False? \newline Answer: False \newline\newline
            Dr. Eric Goosby, a pioneer in the fight against AIDS, is President Obama's choice to run the American effort to combat the disease globally, the White House announced Monday. The President's Emergency Plan For AIDS Relief, known as Pepfar, was championed by President George W. Bush. It is expected to spend \$48 billion over the next five years and is credited with markedly reducing the disease's death rate. Its prevention policy has been controversial because of its emphasis on socially conservative methods. With a new administration and a Democratic majority in the House, organizations seeking prevention choices beyond abstinence and fidelity — including a renewed commitment to distributing condoms — are eager to try to rewrite the guidelines. entails the Pepfar is committed to fighting AIDS. \newline True or False? \newline Answer: True\newline\newline
            [input] \newline True or False? \newline Answer:
        }
    }
    }
    \caption{Few shot prompt template used for NLI}
    \label{fig:nli-prompt}
\end{figure}

\end{document}